%% file: update_arxiv.tex
\documentclass[11pt]{article}
\usepackage{amsgen,amsmath,amstext,amsbsy,amsopn,amssymb,mathabx,mathrsfs,amsthm,bm,bbm,romannum,enumitem,natbib,xr,multirow}

\usepackage{caption}
\usepackage{dsfont}
\usepackage[dvips]{graphicx}
\usepackage{subcaption}
\usepackage[ruled,vlined,linesnumbered,resetcount]{algorithm2e}
\usepackage{hyperref}
\hypersetup{
	colorlinks=true,
	linkcolor=red,
	urlcolor=blue,
	citecolor=blue
}
\usepackage[depth=2]{bookmark}

\usepackage[margin=1in]{geometry}

\usepackage[utf8]{inputenc}

\input{preamble}

\newcommand{\step}[1]{\medskip\noindent\textbf{Step #1.}}

\usepackage{booktabs}
\usepackage{longtable}
\usepackage{array}
\usepackage{multirow}
\usepackage[table]{xcolor}
\usepackage{wrapfig}
\usepackage{float}
\usepackage{colortbl}
\usepackage{pdflscape}
\usepackage{tabu}
\usepackage{threeparttable}
\usepackage{threeparttablex}
\usepackage[normalem]{ulem}
\usepackage{makecell}

\title{Consistency of Lloyd's Algorithm Under Perturbations}

\author{
Hui Shen$^{1}$,
Dhruv Patel$^{2}$,
Shankar Bhamidi$^{2}$,
Vladas Pipiras$^{2}$,
Yufeng Liu$^{3}$
}

\date{}

\begin{document}
	\pagenumbering{arabic}
\maketitle

\vspace{-1.5em}

\begin{center}
{\small
$^{1}$ Department of Mathematics and Statistics, McGill University\\
$^{2}$ Department of Statistics and Operations Research, University of North Carolina at Chapel Hill\\
$^{3}$ Department of Statistics, University of Michigan\\[0.5em]
\texttt{hui.shen2@mail.mcgill.ca},
\texttt{dhruvpat@live.unc.edu},
\texttt{bhamidi@email.unc.edu},\\
\texttt{pipiras@email.unc.edu},
\texttt{yufliu@umich.edu}
}
\end{center}

    \begin{abstract}

In unsupervised learning, Lloyd's algorithm is one of the most widely used clustering algorithms. It has inspired a plethora of work investigating the correctness of the algorithm under various settings with ground truth clusters. It has been shown that the mis-clustering rate of Lloyd’s algorithm on independent samples from a sub-Gaussian mixture is exponentially bounded after $O(\log(n))$ iterations, given proper initialization. However, in many applications, the true samples are unobserved and need to be learned from the data via pre-processing pipelines such as spectral methods on appropriate data matrices. We show that the mis-clustering rate of Lloyd's algorithm on perturbed samples from a sub-Gaussian mixture is also exponentially bounded after $O(\log(n))$ iterations under the assumptions of proper initialization and that the perturbation is small relative to the sub-Gaussian noise. We further establish theoretical guarantees for initialization schemes such as $k$-means++, provide implications for cluster significance testing (e.g., SigClust). 
We apply these results to offer theoretical guarantees on the mis-clustering rate of Lloyd’s algorithm in diverse applications, including high-dimensional time series, multi-dimensional scaling, and community detection in sparse networks via spectral clustering. 
\end{abstract}


\section{Introduction}\label{s:intro}

In many modern data-analytic pipelines, the final clustering is obtained by running $k$-means on a learned low-dimensional representation rather than on the raw data. Examples include principal component analysis (PCA) scores, spectral embedding matrices, multidimensional scaling (MDS) coordinates, and factor estimates in time series. Lloyd’s algorithm is typically used in these settings because it scales well and is easy to implement.

Most existing theory for $k$-means clustering primarily analyzes the properties of its global optimum rather than the algorithmic behavior of iterative procedures such as Lloyd’s method. Under independent sampling and assuming uniqueness of the population minimizer, \cite{pollard:1981} establishes the consistency of $k$-means clustering and \cite{pollard:1982a} proves a central limit theorem for the empirical cluster means. Subsequent work has studied stability of the global solution, such as uniqueness \citep{rakhlin:2007}, multiplicative perturbation stability \citep{awasthi:2012}, and additive perturbation stability \citep{dutta2017clustering}. However, these results focus on the global optimum rather than the convergence behavior of Lloyd’s algorithm, which may reach a local minimum.

To analyze the algorithmic performance of Lloyd's method in settings with ground-truth clusters, a natural metric is the misclustering rate. In the clean setting of independent sampling from a well-separated sub-Gaussian mixture, \cite{lu2016statistical} showed that Lloyd's algorithm, given a sufficiently accurate initialization, achieves an exponentially decreasing misclustering rate after $O(\log n)$ iterations. These guarantees assume access to the raw mixture samples. When Lloyd’s algorithm is run on estimated or embedded data, this assumption is violated. 

When Lloyd’s algorithm is applied to estimated or embedded data, the input is contaminated by perturbations that are dependent across observations. These perturbations propagate through the assignment and centroid-update steps of Lloyd’s iterations, placing such pipelines outside the scope of existing convergence analyses. This issue arises in a wide range of high-dimensional, network-based, and latent-variable settings. Yet existing theory provides very little guidance: the dependence structure of the perturbations and their interaction with Lloyd’s iterations cannot be addressed by classical analyses that treat noise as independent or purely additive.

\subsection{Informal description of our results}
Our results provide a unified framework for analyzing Lloyd’s algorithm when clustering is performed on estimated representations rather than raw observations. We summarize below the main contributions of this paper.

\begin{enumerate}[label = (\alph*), left=0pt]

\item \textbf{Lloyd’s algorithm under perturbed samples.}
We prove that when the observed data consist of perturbed versions of samples from a well-separated sub-Gaussian mixture, Lloyd’s algorithm achieves an exponentially decaying misclustering rate after $\mathcal{O}(\log n)$ iterations, dutta2017clustering the perturbation is sufficiently small relative to the cluster separation. Our result generalizes and recovers the clean-sample guarantee of \cite{lu2016statistical} as a special case.

\item \textbf{Initialization guarantees.}
Our main theorem requires a moderately accurate initialization. In a simplified setting with two balanced clusters and bounded perturbations,
we show that the $k$-means$++$ initializer produces cluster centers that satisfy the required accuracy conditions with high probability when the clusters are sufficiently separated. This result demonstrates that the initialization assumptions in our main theorems are non-vacuous and can be met by a standard and widely used algorithm under controlled perturbations.

\item \textbf{Implications for cluster significance testing.}
Recent work has emphasized assessing the statistical significance of discovered clusters in unsupervised learning \citep{liu2008statistical, shen2024statistical}. We show that our perturbation-based analysis yields consistency guarantees for significance testing procedures such as SigClust \citep{liu2008statistical}, thereby providing theoretical justification for their use when the input to the clustering algorithm is estimated rather than directly observed.

\item \textbf{Canonical applications.}
We illustrate the implications of our results across several representative settings, including high-dimensional time series with latent low-rank structure, datasets undergoing dimension reduction via multidimensional scaling, and community detection in networks through spectral embedding. These applications demonstrate the broad relevance of our framework in modern clustering pipelines.

\end{enumerate}

\subsection{Organization of the paper}
In Section~\ref{s:clustering_model}, we introduce the general clustering framework and perturbation model studied in this paper. Section~\ref{s:theoretical_results} presents the main theoretical results and their proofs, while Section~\ref{s:applications} illustrates their implications across several canonical settings. Appendices~A and~B contain technical proofs for the main results and applications, respectively.

\subsection{Notation}\label{s:notation}
We collect commonly used notation below for reference. All vectors in this paper will be treated as column vectors. Unless otherwise stated, we will use $\|\bx\|$ to denote the Euclidean $2$-norm of a vector $\bx \in \bbR^r$ and for two vectors $\bx_1,\bx_2\in\bbR^r$ write $\langle \bx_1,\bx_2\rangle := \bx_1^{\top}\bx_2$ for the inner product between these two vectors.   For square matrix $\Ab\in\bbR^{r\times r}$, write $\|\Ab\|$ for the spectral norm of $\Ab$. The $2\rightarrow\infty$ bound for any matrix $\Xb \in \bbR^{n\times m}$ is defined as 
$$\|\Xb\|_{2\to\infty} = \max_{i\in [n]} \|\Xb_i\|,$$ 
where $\Xb_i \in \bbR^m$ is the column vector corresponding to the $i$-th row of $\Xb$. Let $\lambda_j(\Ab)$, $\lambda_{\max}(\Ab)$ and $\lambda_{\min}(\Ab)$ be the $j$-th largest, maximal and minimal eigenvalues of any real symmetric matrix $\Ab$. For any finite set $S$, write $|S|$ for the cardinality of $S$. A random vector $\bx\in\bbR^r$ is said to be sub-Gaussian with parameter $\sigma > 0$, denoted as subG($\sigma^2$), if for all $\ba \in\bbR^r$,
$$\bbE e^{\langle \bx - \bbE(\bx), \ba \rangle} \leq e^{\frac{\sigma^2 \|\ba\|^2}{2}}. $$ For sequences $\left\{a_n\right\}$ and $\left\{b_n\right\}$, we write $a_n=o\left(b_n\right)$ or $a_n \ll b_n$ if $\lim _n a_n / b_n=0$, and write $a_n=O\left(b_n\right), a_n \lesssim b_n$ or $b_n \gtrsim a_n$ if there exists a constant $C$ such that $a_n \leq C b_n$ for all $n$. We write $a_n = w(b_n)$ if $b_n = o(a_n)$ and $a_n = \Omega(b_n)$ if $b_n = O(a_n)$. We write $a_n = \Theta(b_n)$ if $a_n = O(b_n)$ and $a_n = \Omega(b_n)$. We write $a_n \asymp b_n$ if $a_n \lesssim b_n$ and $a_n \gtrsim b_n$. We write $o_p(\cdot), O_p(\cdot)$ for the corresponding probabilistic analogs, so for example, for a sequence of random variables $(X_n;\, n\geq 1)$ and constants $(b_n:n\geq 1)$ with $b_n\uparrow \infty$, we write $X_n=o_p(b_n)$ when $X_n/b_n\probc 0$ as $n\rightarrow\infty$. Throughout, $C, C_1, C_2, \ldots$ are constants that can change from line to line. We let $\weakc$, $\probc$, and $\convas$, respectively denote convergence in distribution, convergence in probability, and almost sure convergence.

\section{Clustering Model} \label{s:clustering_model}

\subsection{The $k$-means objective and Lloyd’s algorithm}

Given $n$ points $\{\by_i : i \in [n]\} \subseteq \mathbb{R}^r$ and a fixed number of clusters $K \ge 2$, the $k$-means problem seeks a partition 
$\bC = \{\cC_1, \dots, \cC_K\}$ minimizing
\begin{equation}\label{kmeans_cost1}
    \mathrm{Cost}(\bC)
    := \sum_{k=1}^K \frac{1}{n_k}
    \sum_{\by_i,\,\by_j \in \cC_k} \|\by_i - \by_j\|^2,
\end{equation}
where $n_k = |\cC_k|$.  Writing $\bmu_k$ for the empirical mean of $\cC_k$ and $z_i$ for the cluster label of $\by_i$, the objective is
\begin{equation}\label{kmeans_cost2}
    \mathrm{Cost}(\bC)
    = \frac{1}{n} \sum_{i=1}^n \|\by_i - \bmu_{z_i}\|^2,
\end{equation}
leading to the optimization problem
\begin{equation}\label{kmeans}
    \min_{\bC,\, z} \frac{1}{n} \sum_{i=1}^n \|\by_i - \bmu_{z_i}\|^2.
\end{equation}
This is a nonconvex and NP-hard problem \citep{mahajan2009}. As a result, practical implementations rely on iterative heuristics such as Lloyd’s algorithm \citep{lloyd1982}.

Lloyd’s algorithm proceeds as follows: starting from initial cluster means
$\{\widehat{\bmu}_k^{(0)}\}$, it alternates between

\noindent\textbf{Assignment:}
\[
    \widehat{z}^{(s+1)}_i
    = \arg\min_{k\in[K]} \|\by_i - \widehat{\bmu}_k^{(s)}\|.
\]

\noindent\textbf{Update:}
\[
    \widehat{\bmu}_k^{(s+1)} 
    = \frac{1}{|\{i : \widehat{z}_i^{(s+1)} = k\}|}
      \sum_{i : \widehat{z}_i^{(s+1)} = k} \by_i.
\]
Each iteration decreases \eqref{kmeans_cost2} and the algorithm converges to a local optimum, whose quality depends on the initialization and the geometry of the data. 

\subsection{Sub-Gaussian mixture models}

We assume independent latent samples $\{\by_i^*\}_{i\in[n]}$ drawn from a $K$-component mixture
\begin{equation}\label{mixture_model}
    \cF = \sum_{k=1}^K p_k \cF_k,    
\end{equation}
where $p_k>0$ and each $\cF_k$ is sub-Gaussian in $\mathbb{R}^r$ with common sub-Gaussian parameter $\sigma$. Let $z_i$ denote the true cluster label and $\bmu_k = \mathbb{E}(\cF_k)$ the component means. Then
\begin{equation}\label{y_i^*}
    \by_i^* = \bmu_{z_i} + \bw_i,
\end{equation}
where $\bw_i$ are independent mean-zero sub-Gaussian noise vectors.

\subsection{Perturbed observations and matrix representation}

In practice, the true samples $\{\by_i^*\}$ are often unobserved. Instead, we observe perturbed points
\begin{equation}\label{eqn:model}
    \by_i = \by_i^* + \be_i,
    \qquad 
    \|\be_i\| \le \epsilon,
\end{equation}
where the perturbations $\be_i$ may be dependent across $i$. This framework captures settings such as PCA, MDS, and spectral embedding, where $\by_i$ is an estimated representation of an underlying latent point.

Let $\Yb = (\by_1,\ldots,\by_n)^\top$, $\Mb = (\bmu_1,\ldots,\bmu_K)^\top$, and $\Zb$ the membership matrix
with $\Zb_{i,k}=1$ if $z_i=k$.  
Writing $\Eb_1 = (\be_1,\ldots,\be_n)^\top$ and $\Eb_2 = (\bw_1,\ldots,\bw_n)^\top$, the model becomes
\begin{equation}\label{eqn:model_matrix}
    \Yb = \Zb \Mb + \Eb_1 + \Eb_2,
\end{equation}
where $\|\Eb_1\|_{2\to\infty} \le \epsilon$ and $\Eb_2$ is row-independent sub-Gaussian. The tuple $(\Zb,\Mb,\epsilon,\sigma)$ characterizes the model.

\begin{rem}
Uniform error bounds of the form $|\by_i - \by_i^*| \le \epsilon$ arise in many estimation pipelines such as PCA, MDS, and spectral embeddings; see Section~\ref{s:applications}. Since $k$-means is invariant under orthogonal transformations, the performance of Lloyd's algorithm on $\Yb$ is identical to its performance on $\Yb \Ob$ for any orthonormal matrix $\Ob$. For convenience, we write $\mathcal{X} = \{\by_i : i \in [n]\}$ for the data used as input to Lloyd’s algorithm.
\end{rem}

\section{Main Results} \label{s:theoretical_results}

We now state our main guarantees for Lloyd’s algorithm under the perturbed sub-Gaussian mixture model from Section~\ref{s:clustering_model}. Under explicit signal-to-noise and initialization conditions, we show that Lloyd’s iterations drive the misclustering rate down exponentially fast and reach $A_s$ of order $\exp(-c\,\Delta_{\min}^2/\sigma^2)$ (and the corresponding perturbation-controlled term) after $s=\Theta(\log n)$ steps. Compared with the clean-sample analysis of \cite{lu2016statistical}, our bounds account for the additional bounded perturbations in the observed inputs.

\subsection{Setup and error metrics} 
\label{s:setup_metrics}

We begin by introducing the notation used to describe the evolution of Lloyd's algorithm.  
For each cluster $k \in [K]$, let
\[
\mathcal{C}_k = \{i : z_i = k\}, \qquad n_k = |\mathcal{C}_k|, 
\qquad \alpha = \min_{k\in[K]} \frac{n_k}{n}
\]
denote the true cluster assignments, cluster sizes, and minimal cluster proportion.

After $s$ iterations of Lloyd’s algorithm applied to the perturbed observations
$\{\by_i\}_{i\in[n]}$, let $\widehat{z}^{(s)}$ be the estimated labels and 
$\{\widehat{\bmu}^{(s)}_k\}$ the estimated cluster centers.  
To track misclassifications, define the cross-assignment sets
\[
U_{lk}^{(s)} = \{ i : z_i = l,\; \widehat{z}_i^{(s)} = k \},
\qquad 
\widehat{n}_{lk}^{(s)} = |U_{lk}^{(s)}|,
\qquad
\widehat{n}_k^{(s)} = |\{ i : \widehat{z}_i^{(s)} = k \}|.
\]
Here, $U_{lk}^{(s)}$ denotes the subset of samples from true cluster $l$ that are assigned to cluster $k$ at iteration $s$. The initial centers and corresponding membership function correspond to the $s\equiv0$ setting.

At iteration $s$, the misclustering rate is
\[
A_s = \min_{\pi\in S_K} \frac{1}{n}
\sum_{i=1}^n \mathbb{I}\{\widehat{z}^{(s)}_i \neq \pi(z_i)\},
\]
where $S_K$ is the class of all permutations on $[K]$ and the cluster-wise error is
\begin{equation}\label{G_s}
G_{s}=\max _{k \in[K]}\left\{\frac{1}{\widehat{n}_{k}^{(s)}} \sum_{l \neq k} \widehat{n}_{l k}^{(s)}, \frac{1}{n_{k}} \sum_{l \neq k} \widehat{n}_{k l}^{(s)}\right\}.
\end{equation}
Here $A_s$ is the overall misclustering rate up to label permutation, while $G_s$ controls the worst cluster-wise contamination. By definition, $A_s, G_s \le 1$ and $G_s \alpha \le A_s$.

To analyze the misclustering rate, we impose standard assumptions on the mixture model, including separation of the population means and suitable signal-to-noise conditions. We define the minimal/maximal separation
\begin{equation}\label{eqn:Delta}
\Delta_{\min} = \min_{l\neq k} \|\bmu_l - \bmu_k\|,
\qquad
\Delta_{\max} = \max_{l\neq k} \|\bmu_l - \bmu_k\|,
\end{equation}
the scaled center error
\begin{equation}\label{Gamma_s}
\Gamma_{s}=\max _{k \in[K]} \frac{\|\widehat{\boldsymbol{\bmu}}_{k}^{(s)}-\boldsymbol{\bmu}_{k}\|}{\Delta_{\min}}. 
\end{equation}
Define the signal-to-noise ratios as
\begin{equation}\label{rho}
\rho_\sigma = \frac{\Delta_{\min}}{\sigma}\sqrt{\frac{\alpha n}{n+Kr}}, \qquad
\rho_\epsilon = \frac{\Delta_{\min}}{\epsilon}\sqrt{\alpha}.
\end{equation}
If $\sigma = 0$ or $\epsilon = 0$, we set $\rho_{\sigma} = \infty$ or $\rho_{\epsilon} = \infty$, respectively.  
We write $\delta(n,\sigma,\Delta_{\min},\epsilon)$ for the failure probability appearing below,
\begin{align*}\delta(n,\sigma,\Delta_{\min},\epsilon) = & \frac{1}{n}+2 \exp \left(-\frac{\Delta_{\min}}{\sigma}\right)  + 2 \exp\left( -\frac{\Delta_{\min}}{\sqrt{\epsilon\sigma}} \right).
\end{align*}
Note that $\delta \to 0$  as $\rho_{\sigma},\rho_{\epsilon},n \to \infty$.

\subsection{Misclustering guarantees for Lloyd’s algorithm}\label{s:lloyd_guarantee}

The following assumptions ensure sufficiently large cluster sizes and sufficiently strong separation relative to both noise sources:
\begin{align}
n\alpha &\ge C_1 K \log n, \label{alpha_assump} \\
\rho_\sigma &\ge C_2 \sqrt{K}, \label{rho_sigma_assump}\\
\rho_\epsilon &\ge C_3 \sqrt{K},\label{rho_eps_assump}\\
\frac{\Delta_{\min}^2}{\epsilon\sigma} &\ge C_4 r, 
\label{interaction_assump} \\
\frac{\Delta_{\max}}{\Delta_{\min}} &\le C_5.  \label{aspect_assump}
\end{align}
Lloyd’s algorithm requires a sufficiently accurate initialization. We assume that one of the following conditions holds: 
\begin{align}
G_0 &\le \Big(\tfrac12 - \tfrac{\sqrt{6}+1}{\rho_\sigma}
- \tfrac{(c_{\epsilon}+1)\sqrt{\alpha}+1}{\rho_\epsilon}
- \alpha^{-1/4}\sqrt{\tfrac{\sigma}{\Delta_{\min}}} \Big)
\frac{\Delta_{\min}}{\Delta_{\max}}, \label{initial_conditions1}\\
\Gamma_0 &\le \tfrac12 - \tfrac{1}{\rho_\sigma}
- \tfrac{c_{\epsilon}\sqrt{\alpha}+1}{\rho_\epsilon}
- \alpha^{-1/4}\sqrt{\tfrac{\sigma}{\Delta_{\min}}}, \label{initial_conditions2}
\end{align}
for some constant $c_{\epsilon}>1$. We can now state our main guarantee.

\begin{theorem}[Misclustering bound for perturbed Lloyd]\label{thm-lloyd-guarantee}
Under assumptions \eqref{alpha_assump}–\eqref{aspect_assump} and either initialization condition \eqref{initial_conditions1} or \eqref{initial_conditions2}, we have
\begin{equation}\label{thm-lloyd-A_s_bound}
    A_s \le 
    \max \left\{
    \exp\!\left(-\frac{\Delta_{\min}^2}{16\sigma^2}\right),
    \;\;
    \exp\!\left(-\frac{\Delta_{\min}^2}{8\epsilon\sigma}\right)
    \right\}
    \qquad\text{for all } s\ge 4\log n,
\end{equation}
with probability at least $1 - \delta(n,\sigma,\Delta_{\min},\epsilon)$, where
\[
\delta(n,\sigma,\Delta_{\min},\epsilon)
= \frac{1}{n}
+ 2\exp(-\Delta_{\min}/\sigma)
+ 2\exp\!\left(-\frac{\Delta_{\min}}{\sqrt{\epsilon\sigma}}\right).
\]
\end{theorem}

\begin{rem}[Interpretation of the SNR conditions]
Assumptions \eqref{rho_sigma_assump}--\eqref{interaction_assump} require that the minimal separation $\Delta_{\min}$ is large compared with both the sub-Gaussian noise level $\sigma$ and the perturbation magnitude $\epsilon$, as well as their interaction. Under these conditions, the contraction of Lloyd’s iterations is dominated by the signal rather than by the two noise sources. As $\rho_\sigma$ and $\rho_\epsilon$ increase, the constraints in \eqref{initial_conditions1}--\eqref{initial_conditions2} become less restrictive, meaning that coarser initializations are sufficient for convergence.
\end{rem}

\begin{rem}[Initialization requirements]
The initialization condition requires that the starting cluster centers are not excessively far from their corresponding population means, as quantified by $\Gamma_0$. For example, when $K=2$, the condition $\Gamma_0 < 1/2$ ensures that each initial center lies within a distance
$\Delta_{\min}/2$ of its true mean, so that it is closer to its own cluster than to the competing one. As the signal-to-noise ratios increase, the allowable deviation approaches this natural geometric limit.

The dependence on $\Delta_{\max}/\Delta_{\min}$ in~\eqref{initial_conditions1} is intrinsic to Lloyd’s mean-based updates. Even a small fraction $G_0$ of misassigned points originating from a distant cluster can induce a centroid bias of order $G_0\,\Delta_{\max}$, so controlling the normalized centroid error $\Gamma_0$ requires $G_0 \lesssim \Delta_{\min}/\Delta_{\max}$. Condition~\eqref{initial_conditions2} avoids this effect by directly constraining $\Gamma_0$.
\end{rem}

\begin{rem}[Necessity of good initialization]
Even under favorable signal-to-noise ratios, Lloyd’s algorithm may converge to a suboptimal local minimum if the initialization does not satisfy 
\eqref{initial_conditions1} or \eqref{initial_conditions2}.  
This phenomenon is already present in the noiseless setting and was formalized in \cite{lu2016statistical}.  
Figure~\ref{fig:initial_conditions} provides an example with $\sigma = 0$, $\epsilon = 1$, and 
$\Delta_{\min} = 2\sqrt{2}$, where an initialization violating these conditions causes Lloyd’s algorithm to converge to a poor local optimum and yield a high misclustering rate.  
Such examples underscore that the initialization assumptions in Theorem~\ref{thm-lloyd-guarantee} are not merely technical artifacts but are essentially necessary for correct recovery.
\begin{figure}[t]
    \centering
    \includegraphics[width =12cm, height =8cm]{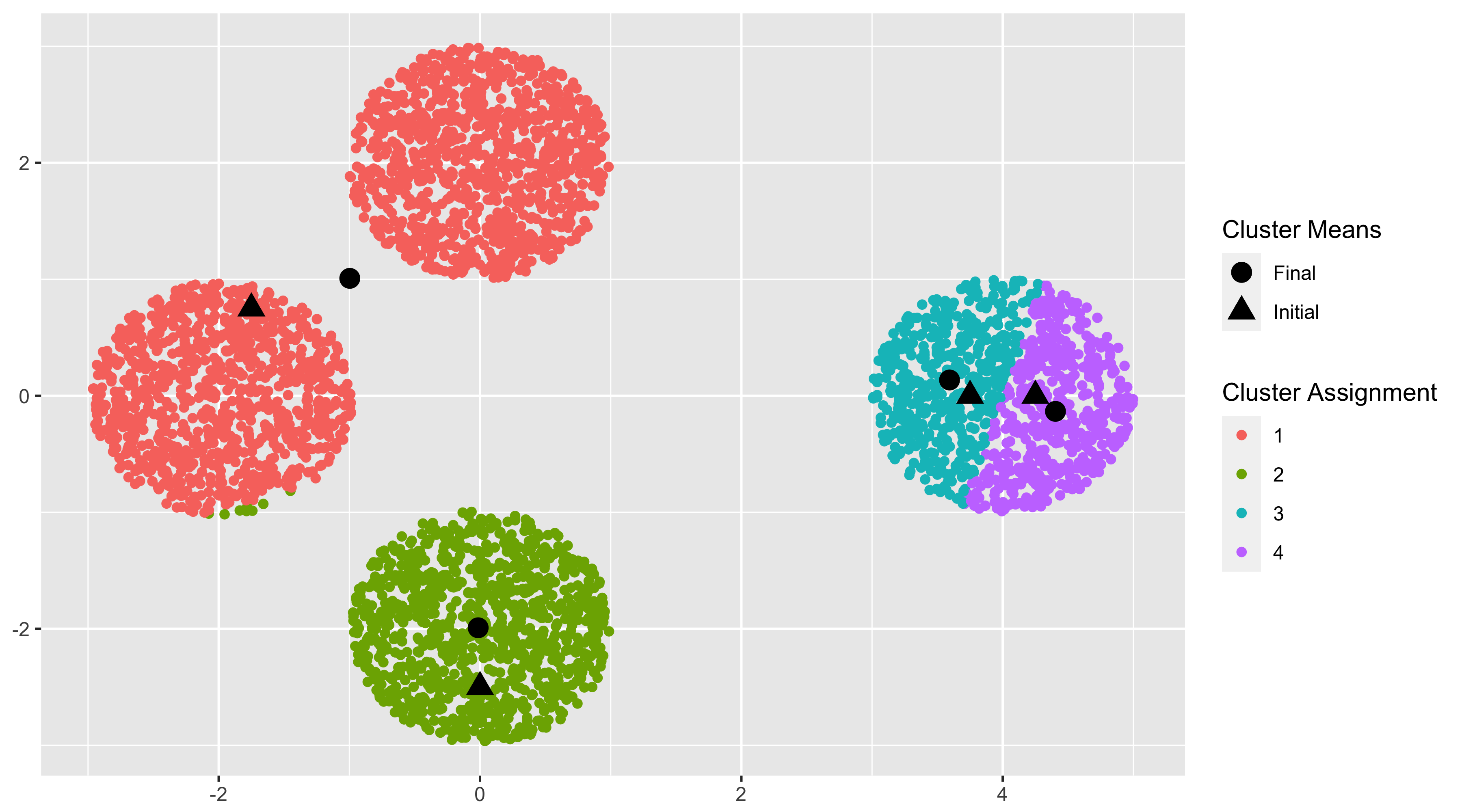}
    \caption{An example illustrating the need for the initial conditions given in \eqref{eq:indicator-rewritten}. Note that even with a clear separation of the four clusters, there exist candidate initial means which lead to local optima with a bad misclustering rate. }
    \label{fig:initial_conditions}
\end{figure}
\end{rem}

\begin{rem}[Condition on maximum–minimum gap]
Assumption~\eqref{aspect_assump} is a mild regularity condition on the cluster configuration. 
It rules out pathological cases where one or several cluster centers lie extremely far from the others 
compared with the minimal separation $\Delta_{\min}$. 
In such scenarios, any reasonable initialization already achieves perfect recovery for those 
far-away clusters, and their labels and estimated centers remain unchanged throughout Lloyd's iterations.  Consequently, these clusters do not interact with the remaining ones and do not play a role in the convergence analysis.  Imposing Assumption~\eqref{aspect_assump} therefore allows us to focus on the nontrivial regime where clusters are all within a comparable geometric scale.
\end{rem}

As an immediate consequence of Theorem~\ref{thm-lloyd-guarantee}, we derive the convergence rate for the estimated cluster centers under our perturbed mixture model. A related result was established in \cite[Theorem~6.2]{lu2016statistical}, but our bound explicitly accounts for the additional perturbation error.

\begin{corollary}\label{corr:lloyd-center}
Assume conditions \eqref{alpha_assump} -- \eqref{aspect_assump} hold for some sufficiently large universal constants. Then given any initializer satisfying
\eqref{initial_conditions1} or \eqref{initial_conditions2}, 
we have for $s \geq 4 \log n$, 
$$
\max_{k\in[K]} \|\widehat{\bmu}_k^{(s)} - \bmu_k\|
\leq  2\sqrt{3}(\sqrt{K}+1)\sigma\sqrt{\dfrac{(n+r)}{ n\alpha^2 }A_s} + 2\Delta_{\min}\frac{A_s}{\alpha} + 6\sigma\sqrt{\dfrac{r+\log n}{ n\alpha}} +  \epsilon ,
$$
where $A_s \leq \max \left\{ \exp \left( -\frac{\Delta_{\min}^2}{16 \sigma^2} \right) , \exp\left( -\frac{\Delta_{\min}^2}{8\epsilon\sigma} \right) \right\}$ and with probability \\
$1-\delta(n,\sigma,\Delta_{\min},\epsilon)$.
\end{corollary}

\subsection{$k$-means$++$ and good initialization}
Note that Theorem \ref{thm-lloyd-guarantee} hinges on good initialization \eqref{initial_conditions1} or \eqref{initial_conditions2}. The aim here is to show that one standard method of finding ``good'' initial seeds, the so-called $k$-means$++$ algorithm \citep{arthur:2007}, is able to provide such seeds with high probability, given enough separation between the clusters. To illustrate the main proof ideas and keep the exposition concise, we work in the following simplified setting, although similar proof techniques should apply more generally:
\begin{enumerate}[label=(\alph*), left=0pt]
    \item Assume $K=2$ clusters with equal sizes $n_1 = n_2$. Thus,
    $\Delta := \Delta_{\min} = \Delta_{\max} = \|\bmu_1 - \bmu_2\|$.
    \item Assume bounded noise perturbations such that $\|\be_i\| \le \epsilon$.
    \item Assume a Gaussian mixture model with $\cF_k \sim N_r(\bmu_k, \sigma^2\mvI_r)$,
    $k=1,2$, with a common scale parameter $\sigma^2 > 0$.
\end{enumerate}
In the $K=2$ setting, the $k$-means$++$ algorithm, applied to the dataset $\cX$ is as follows \citep{arthur:2007}:
\begin{enumerate}[label = (\roman*),  left=0pt]
    \item Choose the first center $\wh\bmu_1^{(0)}$ uniformly at random from $\cX$. 
    \item Choose the second center $\wh\bmu_2^{(0)}$ with probability:
    \[\bbP(\wh\bmu_2^{(0)} =\by|\cX, \wh\bmu_1^{(0)}) = \frac{||\by - \wh\bmu_1^{(0)}||^2}{\sum_{\by^\prime \in \cX} ||\by^\prime - \wh\bmu_1^{(0)}||^2 }, \qquad \by \in \cX. \]
\end{enumerate}
Next, define the constant 
\begin{equation}
    \label{eqn:gamma-r}
    \Psi_r:= \frac{\sqrt{2}\Gamma(\frac{r+1}{2})}{\Gamma(\frac{r}{2})}, 
\end{equation}
where $\Gamma(\cdot)$ denotes the usual Gamma function, and for given $\eta >0$, set 
\begin{align}
    \ell(A):=& 2\exp\left(-\frac{1}{2}\left(\left(\frac{1}{2} -\eta - \frac{\epsilon}{\Delta}\right)\frac{\Delta}{\sigma} - \Psi_r\right)^2\right) \\
    & + 2\exp\left(-\frac{1}{2}\left(A - \left(\Psi_r + \frac{\|\bmu_1\|}{\sigma} + \frac{\epsilon}{\sigma}\right)\right)^2\right) \label{def:l(A)} \\
    &+ \frac{8}{r}\exp\left(-\frac{(1-\eta)}{2}\left(\left(\frac{1}{2} -\eta - \frac{\epsilon}{\Delta}\right)\frac{\Delta}{\sigma} - \Psi_r\right)^2\right). \nonumber
\end{align}

\begin{theorem}
\label{thm:kmeans-init}
Let $\cD$ denote the event that the two initial seeds $\wh\bmu_1^{(0)}, \wh\bmu_2^{(0)}$ belong to different ground-truth clusters. Under the assumption $\Delta \geq C(\epsilon + \sqrt{r}\sigma)$ for some large $C$, we have  
\begin{equation}
\label{eqn:1040}
    \bbP(\cD|\cX) = 1 - O_p\left( \dfrac{r\sigma^2+\epsilon^2 }{\Delta^2} \right).
\end{equation} 
Conditional on $\cD$, without loss of generality, assume $\wh\bmu_k^{(0)}$ is in (true) cluster $k$ (otherwise permute the labeling).  Given any $\eta >0$, $\exists ~\tilde{C} >2\Psi_r$ such that if $\Delta/\sigma > \tilde{C}$, then for any fixed $A > ||\bmu_1||/\sigma + \Psi_r + \epsilon/\sigma$, we have
\begin{equation}
\label{eqn:conc-center}
\bbP\left(\left.\max_{k=1,2}\frac{||\wh\bmu_k^{(0)} - \bmu_k||}{\Delta} > \frac{1}{2} - \eta\right|\cD\right) \leq \ell(A) +o(1). 
\end{equation}
\end{theorem}
\begin{rem}
Note that the parameter $A$ is needed for a truncation argument in the proof and controls the rate of decay of the $o(1)$ term in \eqref{eqn:conc-center}. For this reason, it does not appear on the left-hand side of \eqref{eqn:conc-center}.   
\end{rem}

\begin{rem}
Theorem~\ref{thm:kmeans-init} is established under a simplified setting with $K=2$ balanced clusters and a bounded perturbation error. The result shows that, in order for $k$-means$++$ to yield a valid initialization, the perturbation level $\epsilon$ must be sufficiently small relative to the cluster separation $\Delta$. This requirement is consistent with the well-known sensitivity of $k$-means$++$ to outliers and large perturbations, and highlights that the theorem does not claim robustness to arbitrary noise.
\end{rem}

\section{Applications}\label{s:applications}
In this section, we illustrate how the perturbation model in Section~\ref{s:clustering_model} arises in several standard pipelines. In each case, the goal is to identify the population embedding, verify a uniform $2\to\infty$ bound for the structured perturbation term, and isolate a row-independent sub-Gaussian remainder. To keep the paper to a manageable length, we do not attempt to push these applications to their optimal theoretical regimes and leave such refinements for future work.

\subsection{Community detection in stochastic block models}\label{sec:SBM}
The stochastic block model (SBM) is a canonical model for networks with community structure. A central task is community detection, i.e., partitioning vertices into groups with higher within-group connectivity \citep{newman2004detecting,fortunato2010community,abbe2017community}. 
Many methods have been proposed, including spectral clustering and its variants \citep{rohe2011spectral,lei2015consistency,bordenave2015non,joseph2016impact}, likelihood-based procedures \citep{bickel2009nonparametric,amini2013pseudo,gao2017achieving}, belief propagation \citep{mossel2015reconstruction,mossel2018proof}, and convex relaxations \citep{hajek2016achieving}. 
Since standard spectral clustering ends with $k$-means on a spectral embedding, we analyze Lloyd’s algorithm applied to the basic adjacency spectral embedding given by the top-$K$ eigenvectors of $\Ab$.

  
Consider an SBM with a fixed number of communities $K$. The probability matrix $\Ab^*$ can be represented as $\Ab^*=\Zb \Bb \Zb^{\top}$, where $\Zb \in \mathbb{R}^{n \times K}$ denotes the membership matrix and $\Bb \in \bbR^{K\times K}$ is the low-rank connection probability matrix. The observed graph is represented using an adjacency matrix $\Ab$ with $\Ab_{ij} = \Ab_{ji}\sim \bern(\Ab_{ij}^*)$ for all $i\neq j$ and $\Ab_{ii} = 0$ for $i\in [n]$. Denote the error term as
$$
\Eb = \Ab - \Ab^{*}.
$$
Let $\Ab=\sum_{i=1}^n \lambda_i \bu_i \bu_i^\top$ and 
$\Ab^*=\sum_{i=1}^K \lambda_i^* \bu_i^*(\bu_i^*)^\top$ denote the eigendecompositions of $\Ab$ and $\Ab^*$, respectively. The goal is to analyze the performance of Lloyd's algorithm on the $n\times K$ spectral embedding matrix $\Ub= (\bu_1, \bu_2,\ldots,\bu_K)$. Denote $\Ub^* = (\bu_1^*, \bu_2^*,\ldots,\bu_K^*)$ and $\mvLambda^* = \text{diag}(\lambda_1^*,\lambda_2^*,\ldots, \lambda_K^*)$. The embedding matrix $\Ub$ can be decomposed in a fashion similar to the general model \eqref{eqn:model_matrix} up to some orthogonal matrix:
\begin{equation}
\label{eqn:model_SBM}
\Ub\Ob = \Ub^* + \left[\Ub\Ob-\Ab\Ub^*(\mvLambda^{*})^{-1}\right] + \Eb \Ub^*(\mvLambda^{*})^{-1}=:\Ub^*+\Eb_1+\Eb_2,
\end{equation}
where $\Ob\in \bbR^{K\times K}$ is an orthogonal matrix, $\Ub^*$ is the population embedding (it has exactly $K$ distinct rows, one per community), $\|\Eb_1\|_{2\rightarrow\infty}$ can be controlled using the main theorem in \cite{abbe2020entrywise}, and $\Eb_2$ is the row-independent sub-Gaussian term. To apply Theorem \ref{thm-lloyd-guarantee}, we make the following assumptions on SBM.

\begin{asm}
\label{asm:SBM}
\begin{enumerate}
    \item [(a)]
    There exist strictly positive constants $c_1$ and $C_1$ such that
    $$
    0 < c_1 \leq \liminf _n \inf _k \frac{n_k}{n} \leq \limsup _n \sup _k \frac{n_k}{n} \leq C_1 < 1.
    $$
    \item [(b)]
   The number of communities $K$ is held fixed and the community-wise probability connection matrix
$
\Bb=\Bb_n:= \rho_n \Bb_0,
$
for a sparsity rate  $\rho_n <1$ and a fixed matrix $\Bb_0$ with full rank $K$.
    \item [(c)]
    Assume the sparse regime such that
    $$\rho_n \geq \frac{c}{n},$$
    for some large $c$.
\end{enumerate}
    \end{asm}

In sparse SBM where $\rho_n = o(1)$, we can attain improved concentration bounds for Bernoulli entries that go beyond the scope of sub-Gaussian properties. By following a proof similar to that of Theorem \ref{thm-lloyd-guarantee}, we can establish the following result regarding the performance of Lloyd's algorithm in the context of SBM.

\begin{theorem}
\label{thm:sbm_K}
    Assume the conditions in Assumption \ref{asm:SBM} hold and large enough sample size $n\geq 256 \log n$. Then, for any initializer that satisfies 
\begin{equation}
    \Gamma_0 < \frac{1}{2} - \frac{c_2}{\sqrt{n\rho_n}} - \frac{c_2}{(n\rho_n)^{1/4}},\label{initializer_SBM}
\end{equation}
for some positive constant $c_2$, 
we have 
$$
A_s \leq \exp \left( - Cn\rho_n\right), \quad \text{for all} \quad s\geq 4\log n,
$$
for some constant $C$ with probability $1- \frac{1}{n} - 4\exp \left( -\sqrt{n\rho_n}\right)$. 
\end{theorem}

For the SBM with fixed $K$, it is known that the information-theoretic threshold for exact recovery is $\rho_n \asymp \frac{\log n}{n}$, in the sense that if $\rho_n / (\log n / n) \to 0$, then no algorithm can achieve exact recovery \citep{abbe2017community, lei2019unified}. Here, exact recovery means $A_s = 0$ with probability $1-o(1)$. Standard spectral clustering based on the adjacency matrix is known to be suboptimal in sparse regimes \citep{abbe2017community}, compared with approaches based on non-backtracking operators \citep{krzakala2013spectral,bordenave2015non} or trimming procedures \citep{yun2014accurate}.  

Theorem~\ref{thm:sbm_K} establishes that, for any initializer satisfying \eqref{initializer_SBM}, Lloyd’s algorithm applied to the adjacency spectral embedding achieves an exponentially decaying misclustering rate,
$A_s \le \exp(-C n\rho_n)$, after $O(\log n)$ iterations. In particular, when $\rho_n \ge c \log n / n$ for a sufficiently large constant $c$, this bound implies $A_s = 0$ with probability $1-o(1)$, yielding exact recovery. Consequently, Lloyd’s algorithm achieves the information-theoretic limit for community detection in the SBM with fixed $K$ when initialized appropriately.

\begin{rem}[Initialization condition]
The initialization requirement is stated solely in terms of the normalized centroid error $\Gamma_0$. When the signal-to-noise ratio is large (e.g., $n\rho_n \to \infty$
in the SBM setting), this condition reduces to $\Gamma_0 < \tfrac12 - \epsilon_0$ for some
small $\epsilon_0 > 0$, which ensures that each initial center lies closer to its own
population mean than to any competing cluster. This is a standard local convergence
condition for Lloyd’s algorithm and does not depend on the maximum separation between
clusters.

Such an initialization can be achieved by several existing procedures for example those analyzed in \cite{jana2025provable}.
\end{rem}

\subsection{Community detection in noisy stochastic block models}\label{sec:noisy-SBM}

Although SBM is a powerful tool for analyzing networks with community structure, it does not incorporate potential measurement errors that are prevalent in nearly all network analysis applications. Measurement error refers to inaccuracies or uncertainties in the observed network data, such as missing or noisy edges. Therefore, when modeling networks using SBM or applying community detection methods, it is crucial to account for the impact of measurement error to ensure robust and reliable inference of network communities \citep{priebe2015statistical, tabouy2020variational}.

Consider a noisy version of SBM described in Section~\ref{sec:SBM}. Assume that the adjacency matrix $\Ab \in \bbR^{n \times n}$ is generated from an SBM, with $\Ab_{ij} = \Ab_{ji} \sim \operatorname{Bernoulli}(\Ab_{ij}^*)$ and $\Ab^* = \Zb \Bb \Zb^{\top}$. What we observe is a noisy graph with adjacency matrix $\Yb \in \bbR^{n \times n}$, as described in \cite{chang2022estimation}:
\[
\mathbb{P}\!\left(\Yb_{ij} = 1 \mid \Ab_{ij} = 0\right) = \alpha_n
\quad \text{and} \quad
\mathbb{P}\!\left(\Yb_{ij} = 0 \mid \Ab_{ij} = 1\right) = \beta_n .
\]
Decompose $\Yb$ as $\Yb = \sum_{i=1}^n \lambda_i \bu_i \bu_i^{\top}$. Denote $\Ub = (\bu_1, \bu_2, \ldots, \bu_K)$. We are interested in the misclustering rate of Lloyd’s algorithm applied to the spectral embedding matrix $\Ub$.

\begin{asm}
\label{asm:noisy-SBM}
We assume that conditions (a) and (b) of Assumption~\ref{asm:SBM} hold, and that the noise level satisfies
\[
(1-\alpha_n-\beta_n)\rho_n \geq \frac{c}{n},
\]
and
\[
(1-\alpha_n-\beta_n)\rho_n \geq c \alpha_n,
\]
for some sufficiently large constant $c$.
\end{asm}

Similar to Section~\ref{sec:SBM}, we can control the misclustering rate of Lloyd’s algorithm under the noisy SBM as follows.

\begin{theorem}
\label{thm:noisy-sbm}
Under the noisy SBM, assume that the conditions in Assumption~\ref{asm:noisy-SBM} hold and that the sample size satisfies $n \geq 256 \log n$. Then, for any initializer that satisfies
\begin{equation}
\Gamma_0 < \frac{1}{2} - \frac{c_2}{\sqrt{n\rho_n}} - \frac{c_2}{(n\rho_n)^{1/4}},
\label{initializer_noisy_SBM}
\end{equation}
for some positive constant $c_2$, we have
\begin{equation}
A_s \leq \exp\!\left( - C (1-\alpha_n-\beta_n) n\rho_n \right)
\quad \text{for all} \quad s \geq 4 \log n,
\end{equation}
for some constant $C$, with probability
\[
1 - \frac{1}{n} - 4 \exp\!\left( -\sqrt{(1-\alpha_n-\beta_n)n\rho_n} \right).
\]
\end{theorem}

Compared with Theorem~\ref{thm:sbm_K}, the additional edge-noise reduces the effective signal strength by a factor $(1-\alpha_n-\beta_n)$. Indeed, $\bbE[\Yb\mid \Ab^*]=(1-\alpha_n-\beta_n)\Ab^* + \alpha_n(\mathbf{1}\mathbf{1}^\top-\mathbf{I})$, so the informative low-rank part is scaled by $(1-\alpha_n-\beta_n)$. Under Assumption~\ref{asm:noisy-SBM}, this signal dominates the noise contributions, and the proof follows the same strategy as in Section~\ref{sec:SBM}, with $n\rho_n$ replaced by $(1-\alpha_n-\beta_n)n\rho_n$.

\subsection{Spectral clustering in mixture models }\label{sec:SC}
Sub-Gaussian mixture models (SGMMs) provide a flexible framework for clustering data with light-tailed noise and include, as special cases, Gaussian mixture models and mixtures with bounded support. Spectral methods for SGMMs have been studied extensively; see, for example, \cite{loffler2021optimality,davis2021clustering,ndaoud2022sharp,zhang2022leave,abbe2022ellp}. In this subsection, we show how our perturbation-based analysis applies to spectral clustering for SGMMs when Lloyd’s algorithm is used in the final clustering step.

Consider the sub-Gaussian mixture model: 
$$
\bx_i = \bar{\bx}_i + \bw_i = \bmu_{z_i} + \bw_i \in \bbR^p  \text{,  for }i \in [n].
$$
Here, $\left\{\bmu_l\right\}_{l=1}^K \subseteq \bbR^{p}$ are cluster centers, $\left\{z_i\right\}_{i=1}^n \subseteq[K]^n$ are true labels, and $\bw_i$ are i.i.d. $ \operatorname{subG}(\sigma^2)$ with mean zero. The matrix form is $\Xb = \bar{\Xb} + \Wb$ with $\bar{\Xb} = \Zb\Mb$, where $\Zb\in \bbR^{n\times K}$ is the membership matrix and $\Mb\in\bbR^{K\times p}$ is the low-rank cluster center matrix.  

We focus on the spectral clustering method in \cite{abbe2022ellp} where Lloyd's algorithm is used in the last step for clustering.
We analyze the spectral embedding based on the hollowed Gram matrix, following \cite{abbe2022ellp}, and show that it admits a decomposition compatible with our general perturbation framework.

Define the hollowed Gram matrix $\Gb \in \bbR^{n \times n}$ of samples $\left\{\bx_i\right\}_{i=1}^n$ through $\Gb_{i j}=$ $\left\langle\bx_i, \bx_j\right\rangle \mathbf{1}_{\{i \neq j\}}$, and the Gram matrix $\widebar{\Gb} \in \bbR^{n \times n}$ of signals $\left\{\bar{\bx}_i\right\}_{i=1}^n$ through $\widebar{\Gb}_{i j}=\left\langle\bar{\bx}_i, \bar{\bx}_j\right\rangle$. 
Let $\Gb = \sum_{i=1}^{n}\lambda_i\bu_i\bu_i^\top$ and $\widebar{\Gb} = \sum_{i=1}^{K}\bar{\lambda}_i\bar{\bu}_i\bar{\bu}_i^\top$ be the eigen-decomposition of matrices $\Gb$ and $\bar{\Gb}$, respectively, with $\lambda_1 \geqslant \lambda_2 \geqslant \ldots \geqslant \lambda_n$ and $\bar{\lambda}_1 \geq \bar\lambda_2 \geqslant \ldots \geq \bar\lambda_K$. Define $\Ub = (\bu_1,\ldots,\bu_{K})$ and $\bLambda = \operatorname{diag}\left(\lambda_1,\ldots,\lambda_{K}\right)$. Similarly, define $\bar{\Ub} = (\bar{\bu}_1,\ldots,\bar{\bu}_{K})$ and $\bar{\bLambda} = \operatorname{diag}\left(\bar{\lambda}_1,\ldots,\bar{\lambda}_{K}\right)$. Define $\cH(\cdot)$ as the hollowing operator, zeroing out all diagonal entries of a square matrix. We are interested in the misclustering rate when applying Lloyd's algorithm on the spectral embedding matrix $\Ub\bLambda^{1/2}$. To apply the main theorem, we can show that $\Ub\bLambda^{1/2}$ has the decomposition form as in the general framework \eqref{eqn:model_matrix} 
\begin{align*}
\Ub\bLambda^{1/2}\Ob &= \widebar{\Ub}\widebar{\bLambda}^{1/2} + \cH\left(\Wb \widebar{\Xb}^{\top}\right) \widebar{\Ub}\widebar{\bLambda}^{-1/2} + \cH\left(\Wb \Wb^{\top}\right) \widebar{\Ub}\widebar{\bLambda}^{-1/2} + \Eb_{12}\\
&=: \widebar{\Ub}\widebar{\bLambda}^{1/2} + \Eb_2 + \Eb_{11} + \Eb_{12},
\end{align*}
where $\Ob$ is an orthogonal matrix, $\widebar{\Ub}\widebar{\bLambda}^{1/2}$ represents the population embedding, 
$\Eb_2$ is a row-independent sub-Gaussian term, and $\Eb_{11}$ and $\Eb_{12}$ are perturbation terms that are small with high probability.

Denote $\bar{s} = \min _{l \neq k}\left\|\bmu_l-\bmu_k\right\|$. We make the following assumptions on the sub-Gaussian mixture model.
\begin{asm} \label{asm:SC}
\begin{enumerate}
    \item [(a)] (Regularities)
    Let $\Mb\Mb^\top \in \mathbb{R}^{K \times K}$ be the Gram matrix of $\left\{\boldsymbol{\mu}_l\right\}_{l=1}^K$. Suppose that $\operatorname{rank}(\Mb\Mb^\top)=K$ is fixed and there is a constant $\kappa_0$ that bounds
    $$
    \frac{n}{\min _{k \in[K]}\left|\left\{i \in[n]: z_i=k\right\}\right|}, \quad \frac{\lambda_1(\Mb\Mb^\top)}{\lambda_K(\Mb\Mb^\top)} \quad \text { and } \quad \frac{\max _{l \in[K]}\left\|\bmu_l\right\|}{\min _{l \neq k}\left\|\bmu_l-\bmu_k\right\|}
    $$
    from above. 
    \item [(b)]
Assume 
$
\min\left\{\dfrac{\sqrt{n}\bar{s}^2}{\sqrt{\log n(p+\log n)}\sigma^2}, \dfrac{\bar{s}^2}{\sigma^2} \right\} \rightarrow \infty,
$
as $n\rightarrow \infty$.
   \item [(c)]
   Assume 
$
\dfrac{\bar{s}^2}{\sigma^2} \gtrsim \log n. 
$
    
\end{enumerate}
\end{asm}

When applying Lloyd's algorithm on the spectral embedding matrix $\Ub\bLambda^{1/2}$, we have the following theorem. 
\begin{theorem}
\label{thm:spect-clust}
Assume the conditions in Assumption \ref{asm:SC} hold. Then, for any initializer that satisfies 
$
\Gamma_0 \leq \frac{1}{2}-\epsilon_0
$
for some small $\epsilon_0$, we have \\
\scalebox{0.95}{$
A_s \leq \max \left \{ \exp \left( - \frac{\bar{s}^2}{32\kappa_0^4\beta \sigma^2} \right) , 
\exp\left( -C\frac{\bar{s}^{3}\sqrt{n}}{\sigma^3\sqrt{\log n(p+\log n)}} \right), \exp\left( -C\frac{\sqrt{\log n}\bar{s}}{\sigma} \right)
\right \}, $}\\
where $\beta = \dfrac{\max_l n_l}{n}$ for some constant $C$ and all $s\geq 4\log n$ 
with probability \\
\scalebox{0.94}{$1- \exp \left( - \frac{\bar{s}}{ \sigma}  \right) -  \exp\left( - \left(\frac{\bar{s}}{\sigma} \right)^{3/2} \left(\frac{n}{\log n(p+\log n)} \right)^{1/4} \right) - \exp\left( -(\log n)^{1/4}\sqrt{\frac{\bar{s}}{ \sigma}} \right) - o(1).$}
\end{theorem}

Assumption~\ref{asm:SC} ensures that the main theorem of \cite{abbe2022ellp} applies, which yields
\[
\bu_j = \Gb \bu_j / \lambda_j \approx \Gb \widebar{\bu}_j / \bar{\lambda}_j,
\]
for $j\in[K]$. Therefore, our focus is on the analysis of $\Gb \bar{\bu}_j / \bar{\lambda}_j$, which is a linear combination of elements in $\Gb$. 

We compare Theorem~\ref{thm:spect-clust} with optimal misclustering rates for spectral clustering under SGMMs, such as those in \cite{loffler2021optimality}. For simplicity, consider the balanced two-component case with $K=2$, $n_1=n_2=n/2$, and $p\asymp n$.
Additionally, we assume $\frac{\bar{s}^2}{\sigma^2} =O(\log n)$, otherwise we will have exact recovery $A_s = 0$ with probability $1-o(1)$. Our result can be simplified as 
$$
A_s \leq \exp \left( - \frac{C\bar{s}^2}{\sigma^2} \right) 
$$
for some constant $C$ with probability $1- 3\exp \left( - \frac{\bar{s}}{\sigma} \right) - o(1)$. Under the same setting above, \cite{loffler2021optimality} show that 
$$
A_s \leq \exp \left( - (1-o(1))\frac{\bar{s}^2}{8\sigma^2} \right)
$$
under Gaussian mixture models with probability $1- \exp \left( - \frac{\bar{s}}{\sigma} \right) - o(1)$
and 
$$
A_s \leq \frac{C^{\prime } k \sigma^2\left(1+\frac{p}{n}\right)}{\bar{s}^2}
$$
for some constant $C^{\prime}$ under sub-Gaussian mixture models with probability $1-\exp(-0.08 n)$.

Under the Gaussian mixture setting, our result and \cite{loffler2021optimality} have the same exponential misclustering error format but \cite{loffler2021optimality} have the optimal coefficient in the exponent. Under the sub-Gaussian mixture setting, our result shows an exponential misclustering rate while \cite{loffler2021optimality} have only a polynomial rate. 

\subsection{Mixture models with multidimensional scaling}\label{sec:MDS}

Multidimensional scaling (MDS) is a classical and widely used technique for embedding data into a low-dimensional Euclidean space using only pairwise dissimilarities \citep{young1938discussion,torgerson1952multidimensional,gower1966some}. 
Classical multidimensional scaling (CMDS) preserves Euclidean distances and can be viewed as a form of spectral embedding applied to a centered Gram matrix \citep{borg2005modern}. 
Despite its broad use in clustering and visualization, particularly in high-dimensional applications such as single-cell data analysis \citep{chen2019single}, theoretical guarantees for clustering based on MDS embeddings remain relatively limited.

In this subsection, we analyze the behavior of Lloyd’s algorithm applied to MDS embeddings under low-rank sub-Gaussian mixture models. 
We show that the CMDS embedding admits a decomposition into a low-rank signal component and structured, dependent perturbations that fit naturally into our general perturbation framework, thereby enabling misclustering guarantees for Lloyd’s algorithm.

Assume that the original dataset $\Xb$ is generated from the same sub-Gaussian mixture model as in Section~\ref{sec:SC}, and that we observe a dissimilarity matrix $\mathbf{D} \in \mathbb{R}^{n\times n}$ with entries $d_{ij}$ measuring pairwise distances between samples. 
Multidimensional scaling seeks a low-dimensional embedding $\Yb \in \bbR^{n\times r}$ whose interpoint distances $\|\by_i-\by_j\|_2$ approximate $d_{ij}$. 
In the classical setting (CMDS), this leads to a spectral embedding based on a centered Gram matrix, which is the focus of our analysis.

Mathematically, the MDS matrix can be found by minimizing the total error of representation. For simplicity, consider CMDS, where the Euclidean distance is used to measure the distance in both spaces. CMDS is equivalent to PCA applied to the centered data matrix. However, MDS is much more general than standard PCA and can also perform nonlinear dimension reduction.

The procedure for finding the CMDS matrix $\widetilde{\Yb}$ is as follows. Denote $\mathbf{B} = -\frac{1}{2} \mathbf{J} \mathbf{D}^{(2)} \mathbf{J}$, where $\mathbf{D}_{ij}^{(2)}=d_{ij}^2, \mathbf{J}=\mathbf{I}_{n}-\mathbf{1}\mathbf{1}^\top / n$ is the centering matrix, and $\mathbf{1} \in \mathbb{R}^{n}$ is a column vector of ones. It can be shown \citep{borg2005modern} that 
\begin{equation}
	\Bb = (\Xb - \mathbf{1}\widehat{\bmu}_{\bx}^\top)(\mathbf{X} - \mathbf{1}\widehat{\bmu}_{\bx}^\top)^\top = (\Jb\Xb)(\Jb\Xb)^\top,
\end{equation}
where $\widehat{\bmu}_{\bx}  = \sum_{i = 1}^n \bx_i / n$ is the empirical mean. Consider the eigendecomposition $\Bb = \widetilde{\Pb} \widetilde{\bLambda} \widetilde{\Pb}^\top$.
In CMDS, the solution $\widetilde{\Yb}$ can be represented as $\widetilde{\Yb}=\widetilde{\Pb}_r \widetilde{\bLambda}_r^{1 / 2}$, where $\widetilde{\Pb}_r$ and $ \widetilde{\bLambda}_r$ are the first $r$ eigenvectors and eigenvalues of $\Bb$.

For technical convenience, we consider the CMDS embedding matrix with one additional hollowing operator. This modification removes diagonal self-interaction terms and simplifies entrywise concentration bounds without affecting the clustering geometry. Let $\cH(\Bb) = \cH((\Jb\Xb)(\Jb\Xb)^\top) = \sum_{i=1}^{n}\lambda_i\bu_i\bu_i^{\top}$ and $(\Jb\bar{\Xb})(\Jb\bar{\Xb})^\top = \sum_{i=1}^{K-1}\bar{\lambda}_i\bar{\bu}_i\bar{\bu}_i^\top$ be the SVDs with $\lambda_1 \geqslant \lambda_2 \geqslant \ldots \geqslant \lambda_n$ and $\bar{\lambda}_1 \geqslant \bar\lambda_2 \geqslant \ldots \geqslant \bar\lambda_{K-1}$. Define $\Ub = (\bu_1,\ldots,\bu_{K-1})$ and \\$\bLambda = \operatorname{diag}\left(\lambda_1,\ldots,\lambda_{K-1}\right)$. Similarly, define $\bar{\Ub} = (\bar{\bu}_1,\ldots,\bar{\bu}_{K-1})$ and \\$\bar{\bLambda} = \operatorname{diag}\left(\bar{\lambda}_1,\ldots,\bar{\lambda}_{K-1}\right)$. The MDS embedding matrix we consider in this section is $\Yb = \Ub\bLambda^{1/2}$. The goal is to study the clustering accuracy when applying Lloyd's algorithm to the MDS embedding matrix $\Yb$, which can be decomposed as 
$$
\Ub\bLambda^{1/2}\Ob = \widebar{\Ub}\widebar{\bLambda}^{1/2} + \cH\left((\Jb\Wb) (\Jb\Xb)^\top\right) \widebar{\Ub}\widebar{\bLambda}^{-1/2} + \Eb_{13},
$$
where $\Ob$ is an orthogonal matrix and $\cH\left((\Jb\Wb) (\Jb\Xb)^\top\right) \widebar{\Ub}\widebar{\bLambda}^{-1/2}$ can be further decomposed as 
\begin{align*}
  & \cH\left((\Jb\Wb) (\Jb\Xb)^\top\right) \widebar{\Ub}\widebar{\bLambda}^{-1/2} \\
  = & \cH\left(\Wb (\Jb\widebar\Xb)^\top\right) \widebar{\Ub}\widebar{\bLambda}^{-1/2} + \cH\left((\Jb-\Ib_n)\Wb(\Jb\widebar\Xb)^\top\right) \widebar{\Ub}\widebar{\bLambda}^{-1/2} \\
  & + \cH\left((\Jb\Wb) (\Jb\Wb)^\top\right) \widebar{\Ub}\widebar{\bLambda}^{-1/2} \\
  =&:  \Eb_{2} + \Eb_{11} + \Eb_{12}. 
\end{align*}
In the above decomposition, $\widebar{\Ub}\widebar{\bLambda}^{1/2}$ represents the population embedding, $\Eb_2$ has independent sub-Gaussian rows, $\Eb_{13}$ is a higher-order residual term, and each $\Eb_{1l}$ can be controlled individually for $l\in[3]$.

Define $\bar{s} = \min _{l \neq k}\left\|\bmu_l-\bmu_k\right\|$.
We make the following assumption on the sub-Gaussian mixture model.
\begin{asm}  
\label{asm:MDS}
\begin{enumerate} 
\item [(a)]
Assume $\dfrac{n_l}{n} \rightarrow \pi_l>0$ as $n\rightarrow \infty$ for $l\in[K]$. Define $\bm{\pi} = (\pi_1,\ldots, \pi_K)^\top$, $\widebar\bmu = \sum_{l=1}^K \pi_l \bmu_l$ and $\mathbf{S} = \Ib_K - \bone_K\bm{\pi}^\top$. 
\item [(b)] (Regularities)
Let $\mathbf{S}\Mb\Mb^\top\mathbf{S}^\top \in \bbR^{K \times K}$ be the Gram matrix of $\left\{\bmu_l - \widebar\bmu \right\}_{l=1}^K$. Suppose that $\operatorname{rank}(\mathbf{S}\Mb\Mb^\top\mathbf{S}^\top)=K-1$ is fixed and there is a constant $\kappa_0$ that bounds
$$\frac{\lambda_1(\mathbf{S}\Mb\Mb^\top\mathbf{S}^\top)}{\lambda_{K-1}(\mathbf{S}\Mb\Mb^\top\mathbf{S}^\top)} \quad \text { and } \quad \frac{\max _{l \in[K]}\left\|\bmu_l\right\|}{\min _{l \neq k}\left\|\bmu_l-\bmu_k\right\|}
$$
from above. 
\item [(c)]
Assume 
$
\min\left\{\dfrac{n\bar{s}^2}{p\sigma^2}, \dfrac{\sqrt{n}\bar{s}^2}{\sqrt{\log n(p+\log n)}\sigma^2} \right\} \rightarrow \infty,
$ as $n\rightarrow \infty$.
\item [(d)]
Assume 
$
\dfrac{\bar{s}^2}{\sigma^2} \gtrsim \log n. 
$
\end{enumerate}
\end{asm}

When Lloyd's algorithm is applied to MDS embedding $\Yb$, the misclustering rate can be controlled as follows. 

\begin{theorem}
\label{thm:mds}
Assume the conditions in Assumption \ref{asm:MDS} hold. Then, for any initializer that satisfies 
$
\Gamma_0 \leq \frac{1}{2}-\epsilon_0
$ for some small $\epsilon_0$, we have 
\begin{align*}
A_s \leq \max & \left\{ \exp \left( - \frac{\bar{s}^2}{32\kappa_0^4\beta \sigma^2} \right) , 
\exp\left( -C\frac{\bar{s}^{3}\sqrt{n}}{\sigma^3\sqrt{\log n(p+\log n) }} \right),\right. \\
& \left. \exp\left( -C\frac{n \bar{s}^{3}}{p\sigma^3} \right), \exp\left( -C\frac{\sqrt{\log n}\bar{s}}{\sigma} \right)
\right \}  ,    
\end{align*}
where $\beta = \dfrac{\max_l n_l}{n}$ for some constant $C$ for all $s\geq 4\log n$ and 
with probability 
\begin{align*}
1- & \exp \left( - \frac{\bar{s}}{ \sigma}  \right) -  \exp\left( - \left[\left(\frac{n}{\log n(p+\log n)} \right)^{1/4}+\sqrt{\frac{n}{p}} \right] \left(\frac{\bar{s}}{\sigma} \right)^{3/2}  \right) \\
-& \exp\left( -(\log n)^{1/4}\sqrt{\frac{\bar{s}}{ \sigma}} \right) - o(1).
\end{align*}
\end{theorem}

The above theorem in the MDS setting is similar to Theorem~\ref{thm:spect-clust}, except that we have one additional centering operation on the data. The conditions are slightly stronger than those in Assumption~\ref{asm:SC} in the high-dimensional setting where $p\gtrsim n$, since the mean vector $\widebar\bw$ has dimension $p$ and its influence accumulates as $p$ increases. \cite{little2022analysis}, the first strong theoretical guarantee for CMDS in the literature, study the performance of CMDS on the task of clustering under sub-Gaussian mixture models. 
Their result focuses on sufficient conditions for the exact recovery of cluster labels. Specifically, \cite{little2022analysis} show that when 
$$\frac{\bar{s}^2}{\sigma^2} \gtrsim \max\left\{\dfrac{p}{n} , \sqrt{p}\log n\right\},$$ 
we can achieve exact recovery of cluster labels (i.e., $A_s = 0$) with probability $1-o(1)$ by applying the $k$-means clustering algorithm on MDS embedding. 

To compare our result with \cite{little2022analysis}, we consider the task of exact recovery for cluster labels. Our result, implied by Theorem \ref{thm:mds}, only requires that 
$$
\frac{\bar{s}^2}{\sigma^2} \gtrsim \max\left\{ \left( \frac{p\log n}{n}\right)^{2/3}, \sqrt{\frac{p + \log n}{n}} \log n,\log n\right\},
$$ 
for exact recovery, which is an improvement over their result. 




\subsection{Random dot product graphs}
Although the SBM is a widely used graph model, assuming equal connection probabilities between nodes within the same community is often unrealistic. In many real-world applications, edges form between nodes based on the similarity of their attributes. Latent position random graphs \citep{hoff:2002} address this issue by allowing connection probabilities to depend on latent attributes, which encode node characteristics that directly or indirectly influence community assignments. When the probability of an edge is determined by the dot product of these latent attributes, the graph is called a Random Dot Product Graph (RDPG) \citep{young:2007}. RDPGs simplify the general latent position model by restricting the latent space to vectors whose dot product yields a valid probability, and this structure facilitates analysis using techniques from random matrix theory. Moreover, RDPGs are broadly applicable: they can approximate latent position graphs \citep{tang2013}, and any SBM can be represented as an RDPG with constant latent attributes for nodes within the same community.

We show below that spectral clustering applied to the adjacency matrix of an RDPG, whose latent attributes follow a sub-Gaussian $K$-mixture distribution, fits into our general clustering framework \eqref{eqn:model_matrix}. The cluster memberships, cluster centers, and sub-Gaussian noise matrices arise directly from the mixture structure of the latent attributes. To control the perturbation error, we use the result of \cite{lyzinski:2014}, which provides an upper bound on the $2\to\infty$ norm of the spectral embedding error.

Let $n$ be the number of nodes and suppose $r$ is the dimension of the latent variables. Assume $\cF$ is a distribution on $\bbR^r$ such that for all $\by^*_1,\by^*_2$ in the support of $\cF$, we have $(\by^*_1)^{\top}\by^*_2 \in [0,1]$. Let $\Yb^* = (\by^*_1,\ldots,\by^*_n)^{\top} \in \bbR^{n\times r}$ where $\{\by^*_i\}_{i\in[n]} \sim \cF$ independently. An adjacency matrix $\Ab$ is said to be an RDPG with latent attributes $\Yb^*$, denoted $\Ab\sim \mathrm{RDPG}(\Yb^*)$, if 
\begin{equation}
    \pr (\Ab | \Yb^*) = \prod_{i<j;i,j\in[n]} ((\by^*_i)^{\top}\by^*_j)^{A_{ij}}(1-(\by^*_i)^{\top}\by^*_j)^{1-A_{ij}},
\end{equation}
where $A_{ij}$ is the $(i,j)$-th entry of $\Ab$. Note that if $\Yb^*$ has $K$ distinct rows, then $\Ab$ would be an adjacency matrix observed from an SBM with $K$ communities. Many variations of the SBM can also be written as an RDPG by imposing structure on $\Yb^*$.   

Define the expected adjacency matrix as $
\Pb := \mathbb{E}\,\Ab = \Yb^*(\Yb^*)^{\top}.$
Let $\Pb = \Ub \Sbb \Ub^{\top}$ be its eigendecomposition, where 
$\Ub \in \mathbb{R}^{n\times n}$ is orthogonal and 
$\Sbb \in \mathbb{R}^{n\times n}$ is diagonal with eigenvalues in decreasing order.
Let $\Sbb_r \in \mathbb{R}^{r\times r}$ denote the diagonal matrix containing the $r$ largest 
eigenvalues of $\Pb$, and let $\Ub_r \in \mathbb{R}^{n\times r}$ be the matrix of the 
corresponding eigenvectors. 
The matrix $\Ub_r \Sbb_r^{1/2}$ estimates $\Yb^*$ only up to an orthogonal rotation 
$\Rb \in \mathbb{R}^{r\times r}$, since for any such $\Rb$, $
\Pb = \Yb^*(\Yb^*)^{\top} = \Yb^* \Rb \Rb^{\top} (\Yb^*)^{\top}.$
Following \cite{lyzinski:2014}, we impose the identification $\Yb^* = \Ub_r \Sbb_r^{1/2}$.

To estimate $\Yb^*$, we use the eigendecomposition of the observed adjacency matrix. 
Let $\Ab = \widehat{\Ub}\widehat{\Sbb}\widehat{\Ub}^{\top}$ be the eigendecomposition of $\Ab$, 
and define $\widehat{\Sbb}_r$ and $\widehat{\Ub}_r$ analogously to $\Sbb_r$ and $\Ub_r$. 
Then the spectral estimate of $\Yb^*$ is $
\widehat{\Yb} = \widehat{\Ub}_r \widehat{\Sbb}_r^{1/2}.
$
Define the maximal expected degree
\[
\cD := \max_{i\in[n]} \sum_{j \ne i} P_{ij},
\]
and let
\[
\gamma n := \min_{i \in [r]} \big| S_{ii} - S_{i+1,i+1} \big| > 0
\]
denote the minimum eigengap of $\Sbb_r$. The quantity $\cD$ controls the local concentration of degrees and is directly connected to the eigengap condition through the Gershgorin circle theorem.  By assuming that $\gamma n$ is large enough relative to $\cD$, \cite{lyzinski:2014} are able to prove a $2\to \infty$ norm bound on the estimation error of $\Yb^*$ which depends on $\cD$, $\gamma$, $r$ and $n$. This bound allows us to apply Theorem~\ref{thm-lloyd-guarantee} to the RDPG setting below.

Suppose that $\cF$ is a mixture of $K$ sub-Gaussian distributions, $\{\cF_k\}_{k\in [K]}$, with parameter $\sigma$ and assume that for all $\by^*_1,\by^*_2$ in the support of $\cF$, $(\by^*_1)^{\top}\by^*_2 \in [0,1]$. Define $\bmu_k := \E \cF_k$ and $\Mb = (\bmu_1,\ldots,\bmu_K)^{\top}$. Let $\Zb \in \bbR^{n \times K}$ be the membership matrix given by $\Zb_{ik} = 1$, when $\by^*_i$ is drawn from the $k$-th mixture component $\cF_k$ and $0$ otherwise. Then $\E\Yb^* = \Zb \Mb$ and we may write $\Yb$ as
\begin{equation}
    \Yb =\Zb\Mb + (\Yb - \Yb^*) + (\Yb^* - \Zb\Mb),
\end{equation}
where $\Zb\Mb \in \bbR^{n\times r}$ consists of $K$ distinct rows, $\|\Yb - \Yb^*\|_{2\to\infty}$ can be controlled and $\Yb^* - \Zb\Mb$ is a row-independent sub-Gaussian matrix with parameter $\sigma$ in each entry. Hence, $\Yb$ can be written as the general clustering model in \eqref{eqn:model_matrix}. With parameters $\Delta_{\min}, \gC_s$, and $G_s$ defined as in Section \ref{s:setup_metrics}, we can apply Theorem \ref{thm-lloyd-guarantee} to the RDPG setting. 

\begin{theorem}\label{thm-RDPG}
Suppose $\Pb$ is of rank $r$ and that the non-zero eigenvalues of $\Pb$ are distinct. Assume there exists $0 < \eta < \frac{1}{2}$ such that $\gamma n \geq 4\sqrt{\cD \log\frac{n}{\eta}}$, and
that the assumptions of Theorem \ref{thm-lloyd-guarantee} hold including the initialization condition. Then,
\begin{equation}
A_s \leq \max \left\{ \exp \left( -\dfrac{\Delta_{\min}^2}{16 \sigma^2} \right) , \exp\left( -\frac{\Delta_{\min}^2(\gamma n)^{7/2}}{680r\cD^3 \sigma\log \frac{n}{\eta}} \right) \right\} \quad \text{for all} \quad s\geq 4\log n
\end{equation}
with probability at least $1 -\delta(n,\sigma, \Delta_{\min},\epsilon) - 2\eta$.
\end{theorem}

This dependence on $\cD$ and $\gamma$ is natural, since the eigengap controls the accuracy of the spectral embedding. As $\eta$ decreases, the required eigengap increases, but the corresponding probability bound improves.        
         
\subsection{Community dynamic factor models}\label{sec:CDFM}
The Dynamic Factor Model (DFM) \citep{geweke1977} extends factors models, developed to explain variations in large numbers of variables by a few latent variables called factors, to multivariate time series. How much each factor contributes to the time series is controlled by a weight, or loading, matrix. The ability of DFMs to model multivariate time series when both the number and length of the time series become extremely large has led to widespread adoption \citep{bai:2008,fan2021,forni2000,kim2003,ng1992}. Although dynamic factor models have been extensively studied, most existing approaches do not accommodate the general clustering model \eqref{y_i^*}. The results in this section pertain to the recently developed Community Dynamic Factor Model (CDFM) \citep{bhamidi2023correlation} which extends the DFM by incorporating a mixture structure into the loading matrix where each mixing distribution represents a community. 

The CDFM postulates that an $n$-dimensional stationary time series $\{\mvX_t\}_{t\in \bbZ}$ follows a dynamic factor model (DFM) given by
\begin{equation}\label{DFM}
    \mvX_t = \mvLambda \mvf_t + \mvgee_t, \;\;\; t \in \bbZ,
\end{equation}
where $\{\mvf_t\}_{ t\in \bbZ }, \mvgL,$ and $\{\mvgee_t\}_{t \in \bbZ}$ are defined as follows. The factor series $\{\mvf_t\}_{t\in\mathbb{Z}}$ is an $r$-dimensional stationary, mean-zero process satisfying $\mathbb{E}[\mvf_t\mvf_t^\top]=\mvI_r$. The loading matrix $\mvLambda = (\mvlambda_i)_{i=1}^n$ is an $n\times r$ matrix whose $i$-th row is $\mvlambda_i^\top \in \mathbb{R}^r$. The error process satisfies $\mathbb{E}[\mvgee_t] = \mvzero$ and $\mathbb{E}[\mvgee_t\mvgee_t^\top] = \mvgS_\epsilon$. The errors $\{\mvgee_t\}$ and factors $\{\mvf_t\}$ are assumed to be independent. Lastly, we assume that $\mathbb{E}[X_{i,t}^2]=1$ for all $(i,t)$, which ensures that the covariance and correlation matrices coincide.

Community structure is introduced by assuming that the rows of $\mvgL$ follow a mixture distribution $\cF$. For $K\ge 1$, let ${\cF_k: k\in[K]}$ be distinct sub-Gaussian probability measures on $\mathbb{R}^r$ with common parameter $\sigma>0$, and let $\pb={p_k:k\in[K]}$ be a probability mass function. Define the mixture as 
\begin{equation}\label{CDFM_mixture}
\cF = \sum_{k=1}^K p_k \cF_k
\end{equation}
and suppose that $\mvgl_i$'s are drawn i.i.d. from $\cF$.
Note that each loading vector admits the decomposition 
$$
\mvgl_i = \bmu_{z_i} + \bw_i,
$$
where $z$ is the membership function identifying which mixing distribution the loading distribution is drawn from, $\bmu_{z_i} = \E \cF_{z_i}$, and $\{\bw_i\}$ are centered sub-Gaussian random variables with sub-Gaussian parameter $\sigma>0$. We may then write the estimated loadings in the form of the general clustering model \eqref{y_i^*} as 
\begin{equation}\label{CDFM_clustering_model}
\wh\mvgl_i = \bmu_{z_i} + \bw_i + \wh\mvgl_i-\mvgl_i,
\end{equation}
where the estimation error $\wh\mvlambda_i - \mvlambda_i$ plays the role of the perturbation term $\eb_i$. Hence, under the sub-Gaussian $K$-mixture \eqref{CDFM_mixture} we may define the parameters $\alpha, \Delta_{\min},\sigma, \Delta_{\max}, \rho_{\sigma}, G_s, \gC_s$, and $A_s$ as in Section \ref{s:setup_metrics}.

Suppose we observe $\{\mvX_t\}_{t\in [T]}$ for some $T \in \bbZ$. The loading matrix can then be estimated using PCA. Under suitable assumptions, the PCA estimator satisfies a uniform high-probability error bound of the form \eqref{eqn:model}. More precisely, let $\wh\mvgS_{X} = \wh\mvQ \wh\mvD \wh\mvQ^{\top}$ be the eigendecomposition of the sample covariance matrix where $\wh\mvQ$ consists of the orthogonal eigenvectors and the diagonal matrix $\wh\mvD$ consists of the respective eigenvalues, in a decreasing order. Let $\wh\mvQ_r$ denote the $d\times r$ matrix of the $r$ leading eigenvectors of $\wh\mvgS_X$ associated with the $r$ largest eigenvalues forming a diagonal matrix $\wh\mvD_r$. Then, the PCA estimators are given by
\begin{equation}\label{PCA estimates}
\wh\mvgL = \wh\mvQ_r \wh\mvD_r^{\frac{1}{2}}, \quad \wh\mvf_t = \wh\mvD_r^{-1}\wh\mvgL^{\top}\mvX_t.
\end{equation}
By construction,
\begin{equation}\label{PCA constraints}
\wh\mvgL^{\top}\wh\mvgL = \wh\mvD_r , \quad \dfrac{1}{T}\sum_{t=1}^{T} \wh\mvf_t\wh\mvf_t^{\top} = \mvI_r.
\end{equation}

Let $\mvgL^{\top}\mvgL = \mvQ_{\gl} \mvD_{\gl} \mvQ_{\gl}^{\top}$ be the eigendecomposition of $\mvgL^{\top}\mvgL$. Under suitable assumptions (\cite{bai:2008} and \cite{doz2012}), as $n,T \to \infty$,
\begin{equation}\label{convergence of ests}
\|\wh\mvgL - \mvgL^{(0)}\| \to 0, \quad \wh\mvf_t \stackrel{p}{\to} \mvf_t^{(0)},
\end{equation}
where $\mvgL^{(0)} := \mvgL\mvQ_{\gl}$, $\mvf_t^{(0)} := \mvQ_{\gl}^{\top} \mvf_t $, and $\stackrel{p}{\to}$ denotes convergence in probability.  

To apply Theorem \ref{thm-lloyd-guarantee}, we require a bound on $\|\wh\mvgL - \mvgL\|_{2\to\infty}$, which in turn necessitates additional assumptions on the DFM \eqref{DFM}. To obtain such bounds, we adopt the notation and results of \cite{uematsu:2019}. 

Let $\mvF = (\mvf_1,\ldots,\mvf_T)^{\top}$ denote the factor matrix and $\mvE = (\mvgee_1,\ldots ,\mvgee_T)^{\top}$ the error matrix. Define $L_n = \max \{n,T\}^v - 1$ for some fixed constant $v>0$. We assume that $v$ remains fixed throughout. Define $\gt = \frac{\log T}{\log n}$ so that $T = n^{\gt}$. 
\begin{asm}\label{Uematsu_assumps}
\begin{enumerate}
\item [(a)] The factor matrix $\mvF$ is specified as the vector linear process $\mvf_t = \sum_{\ell=0}^\infty \mvgP_\ell \mvgz_{t-\ell}$, where $\mvgz_t \in \bR^r$  are vectors of i.i.d.\ $subG(\sigma_{\gz}^2)$ entries with standardized second moments and $\sum_{\ell=0}^\infty \mvgP_\ell \mvgP_\ell' = \mvI_r$. Moreover, there are $C_f>0$ and $\ell_f\in\bN$ such that $\|\mvgP_\ell\|_2\leq C_f \ell^{-(\nu+2)}$ for all $\ell\geq \ell_f$. 

\item [(b)] The error matrix $\mvE$ is independent of $\mvF$ and is specified as the vector linear process $\mvgee_t = \sum_{\ell=0}^\infty \mvPsi_\ell \mvxi_{t-\ell}$, where  $\mvxi_t \in \bR^n$ are vectors of i.i.d.\ $subG(\sigma_{\xi}^2)$ entries, and $\mvPsi_0$ is assumed to be nonsingular and lower triangular. Moreover, there are $C_e>0$ and $\ell_e\in\bN$ such that $\|\mvPsi_\ell\|_2\leq C_e \ell^{-(\nu+2)}$ for all $\ell\geq \ell_e$.

\item [(c)] Suppose there exists $\alpha_i \in \bbR$, $i\in [r]$, such that the number of nonzero elements in the $i$-th column is given by $n^{\alpha_i}$. Assume that $0 < \alpha_r \leq \ldots \leq \alpha_1 \leq 1$.

\item[(d)] Assume that $\mvgL^{\top}\mvgL$ is diagonal with entries $\Delta_i^2$, $i \in [r]$, and that $
0 < \Delta_r n^{\ga_r/2} \le \cdots \le \Delta_1 n^{\ga_r/2} < \infty.$
Moreover, whenever $\alpha_j = \alpha_{j-1}$ for some $j$, suppose there exists a constant $g>0$ such that 
$
\Delta_{j-1}^2 - \Delta_j^2 \;\ge\; g^{1/2} \Delta_{j-1}^2.
$

\item [(e)] Suppose $\alpha_r + \gt >1$ and $\alpha_1 + \frac{\max \{1,\gt\}}{2} < \alpha_r + \min\{\alpha_r,\gt\}$.
\end{enumerate}
\end{asm}

Assumptions \ref{Uematsu_assumps} (a) and (b) are stronger than the standard assumptions in the DFM literature (e.g., \cite{bai:2008}), but relaxing them likely hinges on finding suitable technical arguments rather than some other inherent obstacles. Assumption \ref{Uematsu_assumps} (c) allows $\mvgL$ to be sparse, and Assumption \ref{Uematsu_assumps} (d) ensures that $\mvgL^{\top}\mvgL$ is a diagonal matrix whose entries satisfy a gap condition. Assumption \ref{Uematsu_assumps} (e) imposes a condition on the sparsity of $\mvgL$ relative to $n$ and $T$.

A diagonal $\mvgL^T \mvgL$ is needed to enforce identifiability. Without which, PCA would estimate a rotated loadings matrix \cite{bai:2008} and we could not obtain a uniform bound on the estimation error of the loadings. When $\ga_1 = \cdots = \ga_r = 1$, the assumptions imply that the $r$ largest eigenvalues of $\mvgL^T \mvgL$ diverge proportionally to $n$, ensuring the pervasiveness of the factors \cite{guo2023}. This is known as the strong factor model setting and it forces an eigengap condition by separating the $r$ largest eigenvalues from the rest as $n$ increases. However, when $\ga_i < 1$ for some $i \in [r]$, the eigengap condition is weakened by allowing slower convergence rates of the eigenvalues, thus weakening the pervasiveness of the factors. Under these assumptions, \cite{uematsu:2019} show that the estimation error of the loadings can be uniformly bounded. Thus, we may apply Theorem \ref{thm-lloyd-guarantee} to the CDFM setting.

\begin{theorem}
    \label{thm:cdfm}
    Suppose Assumption \ref{Uematsu_assumps} holds for the DFM \eqref{DFM}, 
    \begin{align}
     n\alpha &\geq C_{1} K\log n, \\
     \rho_{\sigma} &\geq C_{2} \sqrt{K}, \\
    \min \bigg \{ \dfrac{\alpha\Delta_{\min}^2}{Kr^2}, \dfrac{\Delta_{\min}^4}{r^4\sigma^2} \bigg \} &\geq C_{3} \dfrac{\log(\max \{n,T\})}{T} \label{CDFM_gee_assumption},
    \end{align}
    and that the initialization condition \eqref{initial_conditions1} or \eqref{initial_conditions2} holds. Then, for some $C>0$, 
    \begin{equation*}
    A_s \leq \max \left\{ \exp \left( -\dfrac{\Delta_{\min}^2}{16 \sigma^2} \right) , \exp\left( -\frac{\Delta_{\min}^2}{8C\sigma r  \sqrt{\frac{\log(\max \{n,T\})}{T}}} \right) \right\} \text{  for all  }  s\geq 4\log n
    \end{equation*}
    with probability at least $1 -\delta(n,\sigma, \Delta_{\min},\epsilon) -  O(\max\{n,T\}^{-v})$.
\end{theorem}

\begin{rem}
Theorem \ref{thm:cdfm} is an application of Theorem \ref{thm-lloyd-guarantee} with $$\gee = C r \left ( \dfrac{\log(\max\{n,T\})}{T} \right)^{\frac{1}{2}}.$$ As shown in the proof in Section \ref{sec:proofs-applications}, we take advantage of a uniform bound on the estimation error obtained from \cite{uematsu:2019} and ensure that the assumptions of Theorem \ref{thm-lloyd-guarantee} are met. 
\end{rem}

\section{Proof roadmap and technical challenges}\label{s:roadmap}

The proof of Theorem~\ref{thm-lloyd-guarantee} follows the recursive error analysis introduced by \cite{lu2016statistical}, but requires several new ideas to handle the perturbed observation model~\eqref{eqn:model}.  In this section, we outline the main steps of the analysis and highlight the key technical challenges that arise in the presence of both sub-Gaussian noise and additive perturbations.

\subsection{Overview of the proof strategy}

The analysis proceeds by tracking two coupled error metrics across Lloyd's iterations: the misclustering rate $G_s$ and the normalized center error $\Gamma_s$. The goal is to show that, under a suitable initialization, both quantities contract at every iteration, yielding the misclustering guarantee stated in Theorem~\ref{thm-lloyd-guarantee}. The proof consists of the following steps.

\noindent\textbf{Step 0: Concentration of sub-Gaussian noise.}
Lemmas~\ref{lem:subg-sum}--\ref{lem:indicator-bernstein} establish high-probability 
bounds for averages of the sub-Gaussian noise components appearing in the updates of the empirical cluster means and assignment scores. These concentration results are used repeatedly throughout the proof: they allow us to treat the empirical Lloyd step as a small perturbation of its population counterpart.

\noindent\textbf{Step 1: One-step contraction bounds.}
Lemmas~\ref{lem:Gamma-step} and~\ref{lem:G-step} provide coupled one-step recursions for the error metrics,
\[
G_{s+1} \le f(\Gamma_s), 
\qquad 
\Gamma_{s+1} \le g(G_{s+1},\Gamma_s),
\]
where the functions $f$ and $g$ strictly contract whenever $(G_s,\Gamma_s)$ lie in a suitable neighborhood of the origin.  The key idea is that accurate empirical centers imply accurate assignments (and vice versa), and the concentration results from Step~0 control the stochastic fluctuations in this iterative improvement.

\noindent\textbf{Step 2: Initialization region.}
We show that if either $G_0$ or $\Gamma_0$ is below the thresholds stated in 
Theorem~\ref{thm-lloyd-guarantee}, then $(G_s,\Gamma_s)$ remain uniformly within the contraction region for all $s$. 

\noindent\textbf{Step 3: Controlling the assignment step.}
The misclustering event for a point $i$ with true label $z_i$ can be reduced to
\[
\mathbb I\!\{ \widehat z_i^{(s+1)} = h,\ z_i \neq h \}
\;\le\;
\mathbb I\!\left(
\beta_1 \|\mu_{z_i}-\mu_h\|^2
< 2\langle y_i - \mu_{z_i},\ \widehat\mu_h^{(s)} - \widehat\mu_{z_i}^{(s)} \rangle
\right).
\]
Using the initialization guarantee from Step~2, the factor $\beta_1$ is strictly bounded. We then decompose
\[
\widehat\mu_h^{(s)} - \widehat\mu_{z_i}^{(s)}
= (\mu_h - \mu_{z_i})
  + \bigl(\widehat\mu_h^{(s)} - \mu_h\bigr)
  - \bigl(\widehat\mu_{z_i}^{(s)} - \mu_{z_i}\bigr),
\]
and show that the perturbation terms are controlled by $G_s$ and $\Gamma_s$ 
via the concentration results from Step~0. Intuitively, when the clustering is accurate, the empirical separation between clusters remains close to the population separation, and the fluctuation terms are small.

\noindent\textbf{Step 4: Divergence from the proof of Lu and Zhou (2016).}
At this stage our argument departs from the proof of Theorem~3.2 in 
\cite{lu2016statistical}. In our setting, the empirical means contain two sources of randomness (sub-Gaussian noise and additive perturbation), so the decomposition above contains additional error terms that do not appear in~\cite{lu2016statistical}. Controlling these terms requires the sharper concentration inequalities developed in Step~0 and leads to a different one-step recursion for $(G_s,\Gamma_s)$.

\noindent\textbf{Step 5: Iterating the recursion.}
Once $(G_s,\Gamma_s)$ are confined to the contraction region, repeated application of the one-step bounds yields exponential decay of the errors. After $s = \Theta(\log n)$ iterations, this gives the final misclustering bound stated in Theorem~\ref{thm-lloyd-guarantee}.

We now briefly explain how this proof strategy adapts to spectral clustering under SBMs.

The proof of Theorem~\ref{thm:sbm_K} is inspired by the analysis of Theorem~\ref{thm-lloyd-guarantee}, but differs in important structural aspects. For spectral clustering under stochastic block models, the sources of error arise from the same underlying edge-wise Bernoulli randomness, rather than from independent sub-Gaussian noise and additive perturbations.

There are two main differences. First, the error matrices $\Eb_1$ and $\Eb_2$, defined in \eqref{eqn:model_SBM}, admit more explicit representations under the SBMs, allowing for sharper bounds. Second, instead of appealing to generic sub-Gaussian concentration, we exploit concentration inequalities tailored to Bernoulli random variables and adjacency matrices.

As a result, the overall proof strategy closely parallels that of
Theorem~\ref{thm-lloyd-guarantee}, but the concentration analysis is both tighter and more direct at each step of the argument. Rather than repeating a high-level proof roadmap, we defer the detailed proofs to the appendix.

\subsection{Technical challenges}
\label{subsec:technical-challenges}
The noiseless setting ($\epsilon \equiv 0$) was analyzed in the foundational work of \cite{lu2016statistical}, where the behavior of Lloyd's algorithm under sub-Gaussian noise was characterized. Extending this framework to the perturbed model considered in the present paper requires addressing several additional challenges. 
\paragraph{(i) Interacting noise sources.} Each Lloyd update now contains three types of error terms: the sub-Gaussian noise, the additive perturbation, and their interaction. Controlling these jointly necessitates a new decomposition of the empirical updates and the development of sharper concentration inequalities than those used in the noiseless setting. \paragraph{(ii) Community detection and sparse Bernoulli observations.} In the stochastic block model application, the adjacency matrix entries are Bernoulli random variables. Although Bernoulli variables are technically sub-Gaussian, the standard sub-Gaussian parameters are far too loose in sparse regimes. To obtain tight bounds, our analysis works directly with sums of Bernoulli variables rather than relying on generic sub-Gaussian inequalities. 
\paragraph{(iii) Verifying the conditions in applications.} Our main theorem requires certain initialization and separation conditions on $(G_0,\Gamma_0)$ and on the cluster structure. We show that these conditions hold in several important applications by combining our Lloyd update bounds with recent results from random matrix theory. This explains the empirical robustness of Lloyd’s algorithm in practical high-dimensional problems. These challenges motivate the new technical tools developed in this work and shape the structure of the proof presented.



\vspace{5mm}

\begin{appendix}
\textbf{Appendix overview:} Appendix~A collects two complementary results that are not required for the main proofs. Section~A.1 shows how the Lloyd guarantees imply consistency of SigClust in a simple sub-Gaussian mixture model. Section~A.2 provides a minimax lower bound in a restricted Gaussian-noise setting, which illustrates how the perturbation level $\epsilon$ affects the achievable error exponent. Appendices~B and C contain the proofs for Sections~\ref{s:theoretical_results} and~\ref{s:applications}, respectively.

\section{Additional theoretical consequences of Lloyd’s algorithm}
\subsection{Implications for testing the significance of clustering} 
We study the second implication of Theorem~\ref{thm-lloyd-guarantee} in the context of testing the statistical significance of clusters produced by $k$-means, focusing on the SigClust procedure \citep{liu2008statistical}. While clustering methods can always produce partitions, determining whether the resulting clusters reflect genuine structure rather than sampling artifacts remains challenging. SigClust addresses this issue by testing whether the observed clustering is consistent with data generated from a single Gaussian distribution.

Under SigClust, the null hypothesis is
\[
H_0:\ \text{the data are generated from a single $r$-dimensional Gaussian distribution}.
\]
Given a partition $\widehat{\cC}=\{\widehat{\cC}_1,\widehat{\cC}_2\}$ of a data set $\cX$, the test statistic is the cluster index
\[
\cC\cI(\widehat{\cC}_1,\widehat{\cC}_2;\cX)
:=\frac{\sum_{j\in\widehat{\cC}_1}\|\by_j-\widehat{\bmu}_1\|^2+\sum_{j\in\widehat{\cC}_2}\|\by_j-\widehat{\bmu}_2\|^2}
{\sum_{j=1}^n\|\by_j-\bar{\by}\|^2},
\]
where $\widehat{\bmu}_1,\widehat{\bmu}_2$ are the cluster means and $\bar{\by}$ is the sample mean. The null distribution of $\cC\cI$ is approximated by estimating the covariance structure under $H_0$, simulating Gaussian data accordingly, reclustering the simulated data using $k$-means, and comparing the observed cluster index to the resulting empirical distribution; see \cite{liu2008statistical} for details.

We consider a two-component sub-Gaussian mixture model of the form
\[
\cF \sim \tfrac12(\mathbf f+a\be_1)+\tfrac12(\mathbf f-a\be_1),
\]
where $\mathbf f$ has independent, mean-zero, sub-Gaussian components with variance proxy $\sigma^2$, $a>0$, and $\be_1$ denotes the first canonical basis vector. In this setting, the mean separation satisfies $\Delta_{\min}=\Delta_{\max}=2a$.

Define
\[
\Psi(t)
:= \left(1-\frac{2}{\pi}\right)t
-48(\sqrt{2}+1)^2 e^{-4t}
-64t e^{-8t}
-64\sqrt{3}(\sqrt{2}+1)\sqrt{t}\, e^{-6t}.
\]
\begin{theorem}
\label{thm:sigclust-consistency}
Assume that the conditions of Theorem~\ref{thm-lloyd-guarantee} hold with $\epsilon=0$, and let
$\{\widehat{\cC}_1,\widehat{\cC}_2\}$ be the partition obtained after $4\log n$ iterations of Lloyd’s algorithm.
Suppose further that
\[
\Psi\!\left(\frac{a^2}{\sigma^2}\right) > \frac{2}{\pi}.
\]
Then the SigClust procedure applied to $\cX$ using the partition
$\{\widehat{\cC}_1,\widehat{\cC}_2\}$ is asymptotically consistent, in the sense that
$\Pr(\text{reject } H_0)\to 1$ as $n\to\infty$.
\end{theorem}

\begin{rem}
Theorem~\ref{thm:sigclust-consistency} extends the theoretical guarantees of SigClust to general sub-Gaussian mixture models in the low-dimensional setting. Combined with Theorem~\ref{thm-lloyd-guarantee}, it shows that SigClust can correctly detect the presence of true clusters recovered by Lloyd’s algorithm.
\end{rem}

\begin{rem}
For high-dimensional data $\Xb\in\bbR^{n\times p}$ with large $p$, an extension of SigClust known as SigClust-MDS \citep{shen2024statistical} first computes a low-dimensional MDS embedding and then applies SigClust. Using Theorem~\ref{thm:mds} together with Corollary~\ref{corr:lloyd-center}, an analogue of Theorem~\ref{thm:sigclust-consistency} can be derived for SigClust-MDS, with the only difference being that the embedding error corresponds to $\epsilon\neq0$.
\end{rem}

\textbf{Proof of Theorem \ref{thm:sigclust-consistency}:}
Denote $v(\sigma^2) = var(f)$ as the variance of the mean-zero, symmetric sub-Gaussian distribution. Using the property of sub-Gaussianity, we have 
$$
v(\sigma^2) \leq \sigma^2. 
$$
It will be convenient to construct the datasets across $n$ on the same probability space so one can define $\limsup$ and $\liminf$ etc of various sequences of random variables. We also let $\cC_1, \cC_2$ denote the true clusters (where these depend on $n$ but we suppress this for simplicity).  By Corollary \ref{corr:lloyd-center} (also see the statement of\cite[Theorem 6.2]{lu2016statistical}), under the Assumptions of Theorem \ref{thm-lloyd-guarantee}, the cluster means $\widehat{\boldsymbol{\bmu}}_{\widehat{\cC_1}}, \widehat{\boldsymbol{\bmu}}_{\widehat{\cC_2}}$ that Lloyd's algorithm converges to satisfy (with the same probability guarantees as in the original result), 
\begin{equation}
    \label{eqn:clust-mean-bound}
    \begin{split}
        \max_{i\in \set{1,2}} ||\widehat{\boldsymbol{\bmu}}_{\widehat{\cC_i}} - \mvmu_{\cC_i}||_2 
        &\leq 4\sqrt{3}(\sqrt{2}+1)\sigma\sqrt{\frac{{n+r}}{n}}\exp\left(-\frac{2a^2}{\sigma^2}\right) \\
        &\quad + 8a\exp\left(-\frac{4a^2}{\sigma^2}\right) + 
        \sigma O\left(\sqrt{\frac{\log{n}}{n}}\right).
    \end{split}
\end{equation}
Next by the definition of Lloyd's algorithm where points are assigned to their nearest centroids, 
\begin{align*}
   \sum_{i=1}^2 \sum_{j\in \widehat{\cC_i}}||\by_j - \widehat{\boldsymbol{\bmu}}_{\widehat{\cC_i}} ||_2^2 &\leq \sum_{i=1}^2 \sum_{j\in \cC_i}||\by_j - \widehat{\boldsymbol{\bmu}}_{\widehat{\cC_i}} ||_2^2  
   = \sum_{i=1}^2 \sum_{j\in \cC_i}||\by_j - {\boldsymbol{\bmu}}_{{\cC_i}} ||_2^2 + \Upsilon_{n},
\end{align*}
where using \eqref{eqn:clust-mean-bound} and laws of large numbers, whp as $n\to\infty$, 
\begin{align*}
\frac{\Upsilon_n}{n} 
\leq & 48(\sqrt{2}+1)^2\sigma^2\frac{n+r}{n}\exp\left( -\frac{4a^2}{\sigma^2}\right) 
+ 64 a^2\exp\left(-\frac{8a^2}{\sigma^2}\right) + \sigma^2 O\left(\frac{\log n}{n}\right) \\
& + 64\sqrt{3}(\sqrt{2}+1)a\sigma \sqrt{\frac{n+r}{n}} \exp\left(-\frac{6a^2}{\sigma^2}\right) \\
& + 8\sqrt{3}(\sqrt{2}+1)\sigma\sqrt{\frac{n+r}{n}}\exp\left(-\frac{2a^2}{\sigma^2}\right) O\left(\sqrt{\frac{\log n}{n}}\right) \\
& + 16a\sigma\exp\left( -\frac{4a^2}{\sigma^2}\right) O\left(\sqrt{\frac{\log n}{n}}\right) 
+ \sqrt{v(\sigma^2)}O\left(\sqrt{\frac{\log n}{n}}\right).
\end{align*}
It is easy to check that the population covariance matrix for $\Gb$ is $\Sigma=\operatorname{diag}(v(\sigma^2)+a^2,v(\sigma^2),\ldots, v(\sigma^2))$. Thus by the laws of large numbers, 
\begin{align} 
\limsup_{n\to\infty}\cC\cI(\widehat{\cC}_1, \widehat{\cC}_2; \cX) 
\leq & \frac{rv(\sigma^2)}{a^2+rv(\sigma^2)}  
+ \frac{48(\sqrt{2}+1)^2\sigma^2}{a^2+rv(\sigma^2)}\exp\left(-\frac{4a^2}{\sigma^2}\right) \nonumber \\
& + \frac{64a^2}{a^2+rv(\sigma^2)}\exp\left(-\frac{8a^2}{\sigma^2} \right) \nonumber \\
& + \frac{64\sqrt{3}(\sqrt{2}+1)a\sigma }{a^2+rv(\sigma^2)}\exp\left(-\frac{6a^2}{\sigma^2}\right). 
\label{eqn:1057} 
\end{align}
Now the comparative distribution under the null hypothesis (since the test statistic is invariant under-scaling), for large $n$, the cluster index is compared to a data from a normal $N_r(\mvzero, \Sigma)$, where $\Sigma = \operatorname{diag}(a^2+v(\sigma^2),v(\sigma^2),\ldots,v(\sigma^2))$. 
We will write $\cX_{H_0}$ for data generated according to this distribution and the corresponding data points as $\set{\by_{i,H_0}: 1\leq i\leq n}$ and $\boldsymbol{Y}_{H_0} \sim N_r(\mvzero, \Sigma)$.  Standard empirical process results (see \cite{pollard:1981,telgarsky2013moment,klochkov2021robust}) imply that in this setting, 
\begin{equation}
    \frac{1}{n} \min_{\bc_1, \bc_2\in \bB^r} \sum_{i=1}^n \min_{j=1,2}||\by_{i,H_0} - \bc_j||_2^2 \probc \E(\min_{\bc_1, \bc_2\in \bB^r} ||\boldsymbol{Y}_{H_0} - \bc_i||_2^2).
\end{equation}
Combining results in \cite{pollard:1981,pollard:1982a,bock1985some}, \cite[Appendix B]{chakravarti2019gaussian} shows that in this case,  the optimal $2$-means cluster centers are given by, 
\begin{align*}
\boldsymbol{\mu}_1 & =\left(-\sqrt{\frac{2 (v(\sigma^2)+a^2)}{\pi}}, 0, \ldots, 0\right)^{\top}, \quad \text { and } \\
\boldsymbol{\mu}_2 & =\left(\sqrt{\frac{2 (v(\sigma^2)+a^2)}{\pi}}, 0, \ldots, 0\right)^{\top} .
\end{align*}
The corresponding optimal population clusters are
\begin{align*}
& A_1=\left\{\mathbf{y}=\left(y_1, \ldots, y_r\right) \in \mathbb{R}^r: y_1 \leq 0\right\}, \quad \text { and } \\
& A_2=\left\{\mathbf{y}=\left(y_1, \ldots, y_d\right) \in \mathbb{R}^r: y_1>0\right\} .
\end{align*}
Using the optimal cluster centers, it is easy to check that
\[\E(\min_{\bc_1, \bc_2\in \bB^r} ||\boldsymbol{Y}_{H_0} - \bc_i||_2^2) = r\E(\min_{c_1, c_2\in \bR}|Y_i - c_i|^2 ) = (1-\frac{2}{\pi})(v(\sigma^2)+a^2)+(r-1)v(\sigma^2).\]
Thus under the null hypothesis, in the large $n\to\infty$ limit, the cluster index under the null hypothesis converges to, 
\[\cC\cI_{H_0} \probc 1-\frac{2}{\pi} \frac{a^2+v(\sigma^2)}{a^2+r v(\sigma^2)}, \qquad \mbox{ as } n\to\infty.  \]
Comparing this with \eqref{eqn:1057} completes the result using the assumption and the fact that $v(\sigma^2)\leq \sigma^2$.

\subsection{A minimax lower bound under Gaussian noise with bounded perturbations}
To complement the upper bound, we provide a minimax lower bound under a Gaussian noise model with bounded perturbations. This lower bound is derived in a restricted setting and illustrates one mechanism by which $\epsilon$ degrades the achievable error exponent. Define a parameter space as follows,
$$
\begin{gathered}
\Theta=\left\{(\bmu, z), \bmu=\left[\bmu_1, \cdots, \bmu_k\right] \in \bbR^{r \times k}, \Delta \leq \min _{g \neq h}\left\|\bmu_g-\bmu_h\right\|,\right. \\
\left.z:[n] \rightarrow[k],\left|\left\{i \in[n], z_i=u\right\}\right| \geq \alpha n, \forall u \in[k]\right\}.
\end{gathered}
$$

\begin{theorem}(Lower bound) For model \eqref{eqn:model}, assume independent Gaussian noise $w_{i j} \stackrel{\text { i.i.d }}{\sim}$ $\mathcal{N}\left(0, \sigma^2\right)$, if $\frac{\Delta}{\sigma\left(1+\frac{\epsilon}{\sqrt{r}\sigma +\Delta }\right)} \rightarrow \infty$, then 
$$
\inf _{\hat{z}} \sup _{(\bmu, z) \in \Theta} \mathbb{E} \ell(\hat{z}, z) \geq  \exp \left(-(1+o(1)) \dfrac{\Delta^2}{8\sigma^2\left(1+\dfrac{\epsilon}{\sqrt{r}\sigma +\Delta }\right)^2} \right). 
$$
\end{theorem}

\begin{proof}
Following the idea in the proof of Theorem 2 in \cite{gao2018community}, we first define a clustering problem on a specific subset of the parameter space, allowing us to bypass the complexity of label permutation. To this end, given $z^*$, for each $u \in[k]$, let $T_u \subset\left\{i:z^*(i)=u\right\}$ with cardinality $\left\lceil n_u\left(z^*\right)-\frac{\alpha n}{4 k}\right\rceil$. Let $T=\cup_{u=1}^k T_u$. 

From the proof of Theorem 3.3 in \cite{lu2016statistical}, we can lower-bound the minimax rate as follows
$$
\inf _{\hat{z}} \sup _{z \in \mathcal{Z}} \mathbb{E} \ell(\hat{z}, z) \geq \frac{\alpha}{6} \frac{1}{\left|T^c\right|} \sum_{i \in T^c}\left[\frac{1}{2 k^2} \inf _{\hat{z}_i}\left(\mathbb{P}_1\left\{\hat{z}_i=2\right\}+\mathbb{P}_2\left\{\hat{z}_i=1\right\}\right)\right],  
$$
where $\mathbb{P}_t, t \in\{1,2\}$ denote the probability distribution of our data given $z_i=t$.

Under our model \eqref{eqn:model} and Neyman-Pearson Lemma,
 we have 
 $$
\begin{aligned}
\inf _{\hat{z}_i}\left(\frac{1}{2} \mathbb{P}_1\left\{\hat{z}_i=2\right\}+\frac{1}{2} \mathbb{P}_2\left\{\hat{z}_i=1\right\}\right) & =\mathbb{P}\left\{\left\|\bmu_1+\bw_i+\be_i-\bmu_2\right\|^2 \leq\left\|\bw_i+\be_i\right\|^2\right\} \\
& = \mathbb{P}\left\{\left\|\bmu_1-\bmu_2\right\|^2 \leq 2\left\langle \bw_i+\be_i, \bmu_1-\bmu_2\right\rangle\right\}.
\end{aligned}
$$
Let $\be_i = \epsilon\dfrac{\bw_i}{\|\bw_i\|}$ where $\bw_i$ and $\be_i$ have correlation $1$, we have 
$$
\begin{aligned}
& \mathbb{P}\left\{\left\|\bmu_1+ \bw_i+ \be_i-\bmu_2\right\|^2 \leq\left\| \bw_i+\be_i\right\|^2\right\} \\
=&\mathbb{P}\left\{\left\|\bmu_1-\bmu_2\right\|^2 \leq 2\left(1+\frac{\epsilon}{\| \bw_i\|}\right)\left\langle \bw_i, \bmu_1-\bmu_2\right\rangle\right\}
\end{aligned}
$$
Recall that $\bw_{i} \sim \cN_r(\bzero, \sigma^2 \Ib)$. From Theorem 2.1 in \cite{hsu2012tail}, 
$$
\mathbb{P}\left\{\left\|\bw_i/\sigma \right\|^2 \geq r +2 \sqrt{r \log (1 / \delta)}+2 \log (1 / \delta)\right\} \leq \delta
$$
For simplicity, denote $\Delta_{\mu} = \bmu_1-\bmu_2$. Choose $\delta = \exp\left\{ -\dfrac{\Delta_{\mu}^2}{2\sigma^2} \right\}$. As a result, 
$$
\|\bw_i\| \leq \sqrt{r}\sigma + \Delta_{\mu},
$$
with probability at least $1 - \exp\left( -\dfrac{\Delta_{\mu}^2}{2\sigma^2} \right) $. 
To continue, 
$$
\begin{aligned}
    & \mathbb{P}\left\{\left\|\Delta_{\mu} \right\|^2 \leq 2\left(1+\frac{\epsilon}{\| \bw_i\|}\right)\left\langle \bw_i, \Delta_{\mu} \right\rangle\right\} \\
    = & \bbP\left\{ \left\|\Delta_{\mu} \right\|^2 \leq 2\left(1+\frac{\epsilon}{\| \bw_i\|}\right)\left\langle \bw_i, \Delta_{\mu}\right\rangle ,\|\bw_i\| \leq \sqrt{r}\sigma + \Delta_{\mu} \right\} \\
    + & \bbP\left\{ \left\|\Delta_{\mu} \right\|^2 \leq 2\left(1+\frac{\epsilon}{\| \bw_i\|}\right)\left\langle \bw_i, \Delta_{\mu}\right\rangle ,\|\bw_i\| > \sqrt{r}\sigma + \Delta_{\mu} \right\} \\
    \geq & \bbP\left\{ \left\|\Delta_{\mu} \right\|^2 \leq 2\left(1+\frac{\epsilon}{\| \bw_i\|}\right)\left\langle \bw_i, \Delta_{\mu}\right\rangle ,\|\bw_i\| \leq \sqrt{r}\sigma + \Delta_{\mu} \right\} \\
    \geq & \bbP\left\{ \left\|\Delta_{\mu} \right\|^2 \leq 2\left(1+\frac{\epsilon}{\sqrt{r}\sigma + \Delta_{\mu}}\right)\left\langle \bw_i, \Delta_{\mu}\right\rangle ,\|\bw_i\| \leq \sqrt{r}\sigma + \Delta_{\mu} \right\} \\
    \geq &  \bbP\left\{ \left\|\Delta_{\mu} \right\|^2 \leq 2\left(1+\frac{\epsilon}{\sqrt{r}\sigma + \Delta_{\mu}}\right)\left\langle \bw_i, \Delta_{\mu}\right\rangle \right\} + \bbP\left\{ \|\bw_i\| \leq \sqrt{r}\sigma + \Delta_{\mu} \right\}
\end{aligned}
$$
where the last inequality uses a standard probability fact 
\[
P(A \cap B) \geq P(A) + P(B) - 1. 
\]
Since $w_{i j} \stackrel{i.i.d.}{\sim}\mathcal{N}\left(0, \sigma^2\right),\left\langle \bw_i, \Delta_{\mu}\right\rangle \sim \mathcal{N}\left(0, \sigma^2\left\|\Delta_{\mu}\right\|^2\right)$. Let $\Phi(t)$ be the cumulative function of $\mathcal{N}(0,1)$ random variable. By calculating the derivatives, it can be easily proved that

$$
1-\Phi(t)=\frac{1}{\sqrt{2 \pi}} \int_t^{\infty} e^{-x^2 / 2} d x \geq \frac{1}{\sqrt{2 \pi}} \frac{t}{t^2+1} e^{-t^2 / 2}. 
$$
Then when $\left\|\Delta_{\mu}\right\| \geq \sigma$, we have
\begin{align*}
    & \bbP\left\{ \left\|\Delta_{\mu} \right\|^2 \leq 2\left(1+\frac{\epsilon}{\sqrt{r}\sigma + \Delta_{\mu}}\right)\left\langle \bw_i, \Delta_{\mu}\right\rangle \right\}\\
    \geq& \frac{\sigma \left(1+\dfrac{\epsilon}{\sqrt{r}\sigma + \Delta_{\mu}}\right) }{\sqrt{2 \pi}\left\|\Delta_{\mu} \right\|} \exp \left(-\frac{\left\|\Delta_{\mu}\right\|^2}{8 \sigma^2 \left(1+\dfrac{\epsilon}{\sqrt{r}\sigma + \Delta_{\mu}}\right)^2 }\right). 
\end{align*}
On the other hand, we have shown 
$$
\bbP\left\{ \|\bw_i\| \leq \sqrt{r}\sigma + \Delta_{\mu} \right\} \geq 1 - \exp\left( -\dfrac{\Delta_{\mu}^2}{2\sigma^2} \right).
$$
Combining two parts together,
$$
\begin{aligned}
    & \mathbb{P}\left\{\left\|\Delta_{\mu} \right\|^2 \leq 2\left(1+\frac{\epsilon}{\| \bw_i\|}\right)\left\langle \bw_i, \Delta_{\mu} \right\rangle\right\} \\
    \geq & \frac{\sigma \left(1+\dfrac{\epsilon}{\sqrt{r}\sigma + \Delta_{\mu}}\right) }{\sqrt{2 \pi}\left\|\Delta_{\mu} \right\|} \exp \left(-\frac{\left\|\theta_1-\theta_2\right\|^2}{8 \sigma^2 \left(1+\dfrac{\epsilon}{\sqrt{r}\sigma + \Delta_{\mu}}\right)^2 }\right) - \exp\left( -\dfrac{\Delta_{\mu}^2}{2\sigma^2} \right). 
\end{aligned}
$$
Under the condition that $\dfrac{\Delta}{\sigma\left(1+\frac{\epsilon}{\sqrt{r}\sigma +\Delta }\right)} \rightarrow \infty$, we can finally simplify the result as 
$$
\inf_{\hat{z}_i}\left(\frac{1}{2} \mathbb{P}_1\left\{\hat{z}_i=2\right\}+\frac{1}{2} \mathbb{P}_2\left\{\hat{z}_i=1\right\}\right)  =\exp\left\{ - (1+o(1)) \dfrac{\Delta^2}{8\sigma^2 (1+\frac{\epsilon}{\sqrt{r}\sigma + \Delta})^2} \right\},
$$
which finishes the proof. 
\end{proof}

\section{Proof of the main results}
\label{sec:proofs-main-res}
This section contains proofs of the main results, as described in Section \ref{s:theoretical_results}. 
\subsection{Proof of Theorem \ref{thm-lloyd-guarantee}}

The proof of Theorem~3.1 proceeds in three stages.  
Section~\ref{subsec:notation} recalls the notation used throughout the paper. First, Lemmas~\ref{lem:subg-sum}--\ref{lem:indicator-bernstein} in Section~\ref{subsec:tech-lemmas} establish concentration bounds for the sub-Gaussian noise. Second, Lemmas~\ref{lem:Gamma-step}--\ref{lem:G-step} in Section~\ref{subsec:one-step} control one Lloyd update in terms of $\Gamma_s$ and $G_s$.  
Finally, in Section~\ref{subsec:proof-lloyd}, we combine these ingredients and prove Theorem~3.1 by iterating the Lloyd recursion.

\subsubsection{Notation and auxiliary quantities}
\label{subsec:notation}

We recall the key quantities used to analyze Lloyd's iterations.
For $s \ge 0$ let
\[
G_s := \max\Bigl\{
  \max_{g\in[K]}\sum_{h\neq g} \frac{\wh n_{gh}^{(s)}}{n_g},
  \;
  \max_{h\in[K]}\sum_{g\neq h} \frac{\wh n_{gh}^{(s)}}{\wh n_h^{(s)}}
\Bigr\},
\]
\[
\Gamma_s := \frac{1}{\Delta_{\min}}
  \max_{h\in[K]} \|\wh\bmu_h^{(s)} - \bmu_h\|,
\qquad
A_s := \frac{1}{n}\sum_{i=1}^n \mathbb{I}\{z_i \neq \wh z_i^{(s)}\}.
\]

We also use the signal-to-noise quantities
\[
\rho_\sigma = \frac{\Delta_{\min}}{\sigma}\sqrt{\frac{\alpha n}{n+Kr}}, \qquad
\rho_\epsilon = \frac{\Delta_{\min}}{\epsilon}\sqrt{\alpha},
\]
and throughout assume $\rho_\sigma \ge C_2\sqrt{K}$ and $\rho_\epsilon \ge C_3$
for constants $C_2,C_3>0$.

For the proof of Theorem~\ref{thm-lloyd-guarantee}, fix a constant $c_\epsilon > 1$ and define
\[
\beta_{1,\epsilon} := c_\epsilon \frac{\epsilon}{\Delta_{\min}},\qquad
\beta_{2,\epsilon} := \frac{1}{\sqrt{\alpha}} \frac{\epsilon}{\Delta_{\min}} = \frac{1}{\rho_\epsilon},\qquad
\beta_{2,\sigma} := \sqrt{\frac{n+Kr}{n\alpha}} \frac{\sigma}{\Delta_{\min}} = \frac{1}{\rho_\sigma},
\]
and choose $\beta_{1,\sigma}>0$ so that
\[
\beta_{1,\epsilon} + \beta_{1,\sigma} + \beta_{2,\epsilon} + \beta_{2,\sigma}
= \frac{C_\Gamma}{2}, 
\]
where $C_\Gamma\in(0,1)$ is specified in Lemma~\ref{lem:G-step}.
Whenever needed we impose $\beta_{1,\sigma}>0$.

\subsubsection{Technical concentration lemmas}
\label{subsec:tech-lemmas}        
We closely follow the proof strategy of \cite{lu2016statistical} and
adapt it to our noisy-mixture setting. In particular, our analysis relies on sub-Gaussian concentration bounds corresponding to Lemmas~A.1–A.4 in \cite{lu2016statistical} and on a sharpened version
of their Lemma~A.5. For completeness, we restate these lemmas below with notation adapted to our setting. Throughout, for $S\subset[n]$ we write
\[
\Wb_S := \sum_{i\in S} \bw_i.
\]

\begin{lemma}[Norm of sub-Gaussian sum]
\label{lem:subg-sum}
$\|\Wb_S\|_2 \leq \sigma \sqrt{3(n+r)|S|} $ for all $ S\subset [n]$ with probability at least $1-\exp(-.3n)$. 
\end{lemma}

\begin{lemma}[Quadratic form bound]
\label{lem:subg-quad}
For any $\mvu \in \bbR^r$ and $S\subset [n]$,
$$\sum_{i\in S} (\bw_i^{\top}\mvu)^2 \leq 6\sigma^2(|S|+r)\|\mvu\|^2_2,$$ with probability at least $1-\exp(-.5n)$. 
\end{lemma}

\begin{lemma}[Inner product of empirical mean and a point]
\label{lem:inner-prod}
For any fixed $i \in [n] $, $S \subset [n]$, $t>0$ and $\delta >0$
$$\bbP \left( \langle \bw_i, \dfrac{1}{|S|}\sum_{j\in S} \bw_j \rangle \geq \dfrac{3\sigma^2(t\sqrt{|S|} + r + \log(1/\delta))}{|S|}\right) \leq \exp \left (-\min\left\{\dfrac{t^2}{4r},\dfrac{t}{4}\right \} \right) + \delta.$$
\end{lemma}

\begin{lemma}[Cluster sum bound]
\label{lem:cluster-sum}
For all $h\in [K]$,
$$\|\Wb_{\cC_h} \|_2 \leq 3\sigma\sqrt{(r+\log n)|\cC_h|}, $$
with prob greater than $1-n^{-3}$. 
\end{lemma}
        
\begin{lemma}[Indicator sum bound]
\label{lem:indicator-bernstein}
Let $g,h \in [K]$ such that $g\neq h$. Then, for any $a>0$,  
\begin{align*}
& \sum_{i\in \cC_g} \cI\left( a\|\bmu_g - \bmu_h\|^2  \leq
\langle \bw_i, \bmu_h - \bmu_g\rangle \right) \\
\leq & n_g\exp\left(-\frac{a^2\Delta^2}{2\sigma^2}\right) + 
\max\left\{ \frac{16}{3}\log(n),
4\exp\left( -\frac{a^2\Delta^2}{4\sigma^2} \right)\sqrt{n_g\log(n)}
\right\},
\end{align*}
with probability at least $1 - n^{-4}$.
\end{lemma}

Note that Lemma \ref{lem:indicator-bernstein} is an immediate consequence of Bernstein's inequality using the fact that indicators are independent and identically distributed Bernoulli random variables. We will also use the following immediate consequence of the Cauchy-Schwarz (CS) inequality. For all $a_1,\ldots,a_n >0 \in \Rb$, 
\begin{equation}\label{Lemma CS}
\sum_{i=1}^n\sqrt{a_i} \leq \sqrt{n\sum_{i=1}^n a_i }.
\end{equation}

\subsubsection{One-step bounds for Lloyd's iterates}
\label{subsec:one-step}

\paragraph{Lemma on center error recursion}
We will first control $G_s$ and $\Gamma_s$. This will then allow us to control $A_s$ as $s$ grows. 

\begin{lemma}[Center error recursion]
\label{lem:Gamma-step}

Conditionally on the events that the results of Lemmas \ref{lem:subg-sum}, \ref{lem:subg-quad}, and \ref{lem:cluster-sum} hold, assume $G_s \leq \frac{1}{2}$, then
\begin{equation}
\Gamma_s \;\le\; 
\frac{\epsilon}{\Delta_{\min}}
\;+\;
\min\Biggl\{
  2G_s \Gamma_{s-1} 
  + \frac{2\sqrt{6}}{\rho_{\sigma}}\sqrt{K G_s}
  + \frac{6\sigma}{\Delta_{\min}}\sqrt{\frac{r+\log n}{\alpha n}},
  \;
  \frac{\sqrt{6}}{\rho_{\sigma}} 
  + \frac{\Delta_{\max}}{\Delta_{\min}}\, G_s
\Biggr\}.
\end{equation}
\end{lemma}

\begin{proof}
We condition throughout on the event that the conclusions of 
Lemmas~\ref{lem:subg-sum}, \ref{lem:subg-quad}, and~\ref{lem:cluster-sum} hold.

\paragraph{Step 1: Decomposition.}
Write
\[
\wh\bmu^{(s)}_h - \bmu_h
  = \frac{1}{\wh n_h^{(s)}} 
    \sum_{i\in U^{(s)}_{hh}} (\by_i - \bmu_h)
  \;+\;
    \sum_{g\neq h}
      \frac{\wh n_{gh}^{(s)}}{\wh n_h^{(s)}}
      \bigl( \wb\by_{U^{(s)}_{gh}} - \bmu_h \bigr).
\label{eq:decomp}
\]

\paragraph{Step 2: Mean of misassigned points.}
For each $i\in U^{(s)}_{gh}$, Lloyd's update rule implies
\[
\|\by_i - \wh\bmu^{(s-1)}_h\|^2 
   \le 
\|\by_i - \wh\bmu^{(s-1)}_g\|^2.
\]
Let $a = \wh\bmu^{(s-1)}_h$ and $b = \wh\bmu^{(s-1)}_g$. Expanding the
squares and cancelling $\|\by_i\|^2$ gives
\[
2(b-a)^\top \by_i \;\le\; \|b\|^2 - \|a\|^2.
\]
Averaging over $i\in U^{(s)}_{gh}$ and writing
$\wb\by_{U^{(s)}_{gh}} = \frac{1}{\wh n^{(s)}_{gh}}\sum_{i\in U^{(s)}_{gh}}\by_i$
yields
\[
2(b-a)^\top \wb\by_{U^{(s)}_{gh}} \;\le\; \|b\|^2 - \|a\|^2,
\]
which is equivalent to
\[
\|\wb\by_{U^{(s)}_{gh}} - \wh\bmu^{(s-1)}_h\|
   \le
\|\wb\by_{U^{(s)}_{gh}} - \wh\bmu^{(s-1)}_g\|.
\]

Next, insert the true center $\bmu_g$ and the clean mean
$\wb\by^*_{U^{(s)}_{gh}}
 = \frac{1}{\wh n^{(s)}_{gh}}\sum_{i\in U^{(s)}_{gh}}\by_i^*$ and apply
the triangle inequality:
\begin{align*}
\|\wb\by_{U^{(s)}_{gh}} - \bmu_h\|
&\le \|\wb\by_{U^{(s)}_{gh}} - \wh\bmu^{(s-1)}_h\|
    + \|\wh\bmu^{(s-1)}_h - \bmu_h\| \\
&\le \|\wb\by_{U^{(s)}_{gh}} - \wh\bmu^{(s-1)}_g\|
    + \|\wh\bmu^{(s-1)}_h - \bmu_h\| \\
&\le \|\wb\by_{U^{(s)}_{gh}} - \wb\by^*_{U^{(s)}_{gh}}\|
   + \|\wb\by^*_{U^{(s)}_{gh}} - \bmu_g\|
   + \|\bmu_g - \wh\bmu^{(s-1)}_g\|
   + \|\wh\bmu^{(s-1)}_h - \bmu_h\|.
\end{align*}
By the model error bound, 
$\|\wb\by_{U^{(s)}_{gh}} - \wb\by^*_{U^{(s)}_{gh}}\|\le \epsilon$; by
Lemma~\ref{lem:subg-sum},
$\|\wb\by^*_{U^{(s)}_{gh}} - \bmu_g\|\le \sigma\sqrt{3(n+r)/\wh n^{(s)}_{gh}}$;
and by the definition of $\Gamma_{s-1}$,
$\|\bmu_g - \wh\bmu^{(s-1)}_g\|,
 \|\wh\bmu^{(s-1)}_h - \bmu_h\|
 \le \Gamma_{s-1}\Delta_{\min}$.
Hence
\[
\|\wb\by_{U^{(s)}_{gh}} - \bmu_h\|
\le
\epsilon
+ \sigma\sqrt{\frac{3(n+r)}{\wh n^{(s)}_{gh}}}
+ 2\Gamma_{s-1}\Delta_{\min}.
\]

Therefore,
\begin{align}
\biggl\|
\sum_{g\neq h}
  \frac{\wh n_{gh}^{(s)}}{\wh n_h^{(s)}}
  \bigl(\wb\by_{U^{(s)}_{gh}} - \bmu_h\bigr)
\biggr\|
&\le
\sum_{g\neq h}
  \frac{\wh n_{gh}^{(s)}}{\wh n_h^{(s)}}
  \Biggl(
    \epsilon
    + \sigma\sqrt{\frac{3(n+r)}{\wh n^{(s)}_{gh}}}
    + 2\Gamma_{s-1}\Delta_{\min}
  \Biggr) \nonumber \\
&\le
\epsilon
+ \sigma\sqrt{
    \frac{3K(n+r)}{\wh n_h^{(s)}}\,G_s
  }
+ 2G_s \Gamma_{s-1}\Delta_{\min},
\label{eq:misassign_bound}
\end{align}
where we used Cauchy--Schwarz and the definition of $G_s$ in the last
inequality.

\paragraph{Step 3: Mean of correctly assigned points.}

Write
\[
\frac{1}{\wh n_h^{(s)}} 
\sum_{i\in U^{(s)}_{hh}} (\by_i - \bmu_h)
  = \frac{\wh n^{(s)}_{hh}}{\wh n_h^{(s)}} \epsilon
  + \frac{1}{\wh n_h^{(s)}}
    \Biggl(
      \sum_{i:z_i=h}(\by_i^*-\bmu_h)
      - 
      \sum_{g\neq h}
        \sum_{i\in U^{(s)}_{hg}}
           (\by_i^*-\bmu_h)
    \Biggr).
\]
By Lemma~\ref{lem:cluster-sum},
\(
\big\|\sum_{i:z_i=h}(\by_i^*-\bmu_h)\big\|
\le 3\sigma\sqrt{r+\log n}\,\sqrt{n_h}
\),
and by Lemma~\ref{lem:subg-sum},
\(
\big\|\sum_{i\in U^{(s)}_{hg}}(\by_i^*-\bmu_h)\big\|
\le \sigma\sqrt{3(n+r)}\,\sqrt{|U^{(s)}_{hg}|}
\).
Using $n_h - \wh n_{hh}^{(s)} = \sum_{g\neq h} |U^{(s)}_{hg}|$ and the triangle inequality, we obtain
\[
\biggl\|
\frac{1}{\wh n_h^{(s)}} 
\sum_{i\in U^{(s)}_{hh}} (\by_i - \bmu_h)
\biggr\|
\le
\epsilon
+ 3\sigma\frac{\sqrt{r+\log n}\,\sqrt{n_h}}{\wh n_h^{(s)}}
+ \sigma\sqrt{\frac{3(n+r)}{\wh n_h^{(s)}}}\,
  \sqrt{\frac{n_h - \wh n^{(s)}_{hh}}{n_h}}.
\label{eq:correct_bound}
\]

\paragraph{Step 4: Combination.}
From $G_s \le \tfrac12$,
\begin{align}
\wh n_h^{(s)} &\ge \frac{n_h}{2} \ge \frac{\alpha n}{2},
&
\frac{1}{\sqrt{\wh n_h^{(s)}}}
  &\le \sqrt{\frac{2}{\alpha n}},
\label{eq:Gs_half}\\
\frac{\sqrt{n_h}}{\wh n_h^{(s)}}
  &\le \frac{2}{\sqrt{n_h}}
  \le \frac{2}{\sqrt{\alpha n}},
&
\frac{\sqrt{n_h-\wh n_{hh}^{(s)}}}{\sqrt{n_h}}
  &\le \sqrt{G_s}.
\end{align}

Combining \eqref{eq:misassign_bound} and \eqref{eq:correct_bound} and using \eqref{eq:Gs_half},
\begin{align*}
\|\wh\bmu^{(s)}_h - \bmu_h\|
&\le
2\sigma\sqrt{\frac{6K(n+r)}{\alpha n}}\,\sqrt{G_s}
 + 2G_s\Gamma_{s-1}\Delta_{\min}
 + \epsilon
 + 6\sigma\sqrt{\frac{r+\log n}{\alpha n}}
\\[1mm]
&=
\Delta_{\min}\Biggl(
\frac{\epsilon}{\Delta_{\min}}
 + 2G_s\Gamma_{s-1}
 + \frac{2\sqrt{6}}{\rho_{\sigma}}\sqrt{K G_s}
\Biggr)
+ 6\sigma\sqrt{\frac{r+\log n}{\alpha n}}.
\end{align*}
Taking the supremum over $h$ and dividing by $\Delta_{\min}$ gives
\[
\Gamma_s
\le
\frac{\epsilon}{\Delta_{\min}}
 + 2G_s\Gamma_{s-1}
 + \frac{2\sqrt{6}}{\rho_{\sigma}}\sqrt{K G_s}
 + \frac{6\sigma}{\Delta_{\min}}\sqrt{\frac{r+\log n}{\alpha n}}.
\label{eq:Gamma_first}
\]

\paragraph{Step 5: Alternative bound via mixture structure.}
We now derive a second, non-recursive bound. From
\[
\wh\bmu^{(s)}_h
  = \sum_{g}\frac{\wh n_{gh}^{(s)}}{\wh n_h^{(s)}}\bmu_g
    + \frac{1}{\wh n_h^{(s)}}
      \sum_{\wh z_i^{(s)}=h} (\by_i - \bmu_{z_i}),
\]
we decompose
\[
\wh\bmu^{(s)}_h - \bmu_h
  = \sum_{g\neq h}\frac{\wh n_{gh}^{(s)}}{\wh n_h^{(s)}}(\bmu_g - \bmu_h)
    + \frac{1}{\wh n_h^{(s)}}
      \sum_{\wh z_i^{(s)}=h} (\by_i - \bmu_{z_i}).
\]
For the first term, by the triangle inequality and the definition of $G_s$,
\[
\biggl\|
\sum_{g\neq h}\frac{\wh n_{gh}^{(s)}}{\wh n_h^{(s)}}(\bmu_g - \bmu_h)
\biggr\|
\le
\sum_{g\neq h}\frac{\wh n_{gh}^{(s)}}{\wh n_h^{(s)}}
   \|\bmu_g - \bmu_h\|
\le
\Delta_{\max} G_s.
\]

For the second term, write
\[
\by_i - \bmu_{z_i}
 = (\by_i - \by_i^*) + (\by_i^* - \bmu_{z_i}),
\]
so that
\[
\frac{1}{\wh n_h^{(s)}}
      \sum_{\wh z_i^{(s)}=h} (\by_i - \bmu_{z_i})
=
\frac{1}{\wh n_h^{(s)}}
      \sum_{\wh z_i^{(s)}=h} (\by_i - \by_i^*)
+
\frac{1}{\wh n_h^{(s)}}
      \sum_{\wh z_i^{(s)}=h} (\by_i^* - \bmu_{z_i}).
\]
By the model error bound \eqref{eqn:model},
\(
\big\|\frac{1}{\wh n_h^{(s)}}\sum_{\wh z_i^{(s)}=h} (\by_i - \by_i^*)\big\|
\le \epsilon
\),
and by Lemma~\ref{lem:subg-sum},
\(
\big\|\frac{1}{\wh n_h^{(s)}}\sum_{\wh z_i^{(s)}=h} (\by_i^* - \bmu_{z_i})\big\|
\le \sigma\sqrt{3(n+r)/\wh n_h^{(s)}}.
\)
Combining these bounds, we obtain
\[
\|\wh\bmu^{(s)}_h - \bmu_h\|
\le
\Delta_{\max} G_s
 + \sigma\sqrt{\frac{3(n+r)}{\wh n_h^{(s)}}}
 + \epsilon.
\]
Since $\wh n^{(s)}_h \ge \alpha n/2$, we have
\[
\sigma\sqrt{\frac{3(n+r)}{\wh n_h^{(s)}}}
\le \sigma\sqrt{\frac{6(n+r)}{\alpha n}}
= \Delta_{\min}\,\frac{\sqrt{6}}{\rho_\sigma},
\]
and therefore
\[
\Gamma_s
\le
\frac{\epsilon}{\Delta_{\min}}
 + \frac{\sqrt{6}}{\rho_{\sigma}}
 + \frac{\Delta_{\max}}{\Delta_{\min}}\,G_s.
\label{eq:Gamma_second}
\]

\paragraph{Step 6: Final conclusion.}
Combining \eqref{eq:Gamma_first} and \eqref{eq:Gamma_second} proves the lemma.
\end{proof}

\paragraph{Lemma on misclustering recursion}
\begin{lemma}[Misclustering recursion]
\label{lem:G-step}
Conditionally on the event that Lemmas~\ref{lem:subg-sum},
\ref{lem:subg-quad}, \ref{lem:cluster-sum}, and
\ref{lem:indicator-bernstein} hold, assume
$\Gamma_s \leq \frac{1-C_{\Gamma}}{2}$ for some $C_\Gamma\in(0,1)$ and
$G_s < \frac{1}{2}$. Let $\beta_{1,\sigma}$ be as in 
Section~\ref{subsec:notation} and suppose $\beta_{1,\sigma}>0$.
Then
\[
G_{s+1} \le \frac{32}{3}\frac{K\log n}{n\alpha}
 + \frac{4\sigma^2}{\alpha\beta_{1,\sigma}^2\Delta_{\min}^2}
   \Bigl(1 + 8\sqrt{\frac{K\log n}{n}}\Bigr) 
 + \frac{8\Gamma_s^2}{\rho_{\epsilon}} 
 + \frac{48\Gamma_s^2}{\rho_{\sigma}}.
\]
\end{lemma}

\begin{proof}

\step{1: Rewriting the misclassification indicator}

To control $G_{s+1}$, we must bound
\[
\widehat n_{gh}^{(s+1)} 
:= \sum_{i=1}^n \mathbb I\{z_i = g,\; \widehat z_i^{(s+1)} = h\},
\qquad g,h\in[K],\ g\neq h,
\]
that is, the number of points from cluster $g$ assigned to center $h$ at
iteration $s+1$.

Fix $g\in[K]$ and $h\neq g$. By Lloyd's update rule,
\[
\{z_i = g,\ \widehat z_i^{(s+1)} = h\}
\subseteq
\bigl\{\|\by_i - \wh\bmu_h^{(s)}\|^2 \le \|\by_i - \wh\bmu_g^{(s)}\|^2\bigr\},
\]
so
\[
\mathbb I\{z_i = g,\ \widehat z_i^{(s+1)} = h\}
\;\le\;
\mathbb I\bigl\{\|\by_i - \wh\bmu_h^{(s)}\|^2 
               \le \|\by_i - \wh\bmu_g^{(s)}\|^2\bigr\}.
\]

Writing $\by_i = \bmu_g + (\by_i - \bmu_g)$ and expanding both sides around
$\bmu_g$ yields
\begin{align*}
\|\by_i - \wh\bmu_h^{(s)}\|^2
&= \|\by_i - \bmu_g + \bmu_g - \wh\bmu_h^{(s)}\|^2 \\
&= \|\by_i - \bmu_g\|^2
   + \|\bmu_g - \wh\bmu_h^{(s)}\|^2
   + 2\langle \by_i - \bmu_g,\;\bmu_g - \wh\bmu_h^{(s)}\rangle,
\\
\|\by_i - \wh\bmu_g^{(s)}\|^2
&= \|\by_i - \bmu_g + \bmu_g - \wh\bmu_g^{(s)}\|^2 \\
&= \|\by_i - \bmu_g\|^2
   + \|\bmu_g - \wh\bmu_g^{(s)}\|^2
   + 2\langle \by_i - \bmu_g,\;\bmu_g - \wh\bmu_g^{(s)}\rangle.
\end{align*}
Cancelling the common term $\|\by_i - \bmu_g\|^2$ gives
\[
\|\by_i - \wh\bmu_h^{(s)}\|^2 \le \|\by_i - \wh\bmu_g^{(s)}\|^2
\]
if and only if
\[
\|\bmu_g - \wh\bmu_h^{(s)}\|^2 - \|\bmu_g - \wh\bmu_g^{(s)}\|^2
\;\le\;
2\langle \by_i - \bmu_g,\;\wh\bmu_h^{(s)} - \wh\bmu_g^{(s)}\rangle.
\]
Hence
\begin{equation}
\label{eq:indicator-rewritten}
\mathbb I\{z_i = g,\ \widehat z_i^{(s+1)} = h\}
\le
\mathbb I\Bigl\{
\|\bmu_g - \wh\bmu_h^{(s)}\|^2 - \|\bmu_g - \wh\bmu_g^{(s)}\|^2
\le 2\langle \by_i - \bmu_g,\;\wh\bmu_h^{(s)} - \wh\bmu_g^{(s)}\rangle
\Bigr\}.
\end{equation}

Next we lower bound the left-hand side of \eqref{eq:indicator-rewritten}.
By the triangle inequality,
\[
\|\bmu_g - \wh\bmu_h^{(s)}\|
\ge \|\bmu_g - \bmu_h\| - \|\bmu_h - \wh\bmu_h^{(s)}\|.
\]
By the definition of $\Gamma_s$ in \eqref{Gamma_s} and the fact that
$\Delta_{\min} \le \|\bmu_g - \bmu_h\|$, we have
\[
\|\bmu_h - \wh\bmu_h^{(s)}\|
\le \Gamma_s \Delta_{\min}
\le \Gamma_s \|\bmu_g - \bmu_h\|,
\]
so
\[
\|\bmu_g - \wh\bmu_h^{(s)}\|
\ge (1-\Gamma_s)\|\bmu_g - \bmu_h\|.
\]
Therefore
\begin{align}
\|\bmu_g - \wh\bmu_h^{(s)}\|^2 - \|\bmu_g - \wh\bmu_g^{(s)}\|^2
&\ge (1-\Gamma_s)^2\|\bmu_g - \bmu_h\|^2
      - \Gamma_s^2 \|\bmu_g - \bmu_h\|^2 \nonumber\\
&= (1 - 2\Gamma_s)\|\bmu_g - \bmu_h\|^2.
\label{eq:1-2Gamma-lower}
\end{align}
Under the assumption $\Gamma_s \le (1-C_\Gamma)/2$, we have
$1-2\Gamma_s \ge C_\Gamma$, and hence
\begin{equation}
\label{eq:Cgamma-lower}
\|\bmu_g - \wh\bmu_h^{(s)}\|^2 - \|\bmu_g - \wh\bmu_g^{(s)}\|^2
\;\ge\; C_\Gamma \|\bmu_g - \bmu_h\|^2.
\end{equation}
Combining \eqref{eq:indicator-rewritten} and \eqref{eq:Cgamma-lower}, we obtain
\[
\mathbb I\{z_i = g,\ \widehat z_i^{(s+1)} = h\}
\le
\mathbb I\Bigl\{
C_\Gamma \|\bmu_g - \bmu_h\|^2
\le
2\langle \by_i - \bmu_g,\;\wh\bmu_h^{(s)} - \wh\bmu_g^{(s)}\rangle
\Bigr\}.
\]

and
\begin{align*}
\|\bmu_g - \wh\bmu\s_h\|^2 &\geq (\|\bmu_g - \bmu_h\| - \|\bmu_h - \wh\bmu\s_h\|)^2 \\
& = \bigg ( \|\bmu_g - \bmu_h\| \bigg ( 1 - \dfrac{\|\bmu_h - \wh\bmu\s_h\|}{\|\bmu_g - \bmu_h\|} \bigg ) \bigg )^2 \\
& \geq \|\bmu_g - \bmu_h\|^2 (1-\Gamma_s)^2, 
\end{align*}
by the definition \eqref{Gamma_s} of $\Gamma_s$. So, 
\begin{align*}
\|\bmu_g - \wh\bmu\s_h\|^2 - \| \bmu_g - \wh\bmu\s_g\|^2 & \geq \|\bmu_g - \bmu_h\|^2 (1-\Gamma_s)^2 - \| \bmu_g - \wh\bmu\s_g\|^2 \nonumber\\ 
&\geq  \|\bmu_g - \bmu_h\|^2 ((1-\Gamma_s)^2  - \Gamma_s^2) \nonumber \\
& = \|\bmu_g - \bmu_h\|^2 (1-2\Gamma_s) \label{1-2Gamma bdd}\\
&  \geq \|\bmu_g - \bmu_h\|^2 C_{\Gamma}, \nonumber
\end{align*}
where the last inequality is obtained using the assumption that $\Gamma_s \leq \frac{1-C_{\Gamma}}{2}$. 

\step{2: Decomposition into four terms}
Recall inequality \eqref{eq:indicator-rewritten} from Step~1 and the
lower bound \eqref{eq:Cgamma-lower}, which together give
\begin{equation}
\label{eq:Gstep-start}
\mathbb I\{z_i = g,\ \widehat z_i^{(s+1)} = h\}
\le
\mathbb I\Bigl\{
C_\Gamma \|\bmu_g - \bmu_h\|^2
\le
2\langle \by_i - \bmu_g,\;\wh\bmu_h^{(s)} - \wh\bmu_g^{(s)}\rangle
\Bigr\}.
\end{equation}

We now expand both the data and the centers.  Write
\[
\by_i - \bmu_g
 = (\by_i - \by_i^*) + (\by_i^* - \bmu_g)
 =: \be_i + \bw_i,
\]
where $\be_i := \by_i - \by_i^*$ is the model error and
$\bw_i := \by_i^* - \bmu_{z_i}$ is sub-Gaussian noise.  For the centers,
set
\[
\bzeta_k := \wh\bmu_k^{(s)} - \bmu_k,\qquad k\in[K],
\]
so that
\[
\wh\bmu_h^{(s)} - \wh\bmu_g^{(s)}
 = (\bmu_h - \bmu_g) + (\bzeta_h - \bzeta_g).
\]
Substituting these decompositions into the right-hand side of
\eqref{eq:Gstep-start} yields
\begin{align*}
\langle \by_i - \bmu_g,\;\wh\bmu_h^{(s)} - \wh\bmu_g^{(s)}\rangle
&= \langle \be_i + \bw_i,\;\bmu_h - \bmu_g + \bzeta_h - \bzeta_g\rangle \\
&= \langle \be_i,\bmu_h - \bmu_g\rangle
 + \langle \bw_i,\bmu_h - \bmu_g\rangle \\
&\quad
 + \langle \be_i,\bzeta_h - \bzeta_g\rangle
 + \langle \bw_i,\bzeta_h - \bzeta_g\rangle.
\end{align*}

We now split the left-hand side of \eqref{eq:Gstep-start} into four
pieces.  Introduce positive constants
\begin{equation}\label{eqn:def_beta}
    \beta_{1,\epsilon} = c_{\epsilon}\frac{\epsilon}{\Delta_{\min}},\qquad
\beta_{2,\sigma} = \sqrt{\frac{n+Kr}{n\alpha}}\,
                   \frac{\sigma}{\Delta_{\min}} = \frac{1}{\rho_\sigma},
\qquad
\beta_{2,\epsilon} = \frac{1}{\sqrt{\alpha}}\,
                     \frac{\epsilon}{\Delta_{\min}} = \frac{1}{\rho_\epsilon},
\end{equation}
where $c_{\epsilon}>1$ and recall that $\beta_{1,\sigma}$ is defined in
Section~\ref{subsec:notation} so that
\begin{equation}
\label{eq:beta-split}
\beta_{1,\epsilon} + \beta_{1,\sigma} + \beta_{2,\epsilon} + \beta_{2,\sigma}
= \frac{C_\Gamma}{2},\qquad
\beta_{1,\sigma} > 0.
\end{equation}
Using \eqref{eq:Gstep-start}, \eqref{eq:beta-split}, and the
decomposition above, we obtain
\begin{align}
\mathbb I\{z_i = g,\ \widehat z_i^{(s+1)} = h\}
&\le
\mathbb I\Bigl\{
\beta_{1,\epsilon}\|\bmu_g - \bmu_h\|^2
 + \beta_{1,\sigma}\|\bmu_g - \bmu_h\|^2 \nonumber\\
&\hspace{2.5cm}
 + \beta_{2,\epsilon}\|\bmu_g - \bmu_h\|^2
 + \beta_{2,\sigma}\|\bmu_g - \bmu_h\|^2 \nonumber\\
&\hspace{1.2cm}
\le
2\langle \be_i,\bmu_h - \bmu_g\rangle
 + 2\langle \bw_i,\bmu_h - \bmu_g\rangle \nonumber\\
&\hspace{2.1cm}
 + 2\langle \be_i,\bzeta_h - \bzeta_g\rangle
 + 2\langle \bw_i,\bzeta_h - \bzeta_g\rangle
\Bigr\}.\label{eq:Gstep-4sum}
\end{align}
By a union bound applied to \eqref{eq:Gstep-4sum}, this event is
contained in the union of the following four events:
\begin{align*}
T_{1i}
&:= \mathbb I\Bigl(
\beta_{1,\epsilon}\|\bmu_g - \bmu_h\|^2
 \le \langle \be_i,\bmu_h - \bmu_g\rangle
\Bigr),\\
T_{2i}
&:= \mathbb I\Bigl(
\beta_{1,\sigma}\|\bmu_g - \bmu_h\|^2
 \le \langle \bw_i,\bmu_h - \bmu_g\rangle
\Bigr),\\
T_{3i}
&:= \mathbb I\Bigl(
\beta_{2,\epsilon}\|\bmu_g - \bmu_h\|^2
 \le \langle \be_i,\bzeta_h - \bzeta_g\rangle
\Bigr),\\
T_{4i}
&:= \mathbb I\Bigl(
\beta_{2,\sigma}\|\bmu_g - \bmu_h\|^2
 \le \langle \bw_i,\bzeta_h - \bzeta_g\rangle
\Bigr).
\end{align*}
Therefore
\begin{equation}
\label{eq:T1-T4}
\mathbb I\{z_i = g,\ \widehat z_i^{(s+1)} = h\}
\le T_{1i} + T_{2i} + T_{3i} + T_{4i},
\end{equation}
which is the desired decomposition used in the subsequent steps.

\step{3: Bound for $T_{1i}$}

Recall from \eqref{eq:T1-T4} that
\[
T_{1i}
:= \mathbb I\Bigl(
\beta_{1,\epsilon}\|\bmu_g - \bmu_h\|^2
 \le \langle \be_i,\bmu_h - \bmu_g\rangle
\Bigr),
\qquad \be_i = \by_i - \by_i^*.
\]
By Lemma~\ref{lem:indicator-bernstein}, we have, on the event under
consideration,
\[
\bigl|\langle \be_i,\bmu_h - \bmu_g\rangle\bigr|
\le \epsilon \|\bmu_h - \bmu_g\|.
\]
In the definition of $\beta_{1,\epsilon}$ we fixed a constant
$c_\epsilon>1$ such that
\[
\beta_{1,\epsilon}\Delta_{\min}
= c_\epsilon\,\epsilon > \epsilon.
\]
Since $\|\bmu_g - \bmu_h\|\ge \Delta_{\min}$, it follows that
\[
\beta_{1,\epsilon}\|\bmu_g - \bmu_h\|
\ge \beta_{1,\epsilon}\Delta_{\min}
> \epsilon
\ge \bigl|\langle \be_i,\bmu_h - \bmu_g\rangle\bigr|.
\]
Therefore the event inside the indicator defining $T_{1i}$ cannot occur,
and hence
\[
T_{1i} = 0
\]
deterministically on the event where Lemma~\ref{lem:indicator-bernstein}
holds.

\step{4: Bound for $T_{2i}$}

Recall from \eqref{eq:T1-T4} that
\[
T_{2i}
:= \mathbb I\Bigl(
\beta_{1,\sigma}\|\bmu_g - \bmu_h\|^2
 \le \langle \bw_i,\bmu_h - \bmu_g\rangle
\Bigr),
\qquad \bw_i = \by_i^* - \bmu_{z_i}.
\]
Let $\cC_g := \{i \in [n] : z_i = g\}$ and $n_g := |\cC_g|$. On the event of
Lemma~\ref{lem:indicator-bernstein}, each $\bw_i$ is sub-Gaussian with
proxy variance $\sigma^2$, so we can apply that lemma (with
$\bv = \bmu_h - \bmu_g$ and threshold
$\beta_{1,\sigma}\|\bmu_g - \bmu_h\|^2$), together with a union bound
over $i \in \cC_g$, to obtain
\begin{align*}
\sum_{i \in \cC_g} T_{2i} 
&\leq n_g\exp\!\left(-\frac{\beta_{1,\sigma}^2\Delta_{\min}^2}{2\sigma^2}\right) + 
\max\left\{ \frac{16}{3}\log n,\,
4\exp\!\left( -\frac{\beta_{1,\sigma}^2\Delta_{\min}^2}{4\sigma^2} \right)
   \sqrt{n_g\log n}
\right\}.
\end{align*}
This is the bound we will use for the contribution of $T_{2i}$ in
Step~6.

\step{5: Bounds for $T_{3i}$ and $T_{4i}$}

Recall from \eqref{eq:T1-T4} that
\[
T_{3i}
:= \mathbb I\Bigl(
\beta_{2,\epsilon}\|\bmu_g - \bmu_h\|^2
 \le \langle \be_i,\bzeta_h - \bzeta_g\rangle
\Bigr),
\qquad
T_{4i}
:= \mathbb I\Bigl(
\beta_{2,\sigma}\|\bmu_g - \bmu_h\|^2
 \le \langle \bw_i,\bzeta_h - \bzeta_g\rangle
\Bigr),
\]
where $\be_i = \by_i - \by_i^*$, $\bw_i = \by_i^* - \bmu_{z_i}$, and
$\bzeta_k = \wh\bmu_k^{(s)} - \bmu_k$.

\medskip\noindent
\emph{Bound for $\sum_{i\in\cC_g} T_{3i}$.}
By the definition of $\epsilon$ in \eqref{eqn:model}, we have
\[
\frac{1}{n_g}\sum_{i\in\cC_g}
\bigl\langle \be_i, v\bigr\rangle^2
\le \epsilon^2 \|v\|^2
\qquad\text{for all }v\in\mathbb R^d.
\]
Therefore,
\begin{align}
\sum_{i\in\cC_g} T_{3i}
&= \sum_{i\in\cC_g}
   \mathbb I\Bigl(
   \beta_{2,\epsilon}\|\bmu_g - \bmu_h\|^2
   \le \langle \be_i,\bzeta_h - \bzeta_g\rangle
   \Bigr) \nonumber\\
&\le \sum_{i\in\cC_g}
   \mathbb I\Bigl(
   1 \le
   \frac{\langle \be_i,\bzeta_h - \bzeta_g\rangle^2}
        {\beta_{2,\epsilon}^2\|\bmu_g - \bmu_h\|^4}
   \Bigr) \nonumber\\
&\le \frac{1}{\beta_{2,\epsilon}^2\Delta_{\min}^4}
    \sum_{i\in\cC_g}
    \bigl\langle \be_i,\bzeta_h - \bzeta_g\bigr\rangle^2
    \nonumber\\
&\le \frac{\epsilon^2\|\bzeta_h - \bzeta_g\|^2}{\beta_{2,\epsilon}^2\Delta_{\min}^4}\,n_g.
\label{eq:T3-pre}
\end{align}
Using
\begin{equation}
\label{eq:zetadiff-gamma}
\|\bzeta_h - \bzeta_g\|
\le \|\bzeta_h\| + \|\bzeta_g\|
\le \|\wh\bmu_h^{(s)} - \bmu_h\|
   + \|\wh\bmu_g^{(s)} - \bmu_g\|
\le 2\Gamma_s\Delta_{\min}
\le 2\Gamma_s\|\bmu_h - \bmu_g\|,
\end{equation}
we obtain from \eqref{eq:T3-pre} that
\begin{align}
\sum_{i\in\cC_g} T_{3i}
&\le \frac{4\epsilon^2\Gamma_s^2}{\beta_{2,\epsilon}^2\Delta_{\min}^2}\,n_g
\le \frac{4\sqrt{\alpha}\,\epsilon\,\Gamma_s^2}{\Delta_{\min}}\,n_g,
\end{align}
where in the last inequality we used the definition of $\beta_{2,\epsilon}$
in \eqref{eqn:def_beta}.

\medskip\noindent
\emph{Bound for $\sum_{i\in\cC_g} T_{4i}$.}
Similarly,
\begin{align}
\sum_{i\in\cC_g} T_{4i}
&= \sum_{i\in\cC_g}
   \mathbb I\Bigl(
   \beta_{2,\sigma}\|\bmu_g - \bmu_h\|^2
   \le \langle \bw_i,\bzeta_h - \bzeta_g\rangle
   \Bigr) \nonumber\\
&\le \sum_{i\in\cC_g}
   \mathbb I\Bigl(
   1 \le
   \frac{\langle \bw_i,\bzeta_h - \bzeta_g\rangle^2}
        {\beta_{2,\sigma}^2\|\bmu_g - \bmu_h\|^4}
   \Bigr) \nonumber\\
&\le \frac{1}{\beta_{2,\sigma}^2\Delta_{\min}^4}
    \sum_{i\in\cC_g}
    \bigl\langle \bw_i,\bzeta_h - \bzeta_g\bigr\rangle^2 \nonumber\\
&\le \frac{\|\bzeta_h - \bzeta_g\|^2}{\beta_{2,\sigma}^2\Delta_{\min}^4}\,
    \lambda_{\max}\!\left(\sum_{i\in\cC_g}\bw_i\bw_i^\top\right).
\label{eq:T4-pre}
\end{align}
By Lemma~\ref{lem:subg-quad}, we have
\[
\lambda_{\max}\!\left(\sum_{i\in\cC_g}\bw_i\bw_i^\top\right)
\le 6\sigma^2(n_g + r),
\]
and combining this with \eqref{eq:zetadiff-gamma} and \eqref{eq:T4-pre}
gives
\begin{align}
\sum_{i\in\cC_g} T_{4i}
&\le \frac{6\sigma^2\|\bzeta_h - \bzeta_g\|^2}{\beta_{2,\sigma}^2\Delta_{\min}^4}
    (n_g + r) \nonumber\\
&\le \frac{24\sigma^2\Gamma_s^2}{\beta_{2,\sigma}^2\Delta_{\min}^2}
    (n_g + r)
\le 24\sqrt{\frac{n\alpha}{n+Kr}}\,
   \frac{\sigma\Gamma_s^2}{\Delta_{\min}}\,(n_g + r),
\end{align}
where in the last inequality we used the definition of $\beta_{2,\sigma}$
in \eqref{eqn:def_beta}.

\step{6: Combining the bounds}

Recall that
\[
\widehat n_{gh}^{(s+1)}
:= \sum_{i=1}^n \mathbb I\{z_i = g,\; \widehat z_i^{(s+1)} = h\},
\qquad g,h\in[K],\ g\neq h,
\]
and that by \eqref{eq:T1-T4},
\[
\mathbb I\{z_i = g,\ \widehat z_i^{(s+1)} = h\}
\le T_{1i} + T_{2i} + T_{3i} + T_{4i}.
\]
Let $\cC_g := \{i\in[n]: z_i = g\}$ and $n_g := |\cC_g|$. Then
\[
\widehat n_{gh}^{(s+1)}
\le \sum_{i\in\cC_g} (T_{1i} + T_{2i} + T_{3i} + T_{4i}).
\]

By definition,
\[
G_{s+1}
= \max\Biggl\{
\max_{g\in[K]}\sum_{h\neq g}\frac{\widehat n_{gh}^{(s+1)}}{n_g},\quad
\max_{h\in[K]}\sum_{g\neq h}\frac{\widehat n_{gh}^{(s+1)}}
                               {\widehat n_h^{(s+1)}}
\Biggr\},
\]
where $\widehat n_h^{(s+1)} := \sum_{g=1}^K \widehat n_{gh}^{(s+1)}$.

\medskip\noindent
\emph{Forward error: normalization by $n_g$.}
Using the bounds from Steps~3–5, we obtain for each fixed $g$,
\begin{align*}
\sum_{h\neq g}\frac{\widehat n_{gh}^{(s+1)}}{n_g}
&\le \sum_{h\neq g}\sum_{i\in\cC_g}
   \frac{T_{1i} + T_{2i} + T_{3i} + T_{4i}}{n_g} \\
&\le \frac{16}{3}\frac{K\log n}{n_g}
    + K\exp\!\left(-\frac{\beta_{1,\sigma}^2\Delta_{\min}^2}{2\sigma^2}\right)
    + 4K\exp\!\left(-\frac{\beta_{1,\sigma}^2\Delta_{\min}^2}{4\sigma^2}\right)
      \sqrt{\frac{\log n}{n_g}}  \\
&\quad +
\frac{4K\sqrt{\alpha}\,\epsilon\,\Gamma_s^2}{\Delta_{\min}}
 + 24K\sqrt{\frac{n\alpha}{n+Kr}}\,
   \frac{\sigma\Gamma_s^2}{\Delta_{\min}}\frac{n_g + r}{n_g}.
\end{align*}
Using $n_g \ge \alpha n$ and $n_g^{-1}\le (\alpha n)^{-1}$, this yields
\begin{align}
\max_{g\in[K]}\sum_{h\neq g}\frac{\widehat n_{gh}^{(s+1)}}{n_g}
&\le \frac{16}{3}\frac{K\log n}{n\alpha}
 + K\exp\!\left(-\frac{\beta_{1,\sigma}^2\Delta_{\min}^2}{2\sigma^2}\right) \nonumber\\
&\quad
 + 4K\exp\!\left(-\frac{\beta_{1,\sigma}^2\Delta_{\min}^2}{4\sigma^2}\right)
      \sqrt{\frac{\log n}{n\alpha}} \nonumber\\
&\quad +
\frac{4K\sqrt{\alpha}\,\epsilon\,\Gamma_s^2}{\Delta_{\min}}
 + 24K\sqrt{\frac{n\alpha}{n+Kr}}\,
   \frac{\sigma\Gamma_s^2}{\Delta_{\min}}\Bigl(1 + \frac{r}{n\alpha}\Bigr).
\label{eq:forward-bound}
\end{align}

\medskip\noindent
\emph{Backward error: normalization by $\widehat n_h^{(s+1)}$.}
Similarly,
\begin{align*}
\sum_{g\neq h}\frac{\widehat n_{gh}^{(s+1)}}{\widehat n_h^{(s+1)}}
&\le \sum_{g\neq h}\sum_{i\in\cC_g}
   \frac{T_{1i} + T_{2i} + T_{3i} + T_{4i}}{\widehat n_h^{(s+1)}}.
\end{align*}
From the definition of $G_s$ and the assumption $G_s < 1/2$,
one checks that
\[
\frac{1}{n_g} \le \frac{1}{\alpha n},\qquad
\frac{n_g}{\widehat n_h^{(s+1)}} \le \frac{2}{\alpha},\qquad
\frac{\sqrt{n_g}}{\widehat n_h^{(s+1)}} \le \frac{2}{\alpha\sqrt n},
\]
for all $g,h\in[K]$. Using these bounds together with the estimates from
Steps~3–5, we obtain
\begin{align}
\max_{h\in[K]}\sum_{g\neq h}\frac{\widehat n_{gh}^{(s+1)}}{\widehat n_h^{(s+1)}}
&\le \frac{32}{3}\frac{K\log n}{n\alpha}
 + \frac{2}{\alpha}
   \exp\!\left(-\frac{\beta_{1,\sigma}^2\Delta_{\min}^2}{2\sigma^2}\right)
 \nonumber\\
&\quad
 + \frac{8}{\alpha}
   \sqrt{\frac{K\log n}{n}}\,
   \exp\!\left(-\frac{\beta_{1,\sigma}^2\Delta_{\min}^2}{4\sigma^2}\right)
 + \frac{8\epsilon\,\Gamma_s^2}{\sqrt{\alpha}\,\Delta_{\min}} \nonumber\\
&\quad
 + 48\frac{\sigma\Gamma_s^2}{\Delta_{\min}}
   \sqrt{\frac{n+Kr}{n\alpha}} \nonumber\\
&= \frac{32}{3}\frac{K\log n}{n\alpha}
 + \frac{2}{\alpha}
   \exp\!\left(-\frac{\beta_{1,\sigma}^2\Delta_{\min}^2}{2\sigma^2}\right)
 \nonumber\\
&\quad
 + \frac{8}{\alpha}
   \sqrt{\frac{K\log n}{n}}\,
   \exp\!\left(-\frac{\beta_{1,\sigma}^2\Delta_{\min}^2}{4\sigma^2}\right)
 + \frac{8\Gamma_s^2}{\rho_\epsilon}
 + \frac{48\Gamma_s^2}{\rho_\sigma},
\label{eq:backward-bound}
\end{align}
where, in the last equality, we used the definitions of
$\rho_\epsilon$ and $\rho_\sigma$.

\medskip

Combining \eqref{eq:forward-bound} and \eqref{eq:backward-bound}, we obtain
\begin{align*}
G_{s+1}
&\le \frac{32}{3}\frac{K\log n}{n\alpha}
 + \frac{2}{\alpha}
   \exp\!\left(-\frac{\beta_{1,\sigma}^2\Delta_{\min}^2}{2\sigma^2}\right) \\
&\quad
 + \frac{8}{\alpha}
   \sqrt{\frac{K\log n}{n}}\,
   \exp\!\left(-\frac{\beta_{1,\sigma}^2\Delta_{\min}^2}{4\sigma^2}\right)
 + \frac{8\Gamma_s^2}{\rho_\epsilon}
 + \frac{48\Gamma_s^2}{\rho_\sigma}.
\end{align*}
Finally, using the elementary bound $\exp(-x)\le x^{-1}$ for $x>0$, we
obtain
\[
\exp\!\left(-\frac{\beta_{1,\sigma}^2\Delta_{\min}^2}{2\sigma^2}\right)
\le \frac{2\sigma^2}{\beta_{1,\sigma}^2\Delta_{\min}^2},\qquad
\exp\!\left(-\frac{\beta_{1,\sigma}^2\Delta_{\min}^2}{4\sigma^2}\right)
\le \frac{4\sigma^2}{\beta_{1,\sigma}^2\Delta_{\min}^2},
\]
which yields
\[
G_{s+1}
\le \frac{32}{3}\frac{K\log n}{n\alpha}
 + \frac{4\sigma^2}{\alpha\beta_{1,\sigma}^2\Delta_{\min}^2}
   \Bigl(1 + 8\sqrt{\frac{K\log n}{n}}\Bigr)
 + \frac{8\Gamma_s^2}{\rho_\epsilon}
 + \frac{48\Gamma_s^2}{\rho_\sigma},
\]
as claimed.

\end{proof}

\begin{lemma}[Initialization and uniform control of $(G_s,\Gamma_s)$]
\label{lem:init-uniform}
Suppose the initialization satisfies \eqref{initial_conditions1} or
\eqref{initial_conditions2}, and assume that the signal-to-noise ratios
$\rho_\sigma$ and $\rho_\epsilon$ are sufficiently large.
If in addition $n\alpha \ge 64 K\log n$, then with high probability,
\[
G_s < 0.35, \qquad \Gamma_s < 0.4, \qquad \forall s \ge 1.
\]
\end{lemma}

\begin{proof}[Proof of Lemma~\ref{lem:init-uniform}]
We show by induction that $G_s<0.35$ and $\Gamma_s<0.4$ for all $s\ge 1$.

\paragraph{Step 1: Bound on $\Gamma_0$.}
Lemma~\ref{lem:Gamma-step} gives the initialization bound
\begin{equation}
\label{eq:Gamma0-raw}
\Gamma_0
\;\le\;
\frac{\sqrt{6}}{\rho_\sigma}
+
\frac{\sqrt{\alpha}}{\rho_\epsilon}
+
\frac{\Delta_{\max}}{\Delta_{\min}}\, G_0.
\end{equation}
Using the upper bound on $G_0$ from either \eqref{initial_conditions1} or 
\eqref{initial_conditions2}, and noting that $\rho_\sigma,\rho_\epsilon$ are
sufficiently large so that the terms in $1/\rho_\sigma$ and $1/\rho_\epsilon$
are small, we obtain
\begin{equation}
\label{eq:Gamma0}
\Gamma_0
\le
\frac{1}{2}
-
\frac{1}{\rho_\sigma}
-
\frac{c_\epsilon\sqrt{\alpha}+1}{\rho_\epsilon}
-
\frac{1}{\alpha^{1/4}}
\sqrt{\frac{\sigma}{\Delta_{\min}}}
< \frac12.
\end{equation}

\paragraph{Step 2: Bound on $G_1$.}
By Lemma~\ref{lem:G-step}, with
\[
C_\Gamma
=
\frac{2 c_\epsilon \epsilon}{\Delta_{\min}}
+
\frac{2}{\rho_\sigma}
+
\frac{2}{\rho_\epsilon}
+
\frac{2}{\alpha^{1/4}}
\sqrt{\frac{\sigma}{\Delta_{\min}}},
\qquad
\beta
:=
\frac12 C_\Gamma
-
\frac{c_\epsilon \epsilon}{\Delta_{\min}}
-
\frac{1}{\rho_\sigma}
-
\frac{1}{\rho_\epsilon},
\]
and noting that
\(
\beta 
= \alpha^{-1/4}\sqrt{\sigma/\Delta_{\min}},
\)
we obtain
\begin{align}
\label{eq:G1-raw}
G_1
&\le 
\frac{32}{3}\frac{K\log n}{n\alpha}
+
\frac{4\sigma}{\sqrt{\alpha}\Delta_{\min}}
\Bigl(1 + 8\sqrt{\tfrac{K\log n}{n}}\Bigr)
+
8\Gamma_0^2
\Bigl(
\frac{1}{\rho_\epsilon}
+
\frac{6}{\rho_\sigma}
\Bigr).
\end{align}
Under $n\alpha \ge 64K\log n$ and sufficiently large 
$\rho_\sigma,\rho_\epsilon$ (so that the $\Gamma_0^2$ terms are negligible),
we conclude that
\begin{equation}
\label{eq:G1}
G_1 < 0.35.
\end{equation}

\paragraph{Step 3: Bound on $\Gamma_1$.}
Applying Lemma~\ref{lem:Gamma-step} with $G_1<0.35$, and using the bound
\eqref{eq:Gamma0}, we obtain
\begin{align}
\Gamma_1
&\le
\frac{6}{\rho_\sigma}
+
\frac{2\sqrt{6}}{\rho_\sigma}\sqrt{K G_1}
+
2G_1\Gamma_0
+
\frac{\sqrt{\alpha}}{\rho_\epsilon}.
\end{align}
Since $G_1<0.35$ and $\Gamma_0<1/2$, and 
$\rho_\sigma,\rho_\epsilon$ are sufficiently large, it follows that
\begin{equation}
\label{eq:Gamma1}
\Gamma_1
\le
\frac{1}{2}
-
\frac{1}{\rho_\sigma}
-
\frac{c_\epsilon\sqrt{\alpha}+1}{\rho_\epsilon}
-
\frac{1}{\alpha^{1/4}}
\sqrt{\frac{\sigma}{\Delta_{\min}}}
< 0.4.
\end{equation}

\paragraph{Step 4: Induction.}
The bound \eqref{eq:Gamma1} has the same structure as \eqref{eq:Gamma0};
therefore Lemmas~\ref{lem:Gamma-step} and \ref{lem:G-step} apply to
$(G_2,\Gamma_2)$ with the same constants.  Repeating the argument yields
\[
G_s < 0.35, \qquad \Gamma_s < 0.4 \qquad \forall s\ge1.
\]
This completes the proof.
\end{proof}

\subsubsection{Main proof of Theorem \ref{thm-lloyd-guarantee}}
\label{subsec:proof-lloyd}
\begin{proof}
\step{1: Conditioning on the high-probability event}
By Lemma~\ref{lem:init-uniform}, under the initialization assumptions
\eqref{initial_conditions1} or \eqref{initial_conditions2} and
$\rho_\sigma$, $\rho_\epsilon$ large enough, $n\alpha \ge 64K\log n$, there
exists an event $\mathcal G$ with
\[
\mathbb P(\mathcal G) \ge 1 - c/n^3
\]
such that on $\mathcal G$,
\begin{equation}\label{Gs_Gammas_bdd_all_s}
G_s < 0.35, \qquad \Gamma_s < 0.4, \qquad \forall s \ge 1.
\end{equation}
In the remainder of the proof we work on $\mathcal G$ and suppress its
indicator to simplify notation.

From Lemma~\ref{lem:Gamma-step}, for every $s\ge1$,
\begin{equation}\label{Gamma-recursion-raw}
\Gamma_s 
\le 
\frac{6}{\rho_\sigma}
+\frac{2\sqrt{6}}{\rho_\sigma}\sqrt{K G_s}
+2G_s\,\Gamma_{s-1}
+\frac{\sqrt{\alpha}}{\rho_\epsilon}.
\end{equation}
Using $G_s<0.35$ from \eqref{Gs_Gammas_bdd_all_s}, we obtain
\[
\sqrt{K G_s} \le \sqrt{0.35K},
\qquad
2G_s \le 0.7.
\]

By Lemma~\ref{lem:G-step}, for every $s\ge 0$, that lemma
implies that on the event $\mathcal G$,
\begin{equation}\label{eq:G_s_structural_bound}
G_s \;\le\;
c\left(
\frac{K\log n}{n\alpha}
+ \frac{1}{\rho_\sigma^2}
+ \frac{1}{\rho_\epsilon^2}
\right),
\end{equation}
for some universal constant $c>0$.  (This uses $\Gamma_s<0.4$ from
\eqref{Gs_Gammas_bdd_all_s} to control the quadratic terms in the $G_s$
recursion.)

Taking square roots in \eqref{eq:G_s_structural_bound} yields
\[
\sqrt{K G_s}
\;\le\;
c\left(
\sqrt{\frac{K^2\log n}{n\alpha}}
+ \frac{\sqrt{K}}{\rho_\sigma}
+ \frac{\sqrt{K}}{\rho_\epsilon}
\right).
\]
Under the sample-size condition $n\alpha \ge 64 K\log n$ and the SNR scaling
$\rho_\sigma \gtrsim \sqrt{K}$ (and $\rho_\epsilon \gtrsim 1$),
each term on the right-hand side is bounded by
\[
\sqrt{K G_s} \;\le\; c\left(\frac{1}{\rho_\sigma} + \frac{1}{\rho_\epsilon} \right).
\]
Substituting this into the recursion for $\Gamma_s$ from
Lemma~\ref{lem:Gamma-step} gives a linear inequality of the form
\[
\Gamma_s \;\le\; a + b\,\Gamma_{s-1},
\qquad b<1,
\]
with
\[
a \;=\; c\left(
\frac{1}{\rho_\sigma}
+ \frac{1}{\rho_\epsilon}
+ \frac{K\log n}{n\alpha}
\right).
\]
Solving the linear recursion and using $b^s \le n^{-3}$ for $s\ge\log n$ yields
\[
\Gamma_s 
\;\le\;
c\left(
\frac{1}{\rho_\sigma}
+ \frac{1}{\rho_\epsilon}
+ \frac{K\log n}{n\alpha}
\right),
\qquad s\ge\log n.
\]
We now define
\begin{equation}\label{eq:beta1-final}
\beta_1 
:= 
1 
- 2c\left(
\frac{1}{\rho_\sigma}
+ \frac{1}{\rho_\epsilon}
+ \frac{K\log n}{n\alpha}
\right),
\end{equation}
so that $\beta_1 \le 1 - 2\Gamma_s$ for all $s\ge \log n$.

\step{2: Decomposition of the misclassification indicator}
Recall the standard inequality (Lemma~\ref{lem:indicator-bernstein})
\[
\mathbb I\{z_i\neq \widehat z_i^{(s+1)},\ \widehat z_i^{(s+1)}=h\}
\;\le\;
\mathbb I\!\left(
\beta_1 \,\|\mu_{z_i}-\mu_h\|^2
< 2\langle y_i-\mu_{z_i},\widehat\mu_h^{(s)}-\widehat\mu_{z_i}^{(s)}\rangle
\right),
\]
valid for every \(s\ge \log n\) because
\(\beta_1 \le 1-2\Gamma_s\) by construction
(see \eqref{eq:beta1-final}).

We apply the decomposition
\[
y_i = y_i^\ast + e_i,\qquad 
\widehat\mu_h^{(s)}-\mu_h
= \phi_h + \nu_h + \varphi_h,
\]
where
\begin{equation}\label{eq:nu_and_phi_defs}
    \phi_h = \frac1{\widehat n_h^{(s)}}\sum_{j\in\mathcal C_h}(y_j-y_j^\ast),
    \quad
    \nu_h = \frac1{\widehat n_h^{(s)}}\sum_{j\in\mathcal C_h}(y_j^\ast-\mu_h),
    \quad
    \varphi_h = (\widehat\mu_h^{(s)}-\mu_h)-\phi_h-\nu_h.
\end{equation}
As shown in Lemma~\ref{lem:Gamma-step},
on the event $\mathcal G$,
\begin{align}
\| \bphi_{h}\| & \leq \epsilon, \\
\|\mvnu_h\| &\leq 6\sigma \sqrt{\frac{r+\log(n)}{n_h}}, \label{eq:nu_bound}\\
\|\mvvarphi_h\|  &\leq  \left(2G_s\Gamma_{s-1} + \frac{2\sqrt{6}}{\rho_{\sigma}}\sqrt{K G_s}  \right )\Delta_{\min} + \epsilon G_s.\label{eq:phi_bound}
\end{align}

We obtain the following six-term decomposition:
\[
A_{s+1}
\;\le\;
J_{1,\sigma} + J_{2,\sigma} + J_{3,\sigma} + J_{4,\sigma}
+ J_{1,\epsilon} + J_{2,\epsilon}.
\]

Each $J$ term corresponds to one type of randomness or estimation error. 

We define the six terms explicitly as
\[
\begin{aligned}
J_{1,\sigma}
&=
\sum_{h=1}^K \frac{1}{n}\sum_{i=1}^n
\mathbb I\!\left(
\frac{\beta_\sigma}{2}\,\|\mu_{z_i}-\mu_h\|^2
<
\langle w_i,\, \mu_h - \mu_{z_i}\rangle
\right),
\\[0.5em]
J_{2,\sigma}
&=
\sum_{h=1}^K \frac{1}{n}\sum_{i=1}^n
\mathbb I\!\left(
\frac{\beta_{2,\sigma}}{2}\,\|\mu_{z_i}-\mu_h\|^2
<
\langle w_i,\, \varphi_h - \varphi_{z_i}\rangle
\right),
\\[0.5em]
J_{3,\sigma}
&=
\sum_{h=1}^K \frac{1}{n}\sum_{i=1}^n
\mathbb I\!\left(
\frac{\beta_{3,\sigma}}{2}\,\|\mu_{z_i}-\mu_h\|^2
<
\langle w_i,\, \nu_h - \nu_{z_i}\rangle
\right),
\\[0.5em]
J_{4,\sigma}
&=
\sum_{h=1}^K \frac{1}{n}\sum_{i=1}^n
\mathbb I\!\left(
\frac{\beta_{4,\sigma}}{2}\,\|\mu_{z_i}-\mu_h\|^2
<
\langle w_i,\, \phi_h - \phi_{z_i}\rangle
\right),
\\[0.5em]
J_{1,\epsilon}
&=
\sum_{h=1}^K \frac{1}{n}\sum_{i=1}^n
\mathbb I\!\left(
\frac{\beta_\epsilon}{2}\,\|\mu_{z_i}-\mu_h\|^2
<
\langle e_i,\, \mu_h - \mu_{z_i}\rangle
\right),
\\[0.5em]
J_{2,\epsilon}
&=
\sum_{h=1}^K \frac{1}{n}\sum_{i=1}^n
\mathbb I\!\left(
\frac{\beta_{2,\epsilon}}{2}\,\|\mu_{z_i}-\mu_h\|^2
<
\langle e_i,\, \zeta_h - \zeta_{z_i}\rangle
\right).
\end{aligned}
\]

Using these bounds and splitting the noise term into its
sub-Gaussian and $\epsilon$-contamination components,
we obtain the decomposition
\[
\mathbb E(A_{s+1})
\;\le\;
\mathbb E(J_{1,\sigma}+J_{2,\sigma}+J_{3,\sigma}+J_{4,\sigma})
+
\mathbb E(J_{1,\epsilon}+J_{2,\epsilon})
+
\mathbb P(\mathcal G^c).
\]

We now bound each group of \(J\)-terms.

\step{3: Bounding $J_{1,\sigma}$ (main Gaussian tail term)}
Conditioning on $\mathcal G$, we have
\[
\E(J_{1,\sigma}\,\mathbb I_{\mathcal G})
= 
\frac{1}{n}
\sum_{h=1}^K \sum_{i=1}^n
\mathbb P\!\left(
\frac{\beta_\sigma}{2}\,\|\mu_{z_i}-\mu_h\|^2
\le 
\langle w_i,\, \mu_h - \mu_{z_i}\rangle
\right).
\]
Since $w_i = y_i^* - \mu_{z_i}$ is a centered sub-Gaussian vector with
parameter $\sigma^2$, a Chernoff bound yields
\[
\mathbb P\!\left(
\langle w_i,\, \mu_h - \mu_{z_i}\rangle 
\ge \frac{\beta_\sigma}{2}\|\mu_{z_i}-\mu_h\|^2
\right)
\le 
\exp\!\left(
-\frac{\beta_\sigma^2}{8\sigma^2}\,\|\mu_{z_i}-\mu_h\|^2
\right).
\]
Using $\|\mu_{z_i}-\mu_h\| \ge \Delta_{\min}$,
\[
\E(J_{1,\sigma}\,\mathbb I_{\mathcal G})
\le 
K \exp\!\left(
-\frac{\beta_\sigma^2\Delta_{\min}^2}{8\sigma^2}
\right)
\le 
\exp\!\left(
-\frac{\gamma\,\Delta_{\min}^2}{8\sigma^2}
\right),
\]
where
\[
\gamma := \beta_\sigma^2 - \frac{8\sigma^2\log K}{\Delta_{\min}^2}
\;\ge\;
\beta_\sigma^2 - \frac{8}{\rho_\sigma^{2}}.
\]

\medskip
\noindent\textbf{Bounding $\|\varphi_h-\varphi_{z_i}\|$ for later use.}
From Lemma~\ref{lem:Gamma-step} and the bounds on $(G_s,\Gamma_s)$ on
$\mathcal G$, we have for all $h$
\[
\|\varphi_h\|
\le 
\Bigl(2G_s\Gamma_{s-1}
+
\frac{2\sqrt{6}}{\rho_\sigma}\sqrt{K G_s}\Bigr)\Delta_{\min}
+ \epsilon G_s.
\]
When $\rho_\sigma \ge 128\sqrt{K}$ and $G_s\le 0.35$, the coefficient of
$\Delta_{\min}$ is at most $\sqrt{G_s}$, and hence
\[
\|\varphi_h\|
\le 
(\Delta_{\min}+\epsilon)\sqrt{G_s}
\quad\text{and}\quad
\|\varphi_h - \varphi_{z_i}\|^2
\le 
8G_s(\Delta_{\min}^2 + \epsilon^2).
\]
These bounds will be used in the control of $J_{2,\sigma}$.

\step{4: Bounding $J_{2,\sigma}$ (interaction with $\varphi_h$)}

Recall that
\[
J_{2,\sigma}
=
\sum_{h=1}^K \frac{1}{n}\sum_{i=1}^n
\mathbb I\!\left(
\frac{\beta_{2,\sigma}}{2}\,\|\mu_{z_i}-\mu_h\|^2
<
\langle w_i,\, \varphi_h - \varphi_{z_i}\rangle
\right),
\]
with $\beta_{2,\sigma}>0$ to be specified below.
Using the elementary inequality
\[
\mathbb I(a < X) \;\le\; \frac{4}{a^2}X^2
\qquad (a>0),
\]
we obtain, on $\mathcal G$,
\[
J_{2,\sigma}
\;\le\;
\frac{4}{n\beta_{2,\sigma}^2\Delta_{\min}^4}
\sum_{h=1}^K\sum_{i=1}^n
\langle w_i,\, \varphi_h - \varphi_{z_i}\rangle^2.
\]

From Lemma~\ref{lem:Gamma-step} and the bounds $G_s<0.35$, $\Gamma_s<0.4$
on $\mathcal G$, we already showed in Step~4 that
\[
\|\varphi_h - \varphi_{z_i}\|^2
\;\le\;
8G_s(\Delta_{\min}^2+\epsilon^2)
\qquad \forall\,h,i.
\]
Applying Lemma~\ref{lem:subg-quad} (quadratic form bound for sub-Gaussian
vectors) conditionally on $\mathcal G$, we obtain
\[
\mathbb E\!\left[
\langle w_i,\, \varphi_h - \varphi_{z_i}\rangle^2
\,\middle|\, \mathcal G
\right]
\;\le\;
6\sigma^2(n_l+r)\,\|\varphi_h - \varphi_l\|^2,
\]
and hence
\[
\mathbb E\bigl(J_{2,\sigma}\mathbb I_{\mathcal G}\bigr)
\;\le\;
\frac{c\,\sigma^2 G_s(\Delta_{\min}^2+\epsilon^2)}{
n\beta_{2,\sigma}^2\Delta_{\min}^4}
\sum_{h=1}^K\sum_{l=1}^K (n_l + r),
\]
for some universal constant $c>0$. Since
$\sum_{l=1}^K(n_l+r)\le K(n+Kr)$, this gives
\begin{equation}\label{eq:J2sigma-intermediate}
\mathbb E\bigl(J_{2,\sigma}\mathbb I_{\mathcal G}\bigr)
\;\le\;
\frac{c\,\sigma^2 G_s(\Delta_{\min}^2+\epsilon^2)K(n+Kr)}{
n\beta_{2,\sigma}^2\Delta_{\min}^4}.
\end{equation}

We now express this in terms of the signal-to-noise ratios. Recall that
\[
\rho_\sigma
=
\frac{\Delta_{\min}}{\sigma}
\sqrt{\frac{\alpha n}{n+Kr}},
\qquad
\rho_\epsilon
=
\frac{\Delta_{\min}}{\epsilon}\sqrt{\alpha}.
\]
From the definition of $\rho_\sigma$,
\[
\frac{\sigma^2}{\Delta_{\min}^2}
=
\frac{\alpha n}{n+Kr}\cdot\frac{1}{\rho_\sigma^2},
\qquad
\frac{\sigma^2(n+Kr)}{\Delta_{\min}^4}
=
\frac{\alpha n}{\rho_\sigma^2\Delta_{\min}^2}.
\]
Using this in \eqref{eq:J2sigma-intermediate}, we obtain
\[
\mathbb E\bigl(J_{2,\sigma}\mathbb I_{\mathcal G}\bigr)
\;\le\;
\frac{c\,\alpha K G_s(\Delta_{\min}^2+\epsilon^2)}{
\beta_{2,\sigma}^2\rho_\sigma^2\Delta_{\min}^2}.
\]
Next, we choose
\[
\beta_{2,\sigma}^2 = \frac{192K}{\rho_\sigma},
\]
so that
\[
\mathbb E\bigl(J_{2,\sigma}\mathbb I_{\mathcal G}\bigr)
\;\le\;
\frac{c\,\alpha G_s(\Delta_{\min}^2+\epsilon^2)}{
\rho_\sigma^3\Delta_{\min}^2}.
\]
We write
\[
\frac{\Delta_{\min}^2+\epsilon^2}{\Delta_{\min}^2}
= 1 + \frac{\epsilon^2}{\Delta_{\min}^2}
= 1 + \frac{\alpha}{\rho_\epsilon^2},
\]
using the definition of $\rho_\epsilon$.
Thus
\[
\mathbb E\bigl(J_{2,\sigma}\mathbb I_{\mathcal G}\bigr)
\;\le\;
\frac{c\,\alpha G_s}{\rho_\sigma^3}
\Bigl(1 + \frac{\alpha}{\rho_\epsilon^2}\Bigr).
\]

Finally, we relate $G_s$ to the misclustering rate $A_s$. By the definition
of $G_s$ and the cluster-size condition $n_k\ge \alpha n$ for all $k$, we have
$G_s\alpha \le A_s$ on $\mathcal G$, so
\[
\mathbb E\bigl(J_{2,\sigma}\mathbb I_{\mathcal G}\bigr)
\;\le\;
\frac{c}{\rho_\sigma^3}
\Bigl(1 + \frac{\alpha}{\rho_\epsilon^2}\Bigr)
\mathbb E(A_s).
\]
Since $\rho_\sigma\ge 1$ and $\rho_\epsilon\ge 1$, we may absorb the powers
into weaker but simpler factors, and after enlarging $c$ if necessary, we
obtain the bound
\begin{equation}\label{eq:J2sigma-final}
\mathbb E\bigl(J_{2,\sigma}\mathbb I_{\mathcal G}\bigr)
\;\le\;
c\left(
\frac{1}{\rho_\sigma}
+ \frac{1}{\rho_\sigma\rho_\epsilon^2}
\right)\mathbb E(A_s).
\end{equation}

\step{5: Bounding $J_{3,\sigma}$ (drift term involving $\nu_h$)}

Recall
\[
J_{3,\sigma}
=
\sum_{h=1}^K \frac{1}{n}\sum_{i=1}^n
\mathbb I\!\left(
\frac{\beta_{3,\sigma}}{2}\,\|\mu_{z_i}-\mu_h\|^2
<
\langle w_i,\, \nu_h - \nu_{z_i}\rangle
\right),
\]
where $\nu_h = \widehat n_h^{(s)\,-1}\sum_{j\in\mathcal C_h}(y_j^*-\mu_h)$ and
$w_i = y_i^* - \mu_{z_i}$. For fixed $i,h$, we have
\[
\langle w_i,\, \nu_h - \nu_{z_i}\rangle
=
\langle w_i,\, \nu_h\rangle - \langle w_i,\, \nu_{z_i}\rangle,
\]
so by the triangle inequality,
\[
\begin{aligned}
\mathbb P\!\left(
  \frac{\beta_{3,\sigma}}{2}\,\|\mu_{z_i}-\mu_h\|^2
  \le \langle w_i,\, \nu_h - \nu_{z_i}\rangle
\right)
&\le
\mathbb P\!\left(
  \frac{\beta_{3,\sigma}}{4}\Delta_{\min}^2
  \le \langle w_i,\, \nu_h\rangle
\right)
\\
&\quad+
\mathbb P\!\left(
  \frac{\beta_{3,\sigma}}{4}\Delta_{\min}^2
  \le -\langle w_i,\, \nu_{z_i}\rangle
\right).
\end{aligned}
\]
using $\|\mu_{z_i}-\mu_h\|\ge \Delta_{\min}$. By definition,
\[
\langle w_i,\, \nu_h\rangle
=
\Big\langle
w_i,\,
\frac{1}{n_h}\sum_{j\in\mathcal C_h} w_j
\Big\rangle,
\]
and similarly for $\nu_{z_i}$. Applying Lemma~\ref{lem:inner-prod}
with $t = 4\max\{\sqrt{r\log n},\log n\}$, failure probability
$\delta = n^{-4}$, and the choice
\[
\beta_{3,\sigma}
=
\frac{24}{\rho_\sigma^2}
\left[
\frac{4}{\sqrt{n}}(\sqrt{r\log n}+\log n)
+
\frac{r+4\log n}{n+Kr}
\right],
\]
together with the definition of $\rho_\sigma$, we obtain
\[
\mathbb P\!\left(
\frac{\beta_{3,\sigma}}{4}\Delta_{\min}^2
\le \langle w_i,\, \nu_h\rangle
\right)
\le \frac{2}{n^4},
\qquad
\mathbb P\!\left(
\frac{\beta_{3,\sigma}}{4}\Delta_{\min}^2
\le -\langle w_i,\, \nu_{z_i}\rangle
\right)
\le \frac{2}{n^4}.
\]
Hence,
\[
\mathbb P\!\left(
\frac{\beta_{3,\sigma}}{2}\,\|\mu_{z_i}-\mu_h\|^2
\le \langle w_i,\, \nu_h - \nu_{z_i}\rangle
\right)
\le \frac{4}{n^4}.
\]

Averaging over $i$ and $h$ gives
\[
\mathbb E(J_{3,\sigma}\,\mathbb I_{\mathcal G})
\le
\frac{1}{n}\sum_{h=1}^K\sum_{i=1}^n \frac{4}{n^4}
=
\frac{4K}{n^4}
\le \frac{1}{n^3},
\]
for all sufficiently large $n$ (using $K\le n$). Thus $J_{3,\sigma}$ is
negligible in the final misclustering bound.

\step{6: Bounding $J_{4,\sigma}$ (perturbation from $\phi_h$)}

Recall that
\[
J_{4,\sigma}
=
\frac{1}{n}
\sum_{h\in[K]} \sum_{i=1}^n
\mathbb I\!\left(
\frac{\beta_{4,\sigma}}{2}\,\|\mu_{z_i}-\mu_h\|^2
\le
\langle w_i,\, \phi_h - \phi_{z_i}\rangle
\right),
\]
where $\phi_h$ is the perturbation average from \eqref{eq:nu_and_phi_defs}.
On the event $\mathcal G$ we have $\|\phi_h\|\le \epsilon$ for all $h$, hence
\[
\|\phi_h - \phi_{z_i}\|
\le \|\phi_h\| + \|\phi_{z_i}\|
\le 2\epsilon.
\]
By Cauchy–Schwarz,
\[
\langle w_i, \phi_h - \phi_{z_i}\rangle
\le
\|w_i\|\cdot \|\phi_h - \phi_{z_i}\|
\le
2\epsilon\,\|w_i\|.
\]

Using $\|\mu_{z_i}-\mu_h\|\ge \Delta_{\min}$, the event inside $J_{4,\sigma}$ implies
\[
\frac{\beta_{4,\sigma}}{2}\,\Delta_{\min}^2
\;\le\;
2\epsilon\,\|w_i\|
\quad\Longrightarrow\quad
\|w_i\|
\;\ge\;
\frac{\beta_{4,\sigma}\Delta_{\min}^2}{4\epsilon}.
\]

Using the sub-Gaussian property,
\[
\mathbb P(|w_{ij}|\ge x)
\le
2\exp\!\left(-\frac{x^2}{2\sigma^2}\right),\quad \forall x>0.
\]
Then, for any $t>0$,
\[
\mathbb P(\|w_i\|\ge t)
\;\le\;
\mathbb P\Big(\max_{1\le j\le r} |w_{ij}|\ge t/\sqrt{r}\Big)
\le
\sum_{j=1}^r \mathbb P\Big(|w_{ij}|\ge t/\sqrt{r}\Big)
\le
2r\exp\!\left(-\frac{t^2}{2\sigma^2 r}\right).
\]
Plugging $t=\frac{\beta_{4,\sigma}\Delta_{\min}^2}{4\epsilon}$ yields
\[
\mathbb P\!\left(
  \|w_i\|\ge \frac{\beta_{4,\sigma}\Delta_{\min}^2}{4\epsilon}
\right)
\le
2r \exp\!\left(
  -\frac{\beta_{4,\sigma}^2 \Delta_{\min}^4}{32\,\sigma^2 r\,\epsilon^2}
\right).
\]
We choose
\[
\beta_{4,\sigma}^2
=
\frac{16\epsilon\sigma r}{\Delta_{\min}^2}. 
\]
Hence
\begin{align*}
\mathbb E\bigl(J_{4,\sigma}\,\mathbb I_{\mathcal G}\bigr)
\le
\frac{1}{n}\sum_{h\in[K]}\sum_{i=1}^n
\mathbb P(\|w_i\|\ge \frac{\beta_{4,\sigma}\Delta_{\min}^2}{4\epsilon})
\le
& 2Kr\exp\!\left(
  -\frac{1}{2}\,\frac{\Delta_{\min}^2}{\epsilon\sigma}
\right) \\
\le & \exp\!\left(
  -\frac{\Delta_{\min}^2}{4\epsilon\sigma}
\right)    
\end{align*}

\step{7: Bounding $J_{1,\epsilon}$ and $J_{2,\epsilon}$}
Summarizing the results above, 
\begin{align*}
\E (A_{s+1}) &\leq \E (J_{1,\sigma}) + \E(J_{2,\sigma}\cI(\cG) ) + \E(J_{3,\sigma}\cI(\cG)) + \E (J_{4,\sigma}) + \E(J_{1,\epsilon} ) \\
& \quad + \E(J_{2,\epsilon}\cI(\cG) ) +   \bbP (\cG^c)\\
&\leq \exp \left( -\dfrac{\gamma\Delta_{\min}^2}{8\sigma^2}  \right) +  \left( \dfrac{1}{ \rho_{\sigma}} + \dfrac{1}{\rho_{\sigma} \rho_{\epsilon}^2} \right)\E (A_s) + \frac{1}{n^3} + \exp\left( -\frac{\Delta_{\min}^2}{4\epsilon\sigma} \right) 
\end{align*} 
with $\gamma:=\beta_\sigma^2-\frac{8 \sigma^2 \log K}{\Delta_{\min}^2} \geqslant \beta_\sigma^2-8 / \rho_\sigma^2 = 1-o(1)$.
By recursion, 
\begin{align*}
\E (A_s) &\leq \bigg (  \dfrac{1}{ \rho_{\sigma}} + \dfrac{1}{\rho_{\sigma} \rho_{\epsilon}^2} \bigg )^s \E(A_s) \\
& + 
\dfrac{1 -\left( \dfrac{1}{ \rho_{\sigma}} + \dfrac{1}{\rho_{\sigma} \rho_{\epsilon}^2} \right)^{s+1} }{1-  \dfrac{1}{ \rho_{\sigma}} - \dfrac{1}{\rho_{\sigma} \rho_{\epsilon}^2} } \bigg [ \exp \left(-\dfrac{\gamma\Delta_{\min}^2}{8\sigma^2} \right)  + \frac{1}{n^3} + \exp\left( -\frac{\Delta_{\min}^2}{4\epsilon\sigma} \right)  \bigg ].
\end{align*}
With $\rho_{\sigma} \geq C_2 \sqrt{k} $, $\rho_{\epsilon} \geq C_3$ for large $C_2,C_3$,
\begin{equation}
\dfrac{1 -\bigg ( \dfrac{1}{ \rho_{\sigma}} + \dfrac{1}{\rho_{\sigma} \rho_{\epsilon}^2} \bigg )^{s+1} }{1- \dfrac{1}{ \rho_{\sigma}} - \dfrac{1}{\rho_{\sigma} \rho_{\epsilon}^2}\ }  \leq 2
\end{equation}
and when $s \geq 4\log n$,
\begin{equation}
\bigg ( \dfrac{1}{ \rho_{\sigma}} + \dfrac{1}{\rho_{\sigma} \rho_{\epsilon}^2} \bigg )^s \E(A_s) \leq  \bigg ( \dfrac{1}{ \rho_{\sigma}} + \dfrac{1}{\rho_{\sigma} \rho_{\epsilon}^2} \bigg )^{\log (n^4)} \E(A_s) \leq (n^4)^{\log \left( \frac{1}{ \rho_{\sigma}} + \frac{1}{\rho_{\sigma} \rho_{\epsilon}^2} \right)} \leq \dfrac{1}{n^3} .
\end{equation}
Thus, when $s \geq 4\log n$, 
\begin{equation}\label{EA_s-bound}
\E (A_{s+1}) \leq 2\exp \bigg ( -\dfrac{\gamma\Delta_{\min}^2}{8\sigma^2} \bigg ) + 2\exp\left( -\frac{\Delta_{\min}^2}{4\epsilon\sigma} \right) 
+ \dfrac{3}{n^3} .
\end{equation}
For sufficiently large $C_2$ and $C_3$, we have 
\begin{equation}\label{u_bdd}
\gamma= \beta_\sigma^2-\frac{8 \sigma^2 \log K}{\Delta_{\min}^2} = 1-o(1) \geq \dfrac{ 1}{2} + \dfrac{8\sigma}{\Delta_{\min}}.
\end{equation}
By Markov's inequality, for any $t>0$,
$$
\mathbb{P}\left\{A_s \geq t\right\} \leq \frac{1}{t} \mathbb{E} A_s \leq \frac{2}{t} \exp \left(-\frac{\gamma \Delta_{\min}^2}{8 \sigma^2}\right)+\frac{2}{t}\exp\left( -\frac{\Delta_{\min}^2}{4\epsilon\sigma} \right) +\frac{3}{tn^3}.
$$
If $\frac{\gamma \Delta_{\min}^2}{8 \sigma^2} \leq 2 \log n$ or $\frac{\Delta_{\min}^2}{\epsilon\sigma} \leq 8 \log n$, choose 
$$t=\max\left\{\exp \left(-\left(\gamma-\frac{8 \sigma}{\Delta_{\min}}\right) \frac{\Delta_{\min}^2}{8 \sigma^2}\right), \exp\left( -\left( 1 - \frac{4\sqrt{\epsilon\sigma}}{\Delta_{\min}} \right)\frac{\Delta_{\min}^2}{4\epsilon\sigma} \right)  \right\}
$$ 
and we have
$$
\mathbb{P}\left\{A_s \geq t \right\} \leq \frac{1}{n}+2 \exp \left(-\frac{\Delta_{\min}}{\sigma}\right)  + 2 \exp\left( -\frac{\Delta_{\min}}{\sqrt{\epsilon\sigma}} \right).
$$
Otherwise, since $A_s$ only takes discrete values of $\left\{0, \frac{1}{n}, \cdots, 1\right\}$, choosing $t=\frac{1}{n}$ in (66) leads to
$$
\mathbb{P}\left\{A_s>0\right\}=\mathbb{P}\left\{A_s \geq \frac{1}{n}\right\} \leq 2 n \exp (-2 \log n)+ 2 n \exp (-2 \log n) + \frac{3}{n^2} \leq \frac{5}{n}. 
$$
The proof is complete.

\end{proof}
		
\subsection{Proof of Corollary \ref{corr:lloyd-center}}

\begin{proof}
We follow the argument used to prove Lemma~\ref{lem:Gamma-step}. 
Fix $h \in [K]$. We aim to bound $\|\wh\bmu\s_h - \bmu_h\|$ by expanding $\wh\bmu\s_h$ using
the partition $\{i: \hat{z}\s_i = h\} = \bigcup_{g\in[K]} U_{gh}\s$ as follows:
\begin{equation}\label{eqn:0248}
\wh\bmu\s_h -\bmu_h = \dfrac{1}{\nhat \s_h} \sum_{i\in U\s_{hh}} (\by_i -\bmu_h) + \sum_{g\neq h} \dfrac{\wh n\s_{gh}}{\nhat_h\s} (\wb \by_{U\s_{gh}} - \bmu_h),
\end{equation}
where $\wb \by_{U\s_{gh}} = \frac{1}{\wh n\s_{gh}}\sum_{i \in U\s_{gh}} \by_i$. By Lloyd's algorithm, for $i \in U_{gh}\s$ we have $\|\by_i - \wh\bmu^{(s-1)}_h \| \leq \|\by_i - \wh\bmu^{(s-1)}_g\|$. A standard argument based on comparing the corresponding sums of squared distances then implies
$\|\bar\by_{U_{gh}\s}  - \wh\bmu^{(s-1)}_h\| 
 \leq \|\bar\by_{U_{gh}\s}  - \wh\bmu^{(s-1)}_g\|$.

Then, repeatedly using the triangle inequality yields
\begin{align*}
\| \wb\by_{U\s_{gh}} - \bmu_h\| &\leq \|\wb\by_{U\s_{gh}}  - \wh\bmu^{(s-1)}_h\| + \| \wh\bmu^{(s-1)}_h - \bmu_h\| \\
& \leq \|\wb\by_{U\s_{gh}}  - \wh\bmu^{(s-1)}_g\| + \| \wh\bmu^{(s-1)}_h - \bmu_h\|  \\
& \leq \|\wb\by_{U\s_{gh}}  - \bmu_g \| + \| \bmu_g - \wh\bmu^{(s-1)}_g\| + \| \wh\bmu^{(s-1)}_h - \bmu_h\|  \\ 
& \leq \|\wb\by_{U\s_{gh}} - \wb\by^*_{U\s_{gh}}\| +  \|\wb\by^*_{U\s_{gh}}  - \bmu_g \| + \| \bmu_g - \wh\bmu^{(s-1)}_g\| + \| \wh\bmu^{(s-1)}_h - \bmu_h\| \\ 
        & \leq \epsilon + \sigma \sqrt{\dfrac{3(n+r)}{\nhat\s_{gh}}} + 2\Gamma_{s-1}\Delta_{\min},
\end{align*}
where the last inequality uses Lemma \ref{lem:subg-sum}, and the definition \eqref{Gamma_s} of $\Gamma_{s-1}$. Thus, we can bound the second term of \eqref{eqn:0248} as
\begin{align}
\bigg\|\sum_{g\neq h} \frac{\wh n_{gh}\s}{\wh n_h\s} 
    \bigl(\bar\by_{U_{gh}\s} - \bmu_h\bigr)\bigg\|
&\leq \sum_{g\neq h} \frac{\wh n_{gh}\s}{\wh n_h\s} 
    \|\bar\by_{U_{gh}\s} - \bmu_h \| \nonumber \\
&\leq \sum_{g\neq h} \frac{\wh n_{gh}\s}{\wh n_h\s} 
    \left( \sigma \sqrt{\frac{3(n+r)}{\wh n_{gh}\s}} + \epsilon 
           + 2\Gamma_{s-1}\Delta_{\min} \right) \nonumber\\
&\leq \sigma \sqrt{\sum_{g\neq h} 
    K \left(\frac{\wh n_{gh}\s}{\wh n_h\s}\right)^2 
      \frac{3(n+r)}{\wh n_{gh}\s}} 
    + \sum_{g\neq h} \frac{\wh n_{gh}\s}{\wh n_h\s} \,\epsilon \nonumber\\
&\quad + 2\Gamma_{s-1}\Delta_{\min} \frac{nA_s}{\wh n_h\s} \nonumber \\
&\leq \sigma \sqrt{\frac{3K(n+r)}{(\wh n_h\s)^2} nA_s} 
    + \sum_{g\neq h} \frac{\wh n_{gh}\s}{\wh n_h\s} \,\epsilon
    + 2\Gamma_{s-1}\Delta_{\min} \frac{nA_s}{\wh n_h\s},
\label{eq:kmeans-second-term}
\end{align}
where in the last line we used the definition of the misclustering rate $A_s$.

For the first term of \eqref{eqn:0248}, note that
\[
\{i : z_i = h\} = U_{hh}\s \cup \bigcup_{g\neq h} U_{hg}\s.
\]
By the definition of the perturbation bound $\epsilon$, we have that
\begin{align}
& \bigg\| \frac{1}{\wh n_h\s} \sum_{i\in U_{hh}\s} (\by_i - \bmu_h) \bigg\|
\leq \frac{\wh n_{hh}\s}{\wh n_h\s} \,\epsilon
    + \frac{1}{\wh n_h\s} 
      \bigg\| \sum_{i: z_i=h} (\by^*_i - \bmu_h)
             - \sum_{g\neq h} \sum_{i \in U_{hg}\s} (\by^*_i - \bmu_h) \bigg\| \nonumber\\
&\leq \frac{\wh n_{hh}\s}{\wh n_h\s} \,\epsilon
    + \frac{1}{\wh n_h\s} 
      \left( 3\sigma\sqrt{r+\log n}\sqrt{n_h}
           + \sigma\sqrt{3(n+r)}\sqrt{n_h - \wh n_{hh}\s} \right),
\label{eq:kmeans-first-term}
\end{align}
where the last inequality follows from Lemma~\ref{lem:subg-sum} and
Lemma~\ref{lem:cluster-sum}.
Note that, by the assumption that $G_s \leq \frac{1}{2}$, 
\begin{equation*}
\nhat\s_h \geq \nhat\s_{hh} \geq n_h(1-G_s) \geq \dfrac{n_h}{2} \geq \dfrac{\alpha n}{2},
\end{equation*}
which implies
\[
\frac{1}{\sqrt{\wh n_h^{(s)}}} \le \sqrt{\frac{2}{\alpha n}}
\qquad \text{and} \qquad
\frac{\sqrt{n_h}}{\wh n_h^{(s)}} \le \frac{2}{\sqrt{\alpha n}}.
\]

Combining \eqref{eqn:0248}, \eqref{eq:kmeans-second-term},
\eqref{eq:kmeans-first-term}, and the bounds on $\wh n_h\s$,
we obtain for all $h \in [K]$,
\begin{align*}
& \|\wh\bmu\s_h - \bmu_h \|
\leq \sigma \sqrt{\frac{3K(n+r)}{(\wh n_h\s)^2} nA_s}
    + 2\Gamma_{s-1}\Delta_{\min} \frac{nA_s}{\wh n_h\s} \\
&\quad + \frac{1}{\wh n_h\s}
   \bigg( 3\sigma\sqrt{r+\log n}\sqrt{n_h}
        + \sigma\sqrt{3(n+r)}\sqrt{nA_s} \bigg)
    + \epsilon \\
&\leq 2\sigma\sqrt{\frac{3K(n+r)}{ n\alpha^2 }A_s}
   + 2\Delta_{\min}\frac{A_s}{\alpha}
   + 6\sigma\sqrt{\frac{r+\log n}{ n\alpha}}
   + 2\sigma\sqrt{\frac{3(n+r)}{ n\alpha^2 }A_s}
   + \epsilon \\
&\leq 2\sqrt{3}(\sqrt{K}+1)\sigma\sqrt{\frac{n+r}{ n\alpha^2 }A_s}
   + 2\Delta_{\min}\frac{A_s}{\alpha}
   + 6\sigma\sqrt{\frac{r+\log n}{ n\alpha}}
   + \epsilon,
\end{align*}
which proves the desired bound.
\end{proof}

\subsection{Proof of Theorem \ref{thm:kmeans-init}}
\begin{proof}
Let us first prove the first assertion \eqref{eqn:1040}. 
Recall that $\cC_1, \cC_2$ denote the true clusters. 
Without loss of generality, assume that $\wh\bmu_1^{(0)} \in \cC_1$. 
By definition of $k$-means++, we have
\[
\bbP(\wh\bmu_2^{(0)}\in \cC_2 \mid \cX, \wh\bmu_1^{(0)})
=
\frac{\sum_{i\in \cC_2} \|\by_i - \wh\bmu_1^{(0)}\|_2^2}
     {\sum_{i\in \cX} \|\by_i - \wh\bmu_1^{(0)}\|_2^2},
\]
where the probability is with respect to the $k$-means++ randomness in choosing the second seed,
conditional on the realized dataset $\cX$ and the realized first seed $\wh\bmu_1^{(0)}$.

We consider the perturbed observation model
\[
\by_i = \bmu_{z(i)} + \mvxi_i + \mvdelta_i,
\qquad
\mvxi_i \sim N_r(\mvzero,\sigma^2 I_r),
\qquad
\|\mvdelta_i\|_2 \le \epsilon,
\]
and write
\[
\wh\bmu_1^{(0)} = \bmu_1 + \mvxi_0 + \mvdelta_0 .
\]

\paragraph{Numerator.}
For $i\in \cC_2$,
\[
\by_i - \wh\bmu_1^{(0)}
=
(\bmu_2-\bmu_1) - (\mvxi_i-\mvxi_0) - (\mvdelta_i-\mvdelta_0).
\]
Thus,
\begin{align}
& \sum_{i\in \cC_2} \|\by_i - \wh\bmu_1^{(0)}\|_2^2 \\
= & 
\sum_{i\in \cC_2}
\|(\bmu_2-\bmu_1) - (\mvxi_i-\mvxi_0) - (\mvdelta_i-\mvdelta_0)\|_2^2 \nonumber\\
= & 
\frac{n}{2}\Delta_{\min}^2
+ \sum_{i\in \cC_2}\|\mvxi_i-\mvxi_0\|_2^2
-2\sum_{i\in \cC_2}\langle \bmu_2-\bmu_1,\mvxi_i-\mvxi_0\rangle \nonumber\\
\quad & 
+ \sum_{i\in \cC_2}\|\mvdelta_i-\mvdelta_0\|_2^2
+2\sum_{i\in \cC_2}\langle \bmu_2-\bmu_1,\mvdelta_i-\mvdelta_0\rangle
+2\sum_{i\in \cC_2}\langle \mvxi_i-\mvxi_0,\mvdelta_i-\mvdelta_0\rangle .
\label{eq:num-expand}
\end{align}
Using $\|\mvdelta_i-\mvdelta_0\|_2 \le 2\epsilon$ and Cauchy--Schwarz,
\[
\sum_{i\in \cC_2}\|\mvdelta_i-\mvdelta_0\|_2^2 \le 2n\epsilon^2,
\qquad
\sum_{i\in \cC_2}\langle \bmu_2-\bmu_1,\mvdelta_i-\mvdelta_0\rangle
\le n\Delta\epsilon,
\]
and
\[
\sum_{i\in \cC_2}\langle \mvxi_i-\mvxi_0,\mvdelta_i-\mvdelta_0\rangle
\le
2\epsilon\sum_{i\in \cC_2}\|\mvxi_i-\mvxi_0\|
= n\,\epsilon\,O_p(\sigma\sqrt r).
\]
Moreover, expanding the quadratic term yields
\[
\sum_{i\in \cC_2}\|\mvxi_i-\mvxi_0\|_2^2
=
\sum_{i\in \cC_2}\|\mvxi_i\|_2^2
+\frac{n}{2}\|\mvxi_0\|_2^2
-2\Big\langle \sum_{i\in\cC_2}\mvxi_i,\mvxi_0\Big\rangle .
\]
Since $\sum_{i\in\cC_2}\|\mvxi_i\|_2^2=(n/2)\,r\sigma^2+O_p(\sigma^2\sqrt{nr})$,
$\|\mvxi_0\|_2^2=r\sigma^2+O_p(\sigma^2\sqrt r)$, and
$\langle \sum_{i\in\cC_2}\mvxi_i,\mvxi_0\rangle=O_p(\sigma^2\sqrt{nr})$,
we obtain
\[
\sum_{i\in \cC_2}\|\mvxi_i-\mvxi_0\|_2^2
=
\frac{n}{2}\cdot 2r\sigma^2
+
O_p(\sigma^2\sqrt{nr})
+
O_p(n\sigma^2\sqrt r).
\]
Similarly,
\[
\sum_{i\in \cC_2}\langle \bmu_2-\bmu_1,\mvxi_i-\mvxi_0\rangle
=
\Big\langle \bmu_2-\bmu_1,\sum_{i\in\cC_2}\mvxi_i\Big\rangle
-\frac{n}{2}\langle \bmu_2-\bmu_1,\mvxi_0\rangle
=
\frac{n}{2}\,\Delta\sigma\,O_p(1).
\]

Combining the above bounds, we obtain
\begin{equation}
\sum_{i\in \cC_2} \|\by_i - \wh\bmu_1^{(0)}\|_2^2
=
\frac{n}{2}\Delta^2
+\frac{n}{2}\cdot 2r\sigma^2
+O_p\!\Big(n(\Delta\sigma+\Delta\epsilon+r\sigma^2+\epsilon^2)\Big).
\label{sumC2-perturb}
\end{equation}

\paragraph{Denominator (within-cluster term).}
For $i\in \cC_1\setminus\{\wh\bmu_1^{(0)}\}$,
\[
\by_i - \wh\bmu_1^{(0)} = (\mvxi_i-\mvxi_0)+(\mvdelta_i-\mvdelta_0),
\]
and hence
\begin{align}
\sum_{i\in \cC_1} \|\by_i - \wh\bmu_1^{(0)}\|_2^2
&=
\sum_{i\in \cC_1}\|(\mvxi_i-\mvxi_0)+(\mvdelta_i-\mvdelta_0)\|_2^2 \nonumber\\
&=
\frac{n}{2}\Big(
r\sigma^2 + \epsilon^2
+ r\sigma^2 O_p(1)
+ \epsilon\,O_p(\sigma\sqrt r)
\Big).
\label{sumC1-perturb}
\end{align}

\paragraph{Ratio bound.}
Let
\[
N_2 := \sum_{i\in \cC_2}\|\by_i-\wh\bmu_1^{(0)}\|_2^2,
\qquad
N_1 := \sum_{i\in \cC_1}\|\by_i-\wh\bmu_1^{(0)}\|_2^2 .
\]
Then
\[
1-\bbP(\cD\mid \cX,\wh\bmu_1^{(0)})
=
\frac{N_1}{N_1+N_2}
\le
\frac{N_1}{N_2}.
\]
Taking conditional expectation with respect to the randomness of $\wh\bmu_1^{(0)}$ given $\cX$ and using the tower property,
\[
1-\bbP(\cD\mid \cX)
=
\mathbb{E}\big[\,1-\bbP(\cD\mid \cX,\wh\bmu_1^{(0)})\,\big|\,\cX\big]
\le
\mathbb{E}\Big[\frac{N_1}{N_2}\Big|\,\cX\Big].
\]
Under the separation condition $\Delta^2 \geq  C (r\sigma^2+\epsilon^2)$ for some large $C$,
\eqref{sumC2-perturb} and \eqref{sumC1-perturb} imply
\[
\frac{N_1}{N_2}
=
O_p\!\left(\frac{r\sigma^2+\epsilon^2}{\Delta^2}\right).
\]
Plugging in, we have 
\[
1-\bbP(\cD\mid \cX)
=
O_p\!\left(\frac{r\sigma^2+\epsilon^2}{\Delta^2}\right),
\]
which completes the proof.

Next, let us prove the second assertion, namely \eqref{eqn:conc-center}. We start with the following preparatory lemma. 
\begin{lemma}
    \label{lem:gauss-conc}
    \begin{enumerate}[label = (\alph*), left=0pt ]
    \item Let $\mvxi_0\sim N_r(\mvzero, \sigma^2 I_r)$. Then for any $t> 0$, with $\Psi_r$ as in \eqref{eqn:gamma-r}, 
    \[\bbP\left(\left| \frac{||\mvxi_0||_2}{\sigma} - \Psi_r  \right| \geq t\right) \leq 2\exp\left(-\frac{t^2}{2}\right). \]
    \item Let $f_\sigma(\cdot)$ denote the density of $N_r(\mvzero, \sigma^2 I_r)$ and  suppose $\tilde{\mvxi_0}$ has density $\tilde{f}(\bx) \propto ||\bx||^2 f_\sigma(\bx)$, $\bx\in \bbR^r$. Then given any $\eta>0$, there exists $C_{\eta}<\infty$ such that for all $t> C_{\eta}$,  
    \[\bbP\left(\left| \frac{||\tilde{\mvxi_0}||_2}{\sigma} - \Psi_r  \right| \geq t\right) \leq 4(r)^{-1} \exp\left(-\frac{(1-\eta) t^2}{2}\right). \]
    \end{enumerate}
\end{lemma}
\begin{proof}
Part~(a) follows from standard Gaussian concentration for Lipschitz functions.
Indeed, $\|\cdot\|_2$ is $1$-Lipschitz and $\E\|\mvxi_0\|_2=\sigma\Psi_r$, so by
Gaussian concentration (e.g., \cite[Theorem~5.6]{boucheron2013concentration}),
\[
\bbP\!\left(\left|\frac{\|\mvxi_0\|_2}{\sigma}-\Psi_r\right|\ge t\right)
\le 2\exp\!\left(-\frac{t^2}{2}\right).
\]

We now prove part~(b). Let $X:=\mvxi_0/\sigma\sim N_r(\mvzero,I_r)$ and
$R:=\|X\|_2$, so that $\E R=\Psi_r$ and $\E R^2=r$.
By the definition of $\tilde{\mvxi}_0$, the law of
$\tilde X:=\tilde{\mvxi}_0/\sigma$ satisfies the size-biasing identity
\[
\bbP(\tilde X\in A)
= \frac{\E\!\left[R^2\mathbf 1\{X\in A\}\right]}{\E R^2}
= \frac{1}{r}\E\!\left[R^2\mathbf 1\{X\in A\}\right]
\]
for any measurable set $A\subset\bbR^r$.
In particular, for any $a\ge 0$,
\[
\bbP\!\left(\frac{\|\tilde{\mvxi}_0\|_2}{\sigma}>a\right)
= \frac{1}{r}\E\!\left[R^2\mathbf 1\{R>a\}\right].
\]

For a nonnegative random variable $R$, the tail-integration identity gives
\[
\E\!\left[R^2\mathbf 1\{R>a\}\right]
= a^2\bbP(R>a) + 2\int_a^\infty s\,\bbP(R>s)\,ds.
\]
Taking $a=\Psi_r+t$ and using part~(a),
\[
\bbP(R>s)
\le 2\exp\!\left(-\frac{(s-\Psi_r)^2}{2}\right),
\qquad s\ge 0.
\]
Therefore,
\[
\E\!\left[R^2\mathbf 1\{R>\Psi_r+t\}\right]
\le 2(\Psi_r+t)^2 e^{-t^2/2}
+ 4\int_{\Psi_r+t}^\infty s
\exp\!\left(-\frac{(s-\Psi_r)^2}{2}\right)ds.
\]
Making the change of variables $u=s-\Psi_r$ and using
$\int_t^\infty u e^{-u^2/2}du=e^{-t^2/2}$ together with
$\int_t^\infty e^{-u^2/2}du\le t^{-1}e^{-t^2/2}$ for $t\ge 1$, we obtain
\[
\E\!\left[R^2\mathbf 1\{R>\Psi_r+t\}\right]
\le C\big(1+(\Psi_r+t)^2\big)e^{-t^2/2}
\]
for a universal constant $C<\infty$.
Consequently,
\[
\bbP\!\left(\left|\frac{\|\tilde{\mvxi}_0\|_2}{\sigma}-\Psi_r\right|\ge t\right)
\le \frac{C}{r}\big(1+(\Psi_r+t)^2\big)e^{-t^2/2}.
\]

Finally, for any $\eta>0$ there exists $C_\eta<\infty$ such that
$1+(\Psi_r+t)^2 \le \exp(\eta t^2/2)$ for all $t>C_\eta$.
Thus, for $t>C_\eta$,
\[
\bbP\!\left(\left|\frac{\|\tilde{\mvxi}_0\|_2}{\sigma}-\Psi_r\right|\ge t\right)
\le 4r^{-1}\exp\!\left(-\frac{(1-\eta)t^2}{2}\right),
\]
which completes the proof.
\end{proof}

Now let us complete the proof the main Theorem. Without loss of generality, assume that the first initializer $\wh\bmu_1^{(0)}\in \cC_1$ (else run the argument below interchanging the roles of $\cC_1$ and $\cC_2$). Then by Lemma \ref{lem:gauss-conc}(a), we have that, 
\begin{align*}
    & \bbP\left(\left.\frac{\|\wh\bmu_1^{(0)}- \bmu_1\|_2}{\Delta} > \frac{1}{2} - \eta \right|\wh\bmu_1^{(0)}\in \cC_1\right) \\
    \leq & \bbP\left(\left.\frac{\|\mvxi_0 \|_2}{\Delta} > \frac{1}{2} - \eta - \frac{\epsilon}{\Delta} \right|\wh\bmu_1^{(0)}\in \cC_1\right) \\
    \leq & 2\exp\left(-\frac{1}{2}\left(\left(\frac{1}{2} -\eta - \frac{\epsilon}{\Delta} \right)\frac{\Delta}{\sigma} - \Psi_r\right)^2\right).       
\end{align*}
\label{eqn:838}

Fix $A>0$ and let $\cG_A$ (a mnemonic for ``good event'') denote the event, 
\[\cG_A:=\set{\wh\bmu_1^{(0)} \in \cC_1,~~ \wh\bmu_2^{(0)}\in \cC_2,~~ \|\wh\bmu_1^{(0)}\|_2 < A\sigma}.\]
Note that, even conditional on $\wh\bmu_2^{(0)} \in \cC_2$, $\wh\bmu_2^{(0)}$ is {\bf not} uniformly distributed as a point in the second cluster, since by the implementation of the $k$-means$++$ algorithm, the second cluster is ``biased'' to be far away from the first mean $\wh\bmu_1^{(0)}$.   
Conditional on the event $\mathcal D$ that the two initial seeds belong to different
ground-truth clusters, assume without loss of generality that
$\wh\bmu_1^{(0)}$ belongs to cluster $1$.
Then, for any $\eta>0$, standard empirical process theory \cite{pollard1990empirical} implies that for any fixed $A$ for the second center, 
\begin{align}
&\bbP\!\left(
\left.
\frac{\|\wh\bmu_2^{(0)}-\bmu_2\|_2}{\Delta}
>
\frac12-\eta
\;\right|\;
\wh\bmu_1^{(0)},\mathcal D,\mathcal G_A
\right)
\nonumber\\[4pt]
&\le
\frac{
\mathbb E\!\left[
\|\bmu_2+\mvxi+\mvdelta-\wh\bmu_1^{(0)}\|_2^{2}
\;
\ind\!\left\{
\frac{\|\mvxi\|_2}{\Delta}
>
\frac12-\eta-\frac{\epsilon}{\Delta}
\right\}
\;\Big|\;
\wh\bmu_1^{(0)}
\right]
}{
\mathbb E\!\left[
\|\bmu_2+\mvxi+\mvdelta-\wh\bmu_1^{(0)}\|_2^{2}
\;\Big|\;
\wh\bmu_1^{(0)}
\right]
}
+ o(1),
\label{eq:updated-step2} \\
&\le 
\frac{2
\mathbb E\!\left[
\|\bmu_2+\mvdelta-\wh\bmu_1^{(0)}\|_2^{2}
\;
\ind\!\left\{
\frac{\|\mvxi\|_2}{\Delta}
>
\frac12-\eta-\frac{\epsilon}{\Delta}
\right\}
\;\Big|\;
\wh\bmu_1^{(0)}
\right]
}{
\mathbb E\!\left[
\|\bmu_2+\mvxi+\mvdelta-\wh\bmu_1^{(0)}\|_2^{2}
\;\Big|\;
\wh\bmu_1^{(0)}
\right]
} \\
& +
\frac{2
\mathbb E\!\left[
\|\mvxi\|_2^{2}
\;
\ind\!\left\{
\frac{\|\mvxi\|_2}{\Delta}
>
\frac12-\eta-\frac{\epsilon}{\Delta}
\right\}
\;\Big|\;
\wh\bmu_1^{(0)}
\right]
}{
\mathbb E\!\left[
\|\bmu_2+\mvxi+\mvdelta-\wh\bmu_1^{(0)}\|_2^{2}
\;\Big|\;
\wh\bmu_1^{(0)}
\right]
} + o(1), 
\end{align}
where $\mvxi \sim N_r(\mvzero,\sigma^2 I_r)$ and $\mvdelta$ satisfies
$\|\mvdelta\|_2\le \epsilon$, independently of $\wh\bmu_1^{(0)}$.

Let
\[
t := \Big(\frac12-\eta\Big)\Delta_{\min}-\epsilon,
\qquad
E_t := \{\|\mvxi\|_2>t\},
\qquad
d := \|\bmu_2-\wh\bmu_1^{(0)}\|_2 .
\]
Since $\mvxi$ is independent of $\wh\bmu_1^{(0)}$ and $\|\mvdelta\|_2\le \epsilon$,
\begin{align*}
\mathbb E\!\left[
\|\bmu_2+\mvdelta-\wh\bmu_1^{(0)}\|_2^{2}\ind(E_t)
\;\Big|\;\wh\bmu_1^{(0)}
\right]
&\le (d+\epsilon)^2\,\mathbb P(E_t),
\end{align*}
Moreover,
\[
\mathbb E\!\left[
\|\bmu_2+\mvxi+\mvdelta-\wh\bmu_1^{(0)}\|_2^{2}
\;\Big|\;\wh\bmu_1^{(0)}
\right]
\ge d^2 + r\sigma^2 - 2d\epsilon
\ge (d-\epsilon)^2 + r\sigma^2.
\]
for some absolute constant $C>0$. Therefore,
\begin{align*}
&\bbP\!\left(
\left.
\frac{\|\wh\bmu_2^{(0)}-\bmu_2\|_2}{\Delta}
>
\frac12-\eta
\;\right|\;
\wh\bmu_1^{(0)},\mathcal D,\mathcal G_A
\right)\\
&\le
\frac{2(d+\epsilon)^2}{(d-\epsilon)^2 + r\sigma^2}\;\bbP(\|\mvxi\|_2>t)
+
\frac{2
\mathbb E\!\left[
\|\mvxi\|_2^{2}
\;
\ind\!\left\{
\frac{\|\mvxi\|_2}{\Delta}
>
\frac12-\eta-\frac{\epsilon}{\Delta}
\right\}
\;\Big|\;
\wh\bmu_1^{(0)}
\right]
}{(d-\epsilon)^2 + r\sigma^2} 
+o(1).
\end{align*}

Thus using Lemma \ref{lem:gauss-conc}(a) for the first term and (b) for the second term finally gives, 
\begin{align}
    & \bbP\left(\frac{||\wh\bmu_2^{(0)}- \bmu_2||_2}{\Delta} > \frac{1}{2} - \eta \bigg|\wh\bmu_1^{(0)}, \cG_A\right) \\
    \leq &  8/r \exp\left(-\frac{(1-\eta)}{2}\left((\frac{1}{2} -\eta - \frac{\epsilon}{\Delta} )\frac{\Delta}{\sigma} - \Psi_r\right)^2\right) \label{eqn:948}
\end{align}
Assuming $A> ||\bmu_1||_2/\sigma+ \Psi_r + \epsilon/\Delta$ and using Lemma \ref{lem:gauss-conc}(a) to bound the event $\bbP(||\wh\bmu_1^{(0)}||_2> A\sigma)$ and combining \eqref{eqn:838} and \eqref{eqn:948} finally gives the asserted bound in \eqref{eqn:conc-center}. 

\end{proof}

\section{Proofs of the applications}
\label{sec:proofs-applications}

This section contains proofs of all applications, as described in Section \ref{s:applications}. 

\subsection{Spectral clustering in stochastic block models}

The proof of Theorem~\ref{thm:sbm_K} proceeds in three stages.  
Section~\ref{SBM_notation} introduces the notation used throughout the proof. First, Lemmas~\ref{lem:SBM}--\ref{Lemma_norm_SBM} in Section~\ref{lem:concentration_SBM} establish concentration properties for the Bernoulli entries in the SBM.  
Second, Lemmas~\ref{Lemma_Gamma_s_SBM}--\ref{Lemma_G_s_SBM} in Section~\ref{subsec:one-step_SBM} control one Lloyd update in terms of $\Gamma_s$ and $G_s$. Finally, in Section~\ref{main_proof_sbm}, we combine these ingredients and prove Theorem~\ref{thm:sbm_K} by iterating the Lloyd recursion.

\subsubsection{Notation}\label{SBM_notation}
Recall the spectral decompositions of the adjacency matrix \(\Ab\) and its population counterpart \(\Ab^*\):
\[
\Ab = \sum_{i=1}^n \lambda_i \bu_i \bu_i^{\top}, 
\qquad
\Ab^* = \sum_{i=1}^K \lambda_i^* \bu_i^* (\bu_i^*)^{\top},
\]
where \(\lambda_1 \ge \lambda_2 \ge \cdots \ge \lambda_n\) and
\(\lambda_1^* \ge \lambda_2^* \ge \cdots \ge \lambda_K^*\) denote the ordered eigenvalues of
\(\Ab\) and \(\Ab^*\), respectively. 
Define
\(\Ub^* = (\bu_1^*, \bu_2^*, \ldots, \bu_K^*)\) and
\(\mvLambda^* = \mathrm{diag}(\lambda_1^*, \lambda_2^*, \ldots, \lambda_K^*)\).

The empirical spectral embedding $\Ub$ admits a decomposition analogous to the general model in \eqref{eqn:model_matrix}, up to an orthogonal rotation:
\begin{equation}
\Ub\Ob = \Ub^* + \left[\Ub\Ob-\Ab\Ub^*(\mvLambda^{*})^{-1}\right] + \Eb \Ub^*(\mvLambda^{*})^{-1}=:\Ub^*+\Eb_1+\Eb_2,
\end{equation}
where $\Ob\in \bbR^{K\times K}$ is an orthogonal matrix.
Let $n_k$, for $k=1, \ldots, K$ denote the number of nodes in community $k$. We define the \(\ell_{2 \to \infty}\) distance between two matrices
\(\Ub, \Ub^* \in \mathbb{R}^{n \times K}\) as
\[
d_{2 \to \infty}(\Ub, \Ub^*)
\triangleq
\inf_{\Ob \in \mathbb{R}^{K \times K} : \Ob^{\top}\Ob = \Ib}
\|\Ub \Ob - \Ub^*\|_{2 \to \infty},
\]
where \(\|\Vb\|_{2 \to \infty} = \max_{i \in [n]} \|\Vb_i\|_2\) and
\(\Vb_i\) denotes the \(i\)-th row of \(\Vb\).

\subsubsection{Technical concentration lemmas}\label{lem:concentration_SBM}
Applying Corollary~3.6 of \cite{lei2019unified}, we obtain the following result.
\begin{lemma}
\label{lem:SBM}
Assume the conditions in Assumption \ref{asm:SBM} hold. Then
\[
d_{2 \to \infty}\!\left(\Ub, \Ab \Ub^* (\mvLambda^*)^{-1}\right)
= \Theta\!\left(\frac{1}{n \sqrt{\rho_n}}\right).
\]
\end{lemma}

\begin{proof}
Without loss of generality, assume that
$$
\Zb=\left[\begin{array}{cccc}
\mathbf{1}_{n_1} & 0 & \cdots & 0 \\
0 & \mathbf{1}_{n_2} & \cdots & 0 \\
\vdots & \vdots & \ddots & \vdots \\
0 & 0 & \cdots & \mathbf{1}_{n_K}
\end{array}\right]
$$
Let $\Mb=\operatorname{diag}\left(\sqrt{n_1}, \ldots, \sqrt{n_K}\right)$ and $\Qb=\Zb \Mb^{-1}$. Then $\Qb^{\top} \Qb=\Ib$ and
$$
\Ab^*=\Qb(\Mb \Bb \Mb) \Qb^{\top} .
$$
Let $\Vb \bLambda^* \Vb^{\top}$ be the spectral decomposition of $\Mb\Bb\Mb$. Then $\Qb \Vb \bLambda^* (\Qb \Vb)^{\top}$ is the spectral decomposition of $\Ab^*$ because $\Qb \Vb$ is an orthogonal matrix. As a result, the eigenvector matrix of $\Ab^*$ is $\Ub^*=\Qb \Vb$. By definition,
$$
\Ub^*=\left[\begin{array}{c}
\frac{\mathbf{1}_{n_1} \Vb_1^{\top}}{\sqrt{n_1}} \\
\frac{\mathbf{1}_{n_2} \Vb_2^{\top}}{\sqrt{n_2}} \\
\vdots \\
\frac{\mathbf{1}_{n_K} \Vb_K^{\top}}{\sqrt{n_K}}
\end{array}\right],
$$
where $\Vb_i^{\top}$ is the $i$-th row of $\Vb$. 
We have $\|\Ub^*\|_{2\rightarrow \infty} = \Theta(\frac{1}{\sqrt{n}})$. 
Let $c(\cdot):[n]\rightarrow [K]$ be the membership vector and $\cC_s = \{i: c(i) = s\}$ for $s\in[K]$. 
Let $\vb_s^*=\Ub_i^*$ for $i \in \mathcal{C}_s$. Using the fact that $\Vb$ is an orthogonal matrix, we have
$$
\left\|\vb_s^*-\vb_{s^{\prime}}^*\right\|_2 = \sqrt{\left\|\vb_s^*\right\|_2^2+\left\|\vb_{s^{\prime}}^*\right\|_2^2-2\left\langle \vb_s^*, \vb_{s^{\prime}}^*\right\rangle}=\sqrt{\frac{1}{n_s}+\frac{1}{n_{s^{\prime}}}} = \Theta(\frac{1}{\sqrt{n}}).
$$
Therefore, 
$$\Delta:= \min_{s\neq s^{\prime}}\|\vb_s^*-\vb_{s^{\prime}}^*\| = \Theta(\frac{1}{\sqrt{n}}).$$
As shown in the proof of Theorem 5.2 of \cite{lei2019unified}, we have  
$\bar{\kappa}^* \leq 2 K \preceq 1, \Delta^* = \Theta(n\rho_n)$ and
$$
R(\delta) \preceq \log n, \quad g(\delta) \preceq \sqrt{n \rho_n}+\frac{\log n}{\log \log n}.
$$
Combined with Assumption \ref{asm:SBM} (b), we have 
$$
\lambda_{\min}(\Ab^*) \succeq \frac{n \rho_n}{\sqrt{n}\left\|\Ub^*\right\|_{2 \rightarrow \infty}}.
$$
We can therefore apply Corollary~3.6 of \cite{lei2019unified} to obtain the desired bound
\[
d_{2\rightarrow\infty}\!\left(\Ub,\,\Ab\Ub^*(\mvLambda^*)^{-1}\right).
\]

\end{proof}

The above lemma implies that there exists an orthogonal matrix $\Ob\in \bbR^{K\times K}$ such that 
$$
 \epsilon :=  \left\| \Eb_1 \right\|_{2 \rightarrow \infty} = \left\|\Ub \Ob -\Ab\Ub^*(\mvLambda^{*})^{-1}\right\|_{2 \rightarrow \infty} =  \Theta(\frac{1}{n\sqrt{\rho_n}}).
$$

\begin{proof}

{\bf Concentration results on sub-Gaussian noise.}
Under the stochastic block model with independent Bernoulli entries, we can show that Lemma \ref{lem:subg-sum} -- \ref{lem:indicator-bernstein} can be improved to achieve a tight concentration result. We adopt the notation used in the proof of Lemma \ref{lem:SBM}. Without loss of generality, assume that
$$
\Zb=\left[\begin{array}{cccc}
\mathbf{1}_{n_1} & 0 & \cdots & 0 \\
0 & \mathbf{1}_{n_2} & \cdots & 0 \\
\vdots & \vdots & \ddots & \vdots \\
0 & 0 & \cdots & \mathbf{1}_{n_K}
\end{array}\right],
$$ 
and define $N_h := \sum_{l=1}^h n_l$. We first give a detailed characterization of the error matrix $\Eb_2= \Eb\Ub^*(\bLambda^*)^{-1}$ by writing  
$$
\bw_i := \Eb_{2,i} = (\Eb\Ub^*(\bLambda^*)^{-1})_i = (\Eb_i^{\top}\Ub^*(\bLambda^*)^{-1})^{\top},
$$
where $\Eb_i$ is the $i$-th row of $\Eb$. 
For any fixed $l\in[K]$ and $i\in[n]$, we have 
\begin{equation}\label{eqn:SBM_error}
\bw_{il} = \Eb_i^{\top}\bu_l^*/\lambda^*_l = \frac{1}{\lambda^*_l}\sum_{s=1}^K \sum_{j\in \cC_s}\Eb_{ij}\bv_{sl}
\end{equation}
where $\bv_s$ is defined in Lemma \ref{lem:SBM} as $\bv_s^*=\Ub_i^*$ for $i \in \cC_s$. From the proof of Lemma \ref{lem:SBM}, 
$$
\Delta_{\min} = \min_{s\neq s^{\prime}}\|\vb_s^*-\vb_{s^{\prime}}^*\| = \min_{s\neq s^{\prime}} \sqrt{\frac{1}{n_s}+\frac{1}{n_{s^{\prime}}}} \geq \sqrt{\frac{2}{\max_s n_s}} \geq \sqrt{\frac{2}{n}},
$$
and 
$$
M = \max_{s\neq s^{\prime}}\|\vb_s^*-\vb_{s^{\prime}}^*\| = \max_{s\neq s^{\prime}} \sqrt{\frac{1}{n_s}+\frac{1}{n_{s^{\prime}}}} \leq \sqrt{\frac{2}{\min_s n_s}} \leq \sqrt{\frac{2}{n\alpha}}. 
$$
Therefore 
\begin{equation}\label{eqn:M_Delta_bound}
    \frac{\Delta_{\max}}{\Delta_{\min}} \leq \sqrt{\frac{\max_s n_s}{\min_s n_s}} \leq \sqrt{\frac{C_1}{c_1}}.  
\end{equation}

The following lemma is the key observation used to improve the concentration result under SBMs, following from Theorem 5.2 in \cite{lei2015consistency}.

\begin{lemma}\label{concentratiob_SBM}
Let $\Ab$ be the adjacency matrix of a random graph on $n$ nodes in which edges occur independently. Set $\mathbb{E}[\Ab]=\Pb =\left(p_{i j}\right)_{i, j=1, \ldots, n}$ and assume that $n \max _{i j} p_{i j} \leq d$ for $d \geq c_0 \log n$ and $c_0>0$. Then, for any $r>0$ there exists a constant $C=C\left(r, c_0\right)$ such that
$$
\|\Ab-\Pb\| \leq C \sqrt{d}
$$
with probability at least $1-n^{-r}$.
\end{lemma}

Fixing $r=3$ and using the fact that $\lambda_{\min}(\bLambda^*) = \lambda_{\min}(\Ab^*) = \Theta(n\lambda_{\min}(\Bb^*)) = \Theta(n\rho_n)$, we have 
$$
\|\Eb_2\| = \|\Eb \Ub^*(\bLambda^*)^{-1}\| \leq \|\Eb\|\|(\bLambda^*)^{-1}\| \lesssim \frac{\sqrt{n\rho_n}}{n\rho_n} \lesssim \frac{1}{\sqrt{n\rho_n}},
$$
that is 
$$
\|\Eb_2\|  \leq \frac{C_1}{\sqrt{n\rho_n}}
$$
for some constant $C_1$, with probability at least $1-n^{-4}$. Let $S \subset [n]$ and define $\Wb_S = \sum_{i \in S} \bw_i$.
     
\begin{lemma}\label{Lemma_1_SBM}
$\|\Wb_S\|_2 \leq  \dfrac{C_1\sqrt{|S|}}{\sqrt{n\rho_n}} $ for all $S \subset [n]$ with probability at least $1-n^{-3}$. 
\end{lemma}
        
{\bf Proof: }
$$
    \left\|\frac{1}{\sqrt{|S|}}\Wb_S\right\| =  \left\|\frac{1}{\sqrt{|S|}}\bone_{S}^{\top}\Eb_2\right\| \leq \|\Eb_2\| \leq \frac{C_1}{\sqrt{n\rho_n}}
$$
with probability at least $1-n^{-3}$. 
$\blacksquare$
\begin{lemma}\label{Lemma_2_SBM} For all  $S\subset [n]$,
    $$\lambda_{\max}\left( \sum_{i\in S} \bw_i \bw_i^{\top} \right) \leq \frac{C_1^2}{n\rho_n},$$ with probability at least $1-n^{-4}$. 
\end{lemma}

\noindent{\bf Proof: }
$$\left\| \sum_{i\in S} \bw_i \bw_i^{\top} \right\| \leq \left\| 
\Eb_2^{\top} \operatorname{diag}\left( \bone_{S} \right)\Eb_2 \right\| \leq \left\| 
\Eb_2^{\top} \right\|_2^2 \leq \frac{C_1^2}{n\rho_n}, $$
with probability at least $1-n^{-4}$
$\blacksquare$
    
\begin{lemma}\label{Lemma_4_SBM} 
For all  $h\subset [K]$,   $$\left\|\Wb_{\mathcal{C}_h}\right\|_2 \leq C_2\frac{\sqrt{n_h\log n}}{n\sqrt{\rho_n}} $$ 
with probability $1-n^{-3}$. 
\end{lemma}

\begin{proof}
Fix $h\in [K]$, 
$$
\left\|\sum_{i\in\cC_h }\bw_{i} \right\|^2 =  \sum_{l=1}^K\left(\sum_{i\in\cC_h }\bw_{il}\right)^2
$$
and from equation \eqref{eqn:SBM_error} 
\begin{align}
\sum_{i\in\cC_h }\bw_{il} &= 
\sum_{i\in\cC_h } \frac{1}{\lambda^*_l} 
\sum_{s=1}^K \sum_{j\in \cC_s}\Eb_{ij}\bv_{sl}\nonumber\\ 
& = \frac{1}{\lambda^*_l} 
\sum_{i\in \cC_h }\left( 
\sum_{j\in \cC_h}\Eb_{ij}\bv_{hl} + \sum_{s\neq h}\sum_{j\in \cC_s}\Eb_{ij}\bv_{sl}
\right) \nonumber\\
& = \frac{1}{\lambda^*_l} \left(
\sum_{i=N_{h-1}+1}^{N_h}
\sum_{j=N_{h-1}+1}^{N_h}\Eb_{ij}\bv_{hl} + \sum_{i\in\cC_h }\sum_{s\neq h}\sum_{j\in \cC_s}\Eb_{ij}\bv_{sl}
\right) \nonumber
\\
& = \frac{1}{\lambda^*_l} \left(2
\sum_{i=N_{h-1}+1}^{N_h}
\sum_{j=i+1}^{N_h}\Eb_{ij}\bv_{hl} + \sum_{i\in\cC_h }\sum_{s\neq h}\sum_{j\in \cC_s}\Eb_{ij}\bv_{sl}
\right) \label{eqn:0314}
\end{align}
To bound the first part of \eqref{eqn:0314}, using Bernstein inequality, for fixed $h,l\in[K]$
\begin{align*}
& \bbP\left( 
\left|
\sum_{i=N_{h-1}+1}^{N_h}
\sum_{j=i+1}^{N_h}\Eb_{ij}\bv_{hl} \right| \geq t \right) \\
\leq & 2\exp\left( -\frac{\dfrac{1}{2}\dfrac{1}{\bv_{hl}^2}t^2}{
p\dfrac{n_h(n_h-1)}{2}+ \dfrac{1}{3\bv_{hl}}t } \right).
\end{align*}
Choosing 
$t = \max\left\{ \dfrac{16}{3}\bv_{hl}\log n, 2\bv_{hl}\sqrt{2pn_h(n_h-1)\log n}\right\} = 2\bv_{hl}\sqrt{2pn_h(n_h-1)\log n}$ under the assumption $p \geq \frac{1}{n}$,
$$
\left|2
\sum_{i=N_{h-1}+1}^{N_h}
\sum_{j=i+1}^{N_h}\Eb_{ij}\bv_{hl} \right| \leq 4\bv_{hl}\sqrt{2pn_h(n_h-1)\log n}, 
$$
with probability at least $1-n^{-4}$. Similarly, we can bound the second part of \eqref{eqn:0314} using Bernstein's inequality
\begin{align*}
& \bbP\left( 
\left|
\sum_{i\in\cC_h }\sum_{s\neq h}\sum_{j\in \cC_s}\Eb_{ij}\bv_{sl} \right| \geq t \right) \\
\leq & 2\exp\left( -\frac{\frac{1}{2}t^2}{
pn_h + \frac{1}{3}\frac{t}{\sqrt{n\alpha} } } \right),
\end{align*}
where the inequality uses that fact that 
$$
\var\left(\sum_{i\in\cC_h }\sum_{s\neq h}\sum_{j\in \cC_s}\Eb_{ij}\bv_{sl} \right) \leq \sum_{i\in\cC_h }\sum_{s\neq h}\sum_{j\in \cC_s}p\bv_{sl}^2 = \sum_{i\in\cC_h }\sum_{s\neq h} p\bv_{sl}^2n_s \leq \sum_{i\in\cC_h } p = pn_h,
$$
since $\sum_{s\neq h}\bv_{sl}^2 n_s \leq \sum_{s=1}^K\bv_{sl}^2 n_s = 1$ and $\left| \Eb_{ij}\bv_{sl}\right| \leq |\bv_{sl}| \leq \frac{1}{\sqrt{n_s}} \leq \frac{1}{\sqrt{n\alpha}}$. 
Choosing 
$$
t = \max\left\{ \dfrac{8}{3}\dfrac{\log n}{\sqrt{n\alpha}}, 4\sqrt{pn_h \log n}\right\} = 4\sqrt{pn_h \log n}
$$ under the assumption $p \geq \frac{1}{n}$, and then we have for fixed $h,l\in[K]$
$$
\left|
\sum_{i\in\cC_h }\sum_{s\neq h}\sum_{j\in \cC_s}\Eb_{ij}\bv_{sl} \right| \leq 4\sqrt{pn_h \log n}, 
$$
with probability at least $1-n^{-4}$. Combining two parts together and a union bound 
\begin{align*}
\left| \sum_{i\in\cC_h }\bw_{il} \right| \leq& \frac{4}{\lambda_l^*} \left( \bv_{hl}\sqrt{2pn_h(n_h-1)\log n} + \sqrt{pn_h \log n} \right) \\
= & \frac{4}{\lambda_l^*} \sqrt{pn_h\log n} \left( \bv_{hl}\sqrt{2n_h} + 1 \right) ,
\end{align*}
for any fixed $h,l\in[K]$ with probability  $1-n^{-3}$ and 
\begin{align*}
\left\|\sum_{i\in\cC_h }\bw_{i} \right\|^2 = \sum_{l=1}^K \left(\sum_{i\in\cC_h }\bw_{il}\right)^2 &\leq \sum_{l=1}^K \left[
\frac{4}{\lambda_l^*} \sqrt{pn_h\log n} \left( \bv_{hl}\sqrt{2n_h} + 1 \right)
\right]^2 \\
& \leq 32(2+K)pn_h\log n /(\lambda_K^{*})^2,
\end{align*}
for any fixed $h\in[K]$ with probability at least $1-n^{-3}$ .
The last inequality uses the facts that $(a+b)^2\leq 2(a^2+b^2)$, $\lambda_l^* \geq \lambda_K^*$ and  $ \sum_{l=1}^K  \bv_{hl}^2 = \dfrac{1}{n_h}$ for fixed $h\in[K]$. 
Using the definition of $\rho_n$ and $\lambda_K^* = \Theta(n\rho_n)$, we finally have 
$$
\left\|\sum_{i\in\cC_h }\bw_{i} \right\| \leq C_2\frac{\sqrt{n_h\log n}}{n\sqrt{\rho_n}},
$$
with probability $1-n^{-3}$. 
\end{proof}

\begin{lemma}\label{Lemma_5_SBM}
Let $g,h \in [K]$ such that $g\neq h$. Then, for any $a>0$,  
\begin{align*}
& \sum_{i\in \cC_g} \cI\left( a\|\bmu_g - \bmu_h\|^2  \leq
\langle \bw_i, \bmu_h - \bmu_g\rangle \right) \\
\leq & 
2n_g\exp\left( -C_3a^2n\rho_n \right) + n_g\exp\left( -C_4a^2n\rho_n \right) + 16\log n \\
& + 4\sqrt{n_g\log n}\left\{2\exp\left( -\frac{C_3}{2}a^2n\rho_n\right) + \exp\left( -\frac{C_4}{2}a^2n\rho_n\right) \right\}
\end{align*}
for some constants $C_3$ and $C_4$ with probability $1 - n^{-3}$.
\end{lemma}    

\begin{proof}
Fix $g,h\in[K]$ and $a>0$. 
By \eqref{eqn:SBM_error} and the definition of $\bv_s$ for $s\in[K]$, we have 
\begin{align*}
& \cI\left( a\|\bmu_g - \bmu_h\|^2  \leq 
\langle \bw_i, \bmu_h - \bmu_g \rangle \right)
= \cI\left( a\|\bv_g - \bv_h\|^2  \leq 
\langle \bw_i, \bv_g - \bv_h \rangle \right) \\
= & \cI\left( a\|\bv_g - \bv_h\|^2  \leq 
\sum_{l=1}^K \frac{1}{\lambda^*_l}\sum_{s=1}^K \sum_{j\in \cC_s}\Eb_{ij}\bv_{sl}(\bv_{gl} - \bv_{hl}) \right) \\ 
\leq & \cI\left( a_1\|\bv_g - \bv_h\|^2  \leq 
\sum_{l=1}^K \frac{1}{\lambda^*_l} \sum_{j\in \cC_g}\Eb_{ij}\bv_{gl}(\bv_{gl} - \bv_{hl}) \right) \\
& + \cI\left( a_2\|\bv_g - \bv_h\|^2  \leq 
\sum_{l=1}^K \frac{1}{\lambda^*_l}\sum_{s\neq g} \sum_{j\in \cC_s}\Eb_{ij}\bv_{sl}(\bv_{gl} - \bv_{hl}) \right)\\ 
\leq & \cI\left( \frac{a_1}{2}\|\bv_g - \bv_h\|^2  \leq 
\sum_{l=1}^K \frac{1}{\lambda^*_l} \sum_{j=N_{g-1}+1}^{i-1}\Eb_{ij}\bv_{gl}(\bv_{gl} - \bv_{hl}) \right) \\
& +\cI\left( \frac{a_1}{2}\|\bv_g - \bv_h\|^2  \leq 
\sum_{l=1}^K \frac{1}{\lambda^*_l} \sum_{j=i+1}^{N_g}\Eb_{ij}\bv_{gl}(\bv_{gl} - \bv_{hl}) \right) \\
& + \cI\left( a_2\|\bv_g - \bv_h\|^2  \leq 
\sum_{l=1}^K \frac{1}{\lambda^*_l}\sum_{s\neq g} \sum_{j\in \cC_s}\Eb_{ij}\bv_{sl}(\bv_{gl} - \bv_{hl}) \right) 
=: X_i + Y_i + Z_i,
\end{align*}
where $a=a_1+a_2$ and the last inequality guarantee independence between different components when we take summation over $i\in\cG_g$. 
It is straightforward to verify that the expectation for each term can be bounded as 
\begin{align*}
    \bbE\left( X_i \right) \leq \exp\left( -C_3a_1^2n\rho_n \right),\\
    \bbE\left( Y_i \right) \leq \exp\left( -C_3a_1^2n\rho_n \right),\\
    \bbE\left( Z_i \right) \leq \exp\left( -C_4a_2^2n\rho_n \right),
\end{align*}
for some constants $C_3$ and $C_4$. By Bernstein's inequality,
\begin{align*}
    \bbP \left( \sum_{i\in\cC_g}(X_i-\bbE(X_i))>t \right) &\leq \exp\left(
    -\dfrac{\frac{1}{2}t^2}{\sigma_{X}^2 + \frac{1}{3}t}
    \right),
\end{align*}
where $\sigma_{X}^2 = \var\left(\sum_{i\in\cC_g} X_i\right) \leq n_g\exp\left( -C_3a_1^2n\rho_n \right)$. Choose \\
$t_0 = \max\{\frac{16}{3}\log n, 4\sigma_{X}\sqrt{\log n}\}$, we have 
$\sum_{i\in\cC_g}\left(X_i-\bbE(X_i) \right) \leq t_0$ with probability at least $1-n^{-4}$. Therefore,
$$
\sum_{i\in\cC_g}X_i \leq n_g\exp\left( -C_3a_1^2n\rho_n \right) +  \frac{16}{3}\log n + 4\exp\left( -\frac{C_3}{2}a_1^2n\rho_n\right)\sqrt{n_g\log n},
$$
with probability at least $1-n^{-4}$. 
Similarly, we have 
$$
\sum_{i\in\cC_g} Y_i \leq n_g\exp\left( -C_3a_1^2n\rho_n \right) +  \frac{16}{3}\log n + 4\exp\left( -\frac{C_3}{2}a_1^2n\rho_n\right)\sqrt{n_g\log n},
$$
with probability at least $1-n^{-4}$, and  
$$
\sum_{i\in\cC_g} Z_i \leq n_g\exp\left( -C_4a_2^2n\rho_n \right) + \frac{16}{3}\log n + 4\exp\left( -\frac{C_4}{2}a_2^2n\rho_n  \right)\sqrt{n_g\log n} ,
$$
with probability at least $1-n^{-4}$. 
In summary, combining three parts together,
\begin{align*}
&  \sum_{i\in \cC_g} \cI\left( a\|\bmu_g - \bmu_h\|^2  \leq
        \langle \bw_i, \bmu_h - \bmu_g\rangle \right)
\leq \sum_{i\in \cC_g}(X_i+ Y_i+ Z_i) \\
\leq & 2n_g\exp\left( -C_3a_1^2n\rho_n \right) + n_g\exp\left( -C_4a_2^2n\rho_n \right) + 16\log n \\
& + 4\sqrt{n_g\log n}\left\{2\exp\left( -\frac{C_3}{2}a_1^2n\rho_n\right) + \exp\left( -\frac{C_4}{2}a_2^2n\rho_n\right) \right\}
\end{align*}
 with probability at least $1 - n^{-3}$. 

\end{proof}

Based on the formulation of $\bw_i$, we have the following technical lemma, which will be useful throughout the proof.

\begin{lemma}\label{Lemma_SBM}
Let $g,h \in [K]$ such that $g\neq h$. Then, for any $a>0$, $i\in\cC_g$
$$
\bbP\left(  a\|\bmu_g - \bmu_h\|^2 \leq \langle \bw_i, \bmu_h - \bmu_g\rangle \right) \leq \exp(-C_5a^2n\rho_n),
$$
for some constant $C_5$.
\end{lemma}

\begin{proof}
Fix $i\in\cC_g$.  
\begin{align*}
& \bbP\left(  a\|\bmu_g - \bmu_h\|^2 \leq \langle \bw_i, \bmu_h - \bmu_g\rangle \right) \\
= & \bbP\left(  a\|\bv_g - \bv_h\|^2 \leq \sum_{l=1}^K \bw_{il}(\bv_{gl} - \bv_{hl}) \right) \\
= & \bbP\left( a\|\bv_g - \bv_h\|^2  \leq 
\sum_{l=1}^K \frac{1}{\lambda^*_l}\sum_{s=1}^K \sum_{j\in \cC_s}\Eb_{ij}\bv_{sl}(\bv_{gl} - \bv_{hl})  \right) \\
\leq& \exp\left(-C_5a^2n\rho_n \right),
\end{align*} 
for some constant $C_5$ using Bernstein inequality. 
\end{proof}

\begin{lemma}\label{Lemma_norm_SBM}
For all $i\in [n]$, we have 
$$\|\bw_i\| \leq C_6 \max\left\{ \frac{\log n}{n\sqrt{n\alpha}\rho_n}, \frac{1}{n}\sqrt{\frac{\log n}{\rho_n}}\right\},$$
for some constant $C_6$ with probability at least $1-n^{-3}$. 
\end{lemma}

\begin{proof}
Fix $i\in [n] $ and $t>0$, by Bernstein's inequality, 
\begin{align*}
    \bbP\left( t\leq \|\bw_i\| \right) 
    & = \sum_{l=1}^K  \bbP\left( \frac{t}{\sqrt{K}}\leq |\bw_{il}| \right)  \\
    &\leq \sum_{l=1}^K  \bbP\left( \frac{t}{\sqrt{K}}\leq \left|\frac{1}{\lambda^*_l}\sum_{s=1}^K \sum_{j\in \cC_s}\Eb_{ij}\bv_{sl}\right| \right)\\
    &\leq \sum_{l=1}^K \exp\left( -\dfrac{\frac{1}{2}\frac{t^2}{K}}{\frac{p}{(\lambda^*_l)^2}+\frac{1}{3}\frac{t}{\sqrt{K}}\frac{1}{\sqrt{n\alpha}\lambda_l^*}} \right)\\
    &\leq K \exp\left( -\dfrac{\frac{1}{2}\frac{t^2}{K}}{\frac{p}{(\lambda^*_l)^2}+\frac{1}{3}\frac{t}{\sqrt{K}}\frac{1}{\sqrt{n\alpha\lambda_K^*}}} \right)\\
    &\leq K \exp\left( -\dfrac{\frac{1}{2}\frac{t^2}{K}}{\frac{p}{(\lambda^*_K)^2}+\frac{1}{3}\frac{t}{\sqrt{K}}\frac{1}{\sqrt{n\alpha}\lambda_K^*}} \right)
\end{align*}
using the fact that 
$$
\var(\frac{1}{\lambda^*_l}\sum_{s=1}^K \sum_{j\in \cC_s}\Eb_{ij}\bv_{sl}) \leq \frac{p}{(\lambda^*_l)^2}
$$
and 
$$
\left| \frac{\bv_{sl}}{\lambda^*_l} \right| \leq \frac{1}{\sqrt{n_s}}\frac{1}{\lambda^*_l} \leq  \frac{1}{\sqrt{n\alpha}}\frac{1}{\lambda^*_l}.
$$
Choosing $t_0 = C_6\max\left\{\frac{1}{n}\sqrt{\frac{\log n}{\rho_n}} ,  \frac{\log n}{n\sqrt{n\alpha}\rho_n}\right\}$ for some constant $C_6$. Then
$$
\bbP\left( t_0\leq \|\bw_i\| \right) \leq \frac{1}{n^4}.
$$
Applying a union bound over $i\in[n]$ yields the desired result.
\end{proof}

In summary, under SBMs, we can view $\sigma_{\operatorname{approx}} = \Theta(\dfrac{1}{n\sqrt{\rho_n}})$ instead of the sub-Gaussian parameter $\sigma = \Theta(\dfrac{1}{n\rho_n})$ used in the general theorem. Under the SBM with $K=2$, we assume 
\begin{equation}
\epsilon=\dfrac{C_0}{n\sqrt{\rho_n}},\label{eqn:epsilon}    
\end{equation}
for some constant $C_0$, by Lemma \ref{lem:SBM}. Define 
\begin{equation}
\rho_{\sigma} = \frac{1}{C_1}\sqrt{n\alpha\rho_n}, \label{rho_sigma_SBM_K_2}
\end{equation}
and 
\begin{equation}
\rho_{\epsilon} = \dfrac{2}{C_0}\sqrt{n\alpha\rho_n}.\label{rho_epsilon_SBM_K_2}
\end{equation} 
We will first control $G_s$ and $\Gamma_s$, which will then allow us to control $A_s$ as $s$ grows. 

\subsubsection*{One-step bounds for Lloyd's iterates under SBMs}
\label{subsec:one-step_SBM}

\noindent{\bf Lemma on center error recursion.}

\begin{lemma}\label{Lemma_Gamma_s_SBM}
Conditionally on the events in Lemmas~\ref{Lemma_1_SBM}, \ref{Lemma_2_SBM},
and \ref{Lemma_4_SBM}, if $G_s \leq \frac{1}{2}$, then
\begin{equation*}
\Gamma_s \leq \dfrac{\epsilon}{\Delta_{\min}}
+ \min \left( 2G_s\Gamma_{s-1} + (1+\sqrt{2})\frac{\sqrt{KG_s}}{\rho_{\sigma}} + \frac{\sqrt{2}C_2}{C_1} \frac{1}{\rho_{\sigma}} \sqrt{\dfrac{\log n}{n}}, \frac{1}{\rho_{\sigma}} + \sqrt{\frac{C_1}{c_1}} G_s  \right).
\end{equation*}
\end{lemma}

\begin{proof}
In order to bound $\Gamma_s$, we need to bound $\|\wh\bmu\s_h -\bmu_h\|$, which we do so expanding $\wh\bmu\s_h$ using $\{i: \hat{z}\s_i = h\} = \bigcup_{g\in [K] } U_{gh}\s$ as follows:
\begin{equation}\label{eqn:1503}
\wh\bmu\s_h -\bmu_h = \dfrac{1}{\nhat \s_h} \sum_{i\in U\s_{hh}} (\by_i -\bmu_h) + \sum_{g\neq h} \dfrac{\wh n\s_{gh}}{\nhat_h\s} (\wb \by_{U\s_{gh}} - \bmu_h),
\end{equation}
where $\wb \by_{U\s_{gh}} = \frac{1}{\wh n\s_{gh}}\sum_{i \in U\s_{gh}} \by_i$. By Lloyd's algorithm, for $i \in U\s_{gh}$, $\|\by_i - \wh\bmu^{(s-1)}_h \| \leq \|\by_i - \wh\bmu^{(s-1)}_g\|$. Thus, it follows that $\|\wb\by_{U\s_{gh}}  - \wh\bmu^{(s-1)}_h\| \leq \|\wb\by_{U\s_{gh}}  - \wh\bmu^{(s-1)}_g\|$. Repeated applications of the triangle inequality yield
\begin{align*}
\| \wb\by_{U\s_{gh}} - \bmu_h\| &\leq \|\wb\by_{U\s_{gh}}  - \wh\bmu^{(s-1)}_h\| + \| \wh\bmu^{(s-1)}_h - \bmu_h\| \\
& \leq \|\wb\by_{U\s_{gh}}  - \wh\bmu^{(s-1)}_g\| + \| \wh\bmu^{(s-1)}_h - \bmu_h\|  \\
& \leq \|\wb\by_{U\s_{gh}}  - \bmu_g \| + \| \bmu_g - \wh\bmu^{(s-1)}_g\| + \| \wh\bmu^{(s-1)}_h - \bmu_h\|  \\
& \leq \|\wb\by_{U\s_{gh}} - \wb\by^*_{U\s_{gh}}\| +  \|\wb\by^*_{U\s_{gh}}  - \bmu_g \| + \| \bmu_g - \wh\bmu^{(s-1)}_g\| + \| \wh\bmu^{(s-1)}_h - \bmu_h\| \\ 
& \leq \epsilon + \dfrac{C_1}{\sqrt{n\rho_n }\sqrt{\nhat\s_{gh}}} + 2\Gamma_{s-1}\Delta_{\min},
\end{align*}
where the last inequality uses Lemma \ref{Lemma_1_SBM}, the definition of $\epsilon$ bound, and the definition \eqref{Gamma_s} of $\Gamma_{s-1}$. Thus, we bound the second term of \eqref{eqn:1503} as
\begin{align}
\|\sum_{g\neq h} \dfrac{\wh n\s_{gh}}{\nhat_h\s} (\wb\by_{U\s_{gh}} - \bmu_h)\| & \leq \sum_{g\neq h} \dfrac{\wh n_{gh}\s}{\wh n_h\s} \|\wb\by_{U\s_{gh}} - \bmu_h \| \nonumber \\
& \leq \sum_{g\neq h} \dfrac{\wh n\s_{gh}}{\nhat_h\s} \left( \dfrac{C_1}{\sqrt{n\rho_n }\sqrt{\nhat\s_{gh}}}  + \epsilon + 2\Gamma_{s-1}\Delta_{\min} \right) \nonumber\\
& \leq \dfrac{C_1}{\sqrt{n\rho_n}}\sqrt{K\sum_{g\neq h} 
\dfrac{\wh n\s_{gh}}{(\nhat_h\s)^2}} +\sum_{g\neq h} \dfrac{\wh n\s_{gh}}{\nhat_h\s} \epsilon + \sum_{g\neq h} \dfrac{\wh n\s_{gh}}{\nhat_h\s} 2\Gamma_{s-1}\Delta_{\min} \nonumber \\
& \leq C_1 \sqrt{\dfrac{KG_s}{n\rho_n \nhat\s_{h}}} + \sum_{g\neq h} \dfrac{\wh n\s_{gh}}{\nhat_h\s} \epsilon + 2\Gamma_{s-1}\Delta_{\min} G_s,
\end{align}
where the last two inequalities are obtained via Cauchy-Schwarz inequality and the definition \eqref{G_s} of $G_s$. For the first term of \eqref{eqn:1503}, we use the fact that $\{i : z_i = h\} = U\s_{hh} + \bigcup_{g\neq h} U\s_{hg}$ and the definition of $\epsilon$, 
\begin{align}
\bigg\| \dfrac{1}{\nhat \s_h} \sum_{i\in U\s_{hh}} (\by_i -\bmu_h)  \bigg\| &\leq \frac{\nhat \s_{hh}}{\nhat \s_h} \epsilon + \dfrac{1}{\nhat \s_h}\bigg \| \sum_{i: z_i=h} (\by^*_i - \bmu_h) - \sum_{g\neq h} \sum_{i \in U\s_{hg}} (\by^*_i - \bmu_h) \bigg \| \label{decomp_nu_phi_SBM}\\
&\leq \frac{\nhat \s_{hh}}{\nhat \s_h}\epsilon + \dfrac{1}{\nhat \s_h} \left( \dfrac{C_2\sqrt{n_h\log n}}{n\sqrt{\rho_n}} + \dfrac{C_1}{\sqrt{n\rho_n}}\sqrt{n_h - \nhat\s_{hh}} \right), \label{bound_avgUhh_SBM}
\end{align}
where the last inequality follows from applications of Lemma \ref{Lemma_4_SBM} and Lemma \ref{Lemma_1_SBM}.  

Note that, under the assumption that $G_s \leq \frac{1}{2}$, 
\begin{equation}\label{Gs<1/2_conseq SBM}
\nhat\s_h \geq \nhat\s_{hh} \geq n_h(1-G_s) \geq \dfrac{n_h}{2} \geq \dfrac{\alpha n}{2},
\end{equation}
which implies 
\begin{equation}\label{nhat_bound1_SBM}
    \dfrac{1}{\sqrt{\nhat\s_{h}}} \leq \sqrt{\dfrac{2}{n\alpha}}, \quad \dfrac{\sqrt{n_h}}{\nhat\s_h} \leq \dfrac{2}{\sqrt{n_h}} \leq \dfrac{2}{\sqrt{n\alpha}}, \quad \dfrac{\sqrt{n_h - \nhat\s_{hh}}}{\sqrt{n_h}} \leq \sqrt{G_s}. 
\end{equation}
Then, for all $h \in [K]$,
\begin{align*}
\|\wh\bmu \s_h - \bmu_h \| &\leq  C_1 \sqrt{\dfrac{KG_s}{n\rho_n \nhat\s_{h}}} +\sum_{g\neq h} \dfrac{\wh n\s_{gh}}{\nhat_h\s} \epsilon + 2\Gamma_{s-1}\Delta_{\min} G_s \\
& \quad +\frac{\nhat \s_{hh}}{\nhat \s_h} \epsilon + \dfrac{1}{\nhat \s_h} \left( \dfrac{C_2\sqrt{n_h\log n}}{n\sqrt{\rho_n}} + \dfrac{C_1}{\sqrt{n\rho_n}}\sqrt{n_h - \nhat\s_{hh}} \right), \\
& \leq \frac{(2+\sqrt{2})C_1}{n}\sqrt{\dfrac{KG_s}{\alpha \rho_n} } + 2\Gamma_{s-1}G_s\Delta_{\min} + \epsilon + \frac{2C_2}{n}\sqrt{\dfrac{\log n
}{n \alpha\rho_n}} \\
& \leq \Delta_{\min} \left( \frac{\epsilon}{\Delta_{\min}} + 2G_s\Gamma_{s-1} + (1+\sqrt{2})C_1\sqrt{\frac{KG_s}{n\alpha\rho_n}} + \frac{\sqrt{2}C_2}{n}\sqrt{\dfrac{\log n
}{\alpha\rho_n}} \right),
\end{align*}
where we use $\Delta_{\min}\geq \sqrt{\dfrac{2}{\max_s n_s}} \geq \sqrt{\dfrac{2}{n}}$ and the second inequality is obtained using \eqref{nhat_bound1_SBM} and the last inequality is obtained using the definition \eqref{rho_sigma_SBM_K_2} of $\rho_{\sigma}$. Thus,
\begin{equation}\label{eq:Gamma_s_bound1_SBM}
\Gamma_s \leq 
\frac{\epsilon}{\Delta_{\min}} + 2G_s\Gamma_{s-1} + (1+\sqrt{2})\frac{\sqrt{KG_s}}{\rho_{\sigma}} + \frac{\sqrt{2}C_2}{C_1} \frac{1}{\rho_{\sigma}} \sqrt{\dfrac{\log n}{n}}.  
\end{equation}
We can get another bound on $\Gamma_s$ by rewriting $\wh\bmu\s_h$ as 
\begin{align*}
\wh\bmu\s_h &= \dfrac{1}{\nhat\s_h} \sum_{i=1}^n (\bmu_{z_i} + (\by_i - \bmu_{z_i}))\cI(\zhat\s_i = h) \\
&= \sum_{g\in [K]} \dfrac{\wh n\s_{gh}}{\nhat\s_h} \bmu_g + \dfrac{1}{\nhat\s_h} \sum_{\zhat\s_i = h} (\by_i - \bmu_{z_i}).
\end{align*}
Using this decomposition and Lemma \ref{Lemma_1_SBM}, 
\begin{align*}\label{eq:Gamma_s_bound2_SBM}
\|\wh\bmu\s_h - \bmu_h\| &\leq \sum_{g \neq h} \dfrac{\wh n\s_{gh}}{\nhat\s_h}\|\bmu_g - \bmu_h\| + \left\|\dfrac{1}{\nhat\s_h} \sum_{\zhat\s_i=h} ( \by^*_i -\bmu_{z_i}) \right \| + \epsilon \\
&\leq M \sum_{g\neq h} \dfrac{\wh n\s_{gh}}{\nhat\s_h} + C_1\sqrt{\dfrac{1}{n\rho_n \nhat\s_h}} + \epsilon \\
&\leq \Delta_{\min} \left( \dfrac{M}{\Delta_{\min}} G_s + C_1 \sqrt{\dfrac{1}{n \alpha\rho_n }}+ \dfrac{\epsilon}{\Delta} \right).
\end{align*}
This implies
\begin{equation}\label{eq:Gamma_bound_SBM_2}
\Gamma_s \leq \dfrac{\Delta_{\max}}{\Delta_{\min}} G_s + C_1 \sqrt{\dfrac{1}{n \alpha\rho_n }}+ \dfrac{\epsilon}{\Delta_{\min}} \leq \sqrt{\frac{C_1}{c_1}} G_s + C_1 \sqrt{\dfrac{1}{n \alpha\rho_n }}+ \dfrac{\epsilon}{\Delta_{\min}},
\end{equation}
where the last inequality uses \eqref{eqn:M_Delta_bound}. Finally, combining \eqref{eq:Gamma_s_bound1_SBM} and \eqref{eq:Gamma_bound_SBM_2}, we conclude that
\begin{equation*}
\Gamma_s \leq \dfrac{\epsilon}{\Delta_{\min}}
+ \min \left( 2G_s\Gamma_{s-1} + (1+\sqrt{2})\frac{\sqrt{KG_s}}{\rho_{\sigma}} + \frac{\sqrt{2}C_2}{C_1} \frac{1}{\rho_{\sigma}} \sqrt{\dfrac{\log n}{n}}, \frac{1}{\rho_{\sigma}} + \sqrt{\frac{C_1}{c_1}} G_s  \right).
\end{equation*}
\end{proof}

\noindent{\bf Lemma on misclustering recursion.}

\begin{lemma}\label{Lemma_G_s_SBM}
Assume Lemmas \ref{Lemma_1_SBM}, \ref{Lemma_2_SBM}, \ref{Lemma_4_SBM}, and \ref{Lemma_5_SBM} hold. Assume  $\Gamma_s \leq \frac{1-C_{\Gamma}}{2}$ for some $0<C_{\Gamma}<1$, $G_s < \frac{1}{2}$ for all $s$. Define 
$$
\beta_{1,\sigma} = \frac{1}{2}C_{\Gamma} - \frac{3C_0}{2\sqrt{2n\rho_n}} - \frac{C_1}{\alpha^{1/2}(n\rho_n)^{1/4}} - \frac{\sqrt{2}C_0}{\alpha^{1/2}(n\rho_n)^{1/4}},
$$ and assume $\beta_{1,\sigma} > 0$. Then
\begin{align*}
G_{s+1} \leq& \frac{4}{\alpha} \exp\left( -C_3\beta_{1,\sigma}^2n\rho_n \right) + \frac{2}{\alpha}\exp\left( -C_4\beta_{1,\sigma}^2n\rho_n \right) + \frac{32\log n}{n\alpha} \\
&+ \frac{8}{\alpha}\sqrt{\frac{2\log n}{n}}\left\{2\exp\left( -\frac{C_3}{2}\beta_{1,\sigma}^2n\rho_n \right) + \exp\left( -\frac{C_4}{2}\beta_{1,\sigma}^2n\rho_n \right) \right\} + \dfrac{4\Gamma_s^2}{ \sqrt{n\alpha \rho_n}} \\
\leq& \frac{2}{\beta_{1,\sigma}^2n\alpha\rho_n}\left(\frac{2}{C_3} + \frac{1}{C_4} \right)\left(1+16\sqrt{\frac{2\log n}{n}} \right) + \frac{32\log n}{n\alpha} + \dfrac{4\Gamma_s^2}{ \sqrt{n\alpha \rho_n}}. 
\end{align*}
\end{lemma}    

\begin{proof}
To control $G_{s+1}$, we need a bound on $\nhat^{(s+1)}_{gh} = \sum_i \cI(z_i = g,\; \zhat^{(s+1)}_i = h)$ for $h,g \in [K]$ with $h \neq g$. Suppose $g$ is fixed. Then for $h\neq g$,
\begin{align}
& \cI(z_i = g, \; \zhat^{(s+1)}_i = h)\\
\leq & \cI(\|\by_i - \wh\bmu\s_h\|^2 \leq \| \by_i - \wh\bmu\s_g\|^2)\nonumber \\
= & \cI(\langle \by_i  - \bmu_g + \bmu_g- \wh\bmu\s_h,\by_i  - \bmu_g + \bmu_g- \wh\bmu\s_h \rangle \nonumber \\
&\quad \qquad \leq \langle \by_i - \bmu_g + \bmu_g - \wh\bmu\s_g, \by_i - \bmu_g + \bmu_g - \wh\bmu\s_g\rangle)\nonumber \\
= & \cI (\langle \bmu_g - \wh\bmu\s_h,\bmu_g - \wh\bmu\s_h\rangle - \langle \bmu_g - \wh\bmu\s_g,\bmu_g - \wh\bmu\s_g\rangle \leq 2\langle \by_i - \bmu_g, \wh\bmu\s_h - \wh\bmu\s_g\rangle ) \nonumber \\
= & \cI(\|\bmu_g - \wh\bmu\s_h\|^2 - \| \bmu_g - \wh\bmu\s_g\|^2 \leq 2\langle \by_i - \bmu_g, \wh\bmu\s_h - \wh\bmu\s_g\rangle )\label{indicator_rewritten_SBM}
\end{align}
and
\begin{align*}
\|\bmu_g - \wh\bmu\s_h\|^2 &\geq (\|\bmu_g - \bmu_h\| - \|\bmu_h - \wh\bmu\s_h\|)^2 \\
& = \bigg ( \|\bmu_g - \bmu_h\| \bigg ( 1 - \dfrac{\|\bmu_h - \wh\bmu\s_h\|}{\|\bmu_g - \bmu_h\|} \bigg ) \bigg )^2 \\
& \geq \|\bmu_g - \bmu_h\|^2 (1-\Gamma_s)^2, 
\end{align*}
by the definition \eqref{Gamma_s} of $\Gamma_s$. So, 
\begin{align}
\|\bmu_g - \wh\bmu\s_h\|^2 - \| \bmu_g - \wh\bmu\s_g\|^2 & \geq \|\bmu_g - \bmu_h\|^2 (1-\Gamma_s)^2 - \| \bmu_g - \wh\bmu\s_g\|^2 \nonumber\\ 
&\geq  \|\bmu_g - \bmu_h\|^2 ((1-\Gamma_s)^2  - \Gamma_s^2) \nonumber \\
& = \|\bmu_g - \bmu_h\|^2 (1-2\Gamma_s) \label{eqn:1643} \\
&  \geq \|\bmu_g - \bmu_h\|^2 C_{\Gamma}, \nonumber
\end{align}
where the last inequality is obtained using the assumption that $\Gamma_s \leq \frac{1-C_{\Gamma}}{2}$. For $k \in [K]$, write $\mvgz_k = \wh\bmu\s_k - \bmu_k$.
Based on the decomposition 
$$\wh\bmu\s_h - \wh\bmu\s_g = \bmu_h - \bmu_g + (\wh\bmu\s_h - \bmu_h) - (\wh\bmu\s_g - \bmu_g) = (\bmu_h - \bmu_g) + (\bzeta_h - \bzeta_g),$$ 
we define coefficients corresponding to the two parts $\bmu_h - \bmu_g$ and $\bzeta_h - \bzeta_g$ whose reasons will be clear later
\begin{equation}
\label{eqn:def_beta_SBM}
\beta_{1,\epsilon} = \frac{3C_0}{2\sqrt{2n\rho_n}}, \beta_{2,\sigma} = \frac{C_1}{(n\alpha\rho_n)^{1/4}}, \beta_{2,\epsilon} = \frac{\sqrt{2}C_0}{(n\alpha\rho_n)^{1/4}},     
\end{equation}
and 
$$
\beta_{1,\sigma} = \frac{1}{2}C_{\Gamma} - \beta_{1,\epsilon} - \beta_{2,\sigma} - \beta_{2,\epsilon} = \frac{1}{2}C_{\Gamma} - \frac{3C_0}{2\sqrt{2n\rho_n}} - \frac{C_1}{(n\alpha\rho_n)^{1/4}} - \frac{\sqrt{2}C_0}{(n\alpha \rho_n)^{1/4}},
$$
where $C_0$ is define in equation \eqref{eqn:epsilon} and $C_1$ defined in Lemma \ref{Lemma_1_SBM}.
Using the definition above,
\begin{align}
\cI(z_i = g, \; \zhat^{(s+1)}_i=h) 
& \leq \cI\left( \left(\frac{1-2\Gamma_s}{2}\right)\|\bmu_g - \bmu_h\|^2  \leq \langle \by_i - \bmu_g, \wh\bmu\s_h - \wh\bmu\s_g\rangle \right) \nonumber \\
&\leq \cI\left( \beta_{1,\epsilon}\|\bmu_g - \bmu_h\|^2 \leq \langle \by_i - \yb_i^*, \bmu_h - \bmu_g\rangle \right) \nonumber \\
& + \cI\left(\beta_{1,\sigma} \|\bmu_g - \bmu_h\|^2 \leq \langle \by_i^* - \bmu_g,  \bmu_h - \bmu_g\rangle \right) \nonumber \\  
&+ \cI\left( \beta_{2,\epsilon}\|\bmu_g - \bmu_h\|^2 \leq \langle \by_i - \yb_i^*,  \bzeta_h -\bzeta_g \rangle \right) \nonumber \\
& + \cI\left(\beta_{2,\sigma} \|\bmu_g - \bmu_h\|^2 \leq \langle \by_i^* - \bmu_g,  \bzeta_h -\bzeta_g \rangle \right) \nonumber \\    
&=: T_{1i} + T_{2i} + T_{3i} + T_{4i} .\label{eq:indicator_for_incorrect_clustering_SBM}
\end{align}

For the first part of \eqref{eq:indicator_for_incorrect_clustering_SBM}, we have
\begin{align}
T_{1i} = & \cI\left( \beta_{1,\epsilon}\|\bmu_g - \bmu_h\|^2 \leq \langle \by_i - \yb_i^*, \bmu_h - \bmu_g\rangle \right) \\
\leq & \cI\left( \beta_{1,\epsilon}\Delta_{\min} \leq \epsilon \right) \leq \cI\left( \beta_{1,\epsilon}\sqrt{\frac{2}{n}} \leq \epsilon  \right) \\
\leq & \cI\left( \frac{3C_0}{2\sqrt{2n\rho_n}} \sqrt{\frac{2}{n}} \leq \frac{C_0}{n\sqrt{\rho_n}}  \right) = 0, 
\label{T1_ineq_SBM} 
\end{align}
where the second inequality uses the fact that $\Delta_{\min}\geq \sqrt{\dfrac{2}{n}}$. The term related to $T_{2i}$ of \eqref{eq:indicator_for_incorrect_clustering_SBM} can be bounded using Lemma \ref{Lemma_5_SBM} as follows:
\begin{align*}
\sum_{i \in \cC_g} T_{2i} &\leq 2n_g\exp\left( -C_3\beta_{1,\sigma}^2n\rho_n \right) + n_g\exp\left( -C_4\beta_{1,\sigma}^2n\rho_n \right) + 16\log n \\
& + 4\sqrt{n_g\log n}\left\{2\exp\left( -\frac{C_3}{2}\beta_{1,\sigma}^2n\rho_n \right) + \exp\left( -\frac{C_4}{2}\beta_{1,\sigma}^2n\rho_n \right) \right\}\\
& \leq \dfrac{2n_g}{C_3\beta_{1,\sigma}^2n\rho_n} + \dfrac{n_g}{C_4\beta_{1,\sigma}^2n\rho_n} + 16\log n + \dfrac{16\sqrt{n_g\log n}}{C_3\beta_{1,\sigma}^2n\rho_n} + \dfrac{8\sqrt{n_g\log n}}{C_4\beta_{1,\sigma}^2n\rho_n},
\end{align*}
where $C_3$ and $C_4$ are constants defined in Lemma \ref{Lemma_5_SBM} and we use the fact $\exp(-x)\leq \frac{1}{x}$ for $x>0$.  

Based on the following result
\begin{equation}
\| \bzeta_h -\bzeta_g\| \leq  \|\bzeta_h\| + \|\bzeta_g\|  = \|\wh\bmu\s_h- \bmu_h \| + \|\wh\bmu\s_g - \bmu_g \| \leq 2\Gamma_s\Delta_{\min}\leq 2\Gamma_s \|\bmu_h-\bmu_g\|, \label{eqn:1045}
\end{equation}  
we can bound the third term of \eqref{eq:indicator_for_incorrect_clustering_SBM} as 
\begin{align*}
\sum_{i:z_i = g} T_{3i} & = \sum_{i:z_i = g}\cI\left( \beta_{2,\epsilon} \|\bmu_g - \bmu_h\|^2 \leq \langle \yb_i - \yb_i^*, \bzeta_h -\bzeta_g \rangle \right)  \nonumber \\
& = \sum_{i:z_i = g}\cI\left(1 \leq \dfrac{1}{\beta_{2,\epsilon}^2 \|\bmu_g - \bmu_h\|^4}\langle \yb_i - \yb_i^*, \bzeta_h -\bzeta_g \rangle^2 \right)  \nonumber \\
&\leq \sum_{i:z_i = g}  \dfrac{1}{\beta_{2,\epsilon}^2\Delta_{\min}^4}\left(\langle \by_i - \by_i^*, \mvgz_h - \mvgz_g \rangle\right)^2 \nonumber \\ 
&\leq \dfrac{\epsilon^2 \|\bzeta_h -\bzeta_g\|^2}{\beta_{2,\epsilon}^2\Delta_{\min}^4} n_g  \leq \dfrac{2 C_0^2 \Gamma_s^2}{\beta_{2,\epsilon}^2n\rho_n}n_g 
 = \sqrt{\dfrac{\alpha}{n\rho_n}}n_g \Gamma_s^2
\end{align*}
using the definition of $\epsilon$ in \eqref{eqn:epsilon}, the inequality \eqref{eqn:1045} and $\Delta_{\min}\geq \sqrt{\dfrac{2}{n}}$.

Similarly, we can bound the fourth term of \eqref{eq:indicator_for_incorrect_clustering_SBM} as 
\begin{align}
\sum_{i:z_i = g} T_{4i} & = \sum_{i:z_i = g}\cI\left( \beta_{2,\sigma}\|\bmu_g - \bmu_h\|^2 \leq \langle \by_i^* - \bmu_g, \bzeta_h -\bzeta_g \rangle \right)  \nonumber \\ 
&\leq \dfrac{1}{\beta_{2,\sigma}^2\Delta_{\min}^4 } \sum_{i:z_i = g} ( \langle \by^*_i - \bmu_g,\mvgz_h - \mvgz_g\rangle ) ^2 \nonumber \\
&\leq \dfrac{\|\bzeta_h -\bzeta_g\|^2}{\beta_{2,\sigma}^2\Delta_{\min}^4} 
\lambda_{\max}\left( \sum_{i:z_i = g}\bw_i\bw_i^{\top}\right)\label{T4_ineq_1_SBM}
\\
&\leq \dfrac{4\Gamma_s^2}{\beta_{2,\sigma}^2 \Delta_{\min}^2} \frac{C_1^2}{n\rho_n}
\leq \dfrac{C_1^2 \Gamma_s^2}{ \beta_{2,\sigma}^2 \rho_n} = \sqrt{\frac{n\alpha}{\rho_n}} \Gamma_s^2, 
\label{T4_ineq_2_SBM} 
\end{align} 
where \eqref{T4_ineq_1_SBM} uses Lemma \ref{Lemma_2_SBM}, and \eqref{T4_ineq_2_SBM} uses the definition of $\beta_{2,\sigma}$ in \eqref{eqn:def_beta_SBM}.  Note that \eqref{Gs<1/2_conseq SBM}, which uses the assumption that $G_s \leq \frac{1}{2}$, implies
$$\dfrac{1}{n_g} \leq \dfrac{1}{\alpha n}, \quad \dfrac{n_g}{\wh n_h^{(s+1)}}  \leq \dfrac{2n_g}{\alpha n} \leq \dfrac{2}{\alpha}, \quad \dfrac{\sqrt{n_g}}{\wh n_h^{(s+1)}} \leq \dfrac{2\sqrt{n_g}}{\alpha n} \leq \dfrac{2}{\alpha \sqrt{n}}. $$

Combining the four parts together, we have
\begin{align*}
&\max_{g\in[K]}\sum_{h\neq g} \dfrac{\nhat^{(s+1)}_{gh}}{n_g} 
= \max_{g\in[K]} \sum_{h\neq g} \sum_{i\in[n]} \dfrac{T_{1i} + T_{2i} + T_{3i} +  T_{4i}}{n_g} \\
\leq & 2 \exp\left( -C_3\beta_{1,\sigma}^2n\rho_n \right) + \exp\left( -C_4\beta_{1,\sigma}^2n\rho_n \right) + \frac{16\log n}{n_g} + 4\sqrt{\frac{\log n}{n_g}}   \\
& \left\{2\exp\left( -\frac{C_3}{2}\beta_{1,\sigma}^2n\rho_n \right) + \exp\left( -\frac{C_4}{2}\beta_{1,\sigma}^2n\rho_n \right) \right\} + \sqrt{\dfrac{\alpha}{n\rho_n}} \Gamma_s^2 + \sqrt{\frac{n\alpha}{\rho_n}}\frac{\Gamma_s^2}{n_g} \\
\leq & 2 \exp\left( -C_3\beta_{1,\sigma}^2n\rho_n \right) + \exp\left( -C_4\beta_{1,\sigma}^2n\rho_n \right) + \frac{16\log n}{n\alpha} + 4\sqrt{\frac{\log n}{n\alpha}} \\
& \left\{2\exp\left( -\frac{C_3}{2}\beta_{1,\sigma}^2n\rho_n \right) + \exp\left( -\frac{C_4}{2}\beta_{1,\sigma}^2n\rho_n \right) \right\} + \sqrt{\dfrac{\alpha}{n\rho_n}} \Gamma_s^2 + \frac{\Gamma_s^2 }{\sqrt{n\alpha\rho_n}}
\end{align*} 
and 
\begin{align*}
&\max_{h\in[K]}\sum_{g\neq h} \dfrac{\nhat^{(s+1)}_{gh}}{\wh n^{(s+1)}_h}
= \max_{h\in[K]}\sum_{g\neq h} \sum_{i\in[n]} \dfrac{T_{1i} + T_{2i} + T_{3i} + T_{4i}}{\widehat{n}_h^{(s+1)}} \\
\leq& \frac{2n_g}{\widehat{n}_h^{(s+1)}} \exp\left( -C_3\beta_{1,\sigma}^2n\rho_n \right) + \frac{n_g}{\widehat{n}_h^{(s+1)}}\exp\left( -C_4\beta_{1,\sigma}^2n\rho_n \right) + \frac{16\log n}{\widehat{n}_h^{(s+1)}} \\
&+ 4\frac{\sqrt{n_g\log n}}{\widehat{n}_h^{(s+1)}}\left\{2\exp\left( -\frac{C_3}{2}\beta_{1,\sigma}^2n\rho_n \right) + \exp\left( -\frac{C_4}{2}\beta_{1,\sigma}^2n\rho_n \right) \right\} \\
& +  \sqrt{\dfrac{\alpha}{n\rho_n}} \Gamma_s^2 \frac{n_g}{\widehat{n}_h^{(s+1)}} + \sqrt{\frac{n\alpha}{\rho_n}}\frac{ \Gamma_s^2}{\widehat{n}_h^{(s+1)}} \\
\leq& \frac{4}{\alpha} \exp\left( -C_3\beta_{1,\sigma}^2n\rho_n \right) + \frac{2}{\alpha}\exp\left( -C_4\beta_{1,\sigma}^2n\rho_n \right) + \frac{32\log n}{n\alpha} \\
&+ \frac{8}{\alpha}\sqrt{\frac{2\log n}{n}}\left\{2\exp\left( -\frac{C_3}{2}\beta_{1,\sigma}^2n\rho_n \right) + \exp\left( -\frac{C_4}{2}\beta_{1,\sigma}^2n\rho_n \right) \right\} + \dfrac{4\Gamma_s^2}{ \sqrt{n\alpha \rho_n}} . 
\end{align*} 

Finally, we have 
\begin{align*}
G_{s+1} \leq& \frac{4}{\alpha} \exp\left( -C_3\beta_{1,\sigma}^2n\rho_n \right) + \frac{2}{\alpha}\exp\left( -C_4\beta_{1,\sigma}^2n\rho_n \right) + \frac{32\log n}{n\alpha} \\
+& \frac{8}{\alpha}\sqrt{\frac{2\log n}{n}}\left\{2\exp\left( -\frac{C_3}{2}\beta_{1,\sigma}^2n\rho_n \right) + \exp\left( -\frac{C_4}{2}\beta_{1,\sigma}^2n\rho_n \right) \right\} + \dfrac{4\Gamma_s^2}{ \sqrt{n\alpha \rho_n}} \\
\leq& \frac{2}{\beta_{1,\sigma}^2n\alpha\rho_n}\left(\frac{2}{C_3} + \frac{1}{C_4} \right)\left(1+16\sqrt{\frac{2\log n}{n}} \right) + \frac{32\log n}{n\alpha} + \dfrac{4\Gamma_s^2}{ \sqrt{n\alpha \rho_n}}. 
\end{align*}
\end{proof}

\subsubsection*{Main proof of Theorem~\ref{thm:sbm_K}}\label{main_proof_sbm}

\step{1: Initialization and uniform control of $(G_s,\Gamma_s)$}
In order to bound the misclustering rate $A_{s+1}$, we first show that under the initialization condition \eqref{initial_conditions1} or \eqref{initial_conditions2}, Lemmas \eqref{lem:Gamma-step}--\eqref{lem:G-step} hold for all $s\geq 1$. Next, we find a bound for $\Gamma_s$ which does not depend on $s$. Then we decompose $\cI(z_i \neq \zhat_i^{(s+1)}, \zhat_i^{(s+1)} = h)$
using \eqref{eq:nu_and_phi_defs} and the bound on $\Gamma_s$, which allows us to decompose $A_{s+1}$ into three components that can be bounded in expectation.

Note that if $G_0$ satisfies the initial condition \eqref{initializer_SBM}, it follows from Lemma \ref{Lemma_G_s_SBM} that 
\begin{align}
    \Gamma_0 & \leq \frac{\epsilon}{\Delta_{\min}} + \frac{1}{\rho_{\sigma}} + \sqrt{\frac{C_1}{c_1}} G_0 \nonumber \\
    &\leq \frac{\epsilon}{\Delta_{\min}} + \frac{1}{\rho_{\sigma}} + \left( \frac{1}{2} - \frac{5C_0}{2\sqrt{2n\rho_n}} - \frac{C_1}{\sqrt{n\alpha\rho_n}} - \frac{\sqrt{2}C_0+C_1+1}{(n\alpha\rho_n)^{1/4}} \right) \nonumber \\
    &\leq \frac{1}{2} - \frac{C_0+C_1}{(n\rho_n)^{1/4}}\\
    &= \frac{1}{2} - \frac{3C_0}{2\sqrt{2n\rho_n}} - \frac{\sqrt{2}C_0+C_1+1}{(n\alpha\rho_n)^{1/4}} \label{Gamma0_SBM},
\end{align}
where the last inequality follows from the fact that $n\rho_n\gg 1$. Therefore, regardless of which initial condition \eqref{initializer_SBM} holds, we have \eqref{Gamma0_SBM}. 
Plugging \eqref{Gamma0_SBM} into Lemma \ref{Lemma_G_s_SBM} with 
$$C_{\Gamma} := \frac{3C_0}{\sqrt{2n\rho_n}} + \frac{2(\sqrt{2}C_0+C_1+1)}{(n\alpha\rho_{n})^{1/4}} ,$$ 
yields 
\begin{align*}
\beta := \frac{1}{2}C_{\Gamma} - \frac{3C_0}{2\sqrt{2n\rho_n}} - \frac{C_1}{(n\alpha\rho_n)^{1/4}} - \frac{\sqrt{2}C_0}{(n\alpha\rho_n)^{1/4}} 
= \frac{1}{(n\alpha\rho_{n})^{1/4}}
\end{align*}
and 
\begin{align*}
    G_1 & \leq \frac{2}{\beta_{1,\sigma}^2n\alpha\rho_n}\left(\frac{2}{C_3} + \frac{1}{C_4} \right)\left(1+16\sqrt{\frac{2\log n}{n}} \right) + \frac{32\log n}{n\alpha} + \dfrac{4\Gamma_0^2}{ \sqrt{n\alpha\rho_n}} \\
    &\leq  \frac{2}{(n\alpha\rho_n)^{1/2}}\left(\frac{2}{C_3} + \frac{1}{C_4} \right)\left(1+16\sqrt{\frac{2\log n}{n}} \right) + \frac{32\log n}{n\alpha} + \dfrac{4\Gamma_0^2}{ \sqrt{n\alpha \rho_n}} \\
    &\leq 0.35,
\end{align*}
using the fact that $n\alpha\rho_n\gg 1$ and $\dfrac{n\alpha}{\log n} \geq 256$. Then Lemma \ref{Lemma_Gamma_s_SBM} yields 
\begin{align}
    \Gamma_1 & \leq \frac{\epsilon}{\Delta_{\min}} + 2G_1\Gamma_{0} + (1+\sqrt{2})\frac{\sqrt{KG_1}}{\rho_{\sigma}} + \frac{\sqrt{2}C_2}{C_1} \frac{1}{\rho_{\sigma}} \sqrt{\dfrac{\log n}{n}}  \nonumber\\ 
    & \leq  \frac{C_1}{\sqrt{2n\rho_n}}
    + 2(0.35)\left( \frac{1}{2} - \frac{3C_0}{2\sqrt{2n\rho_n}} - \frac{\sqrt{2}C_0+C_1+1}{(n\alpha\rho_n)^{1/4}} \right) \\
    & + (1+\sqrt{2})\frac{\sqrt{0.35}}{\rho_{\sigma}} + \frac{\sqrt{2}C_2}{C_1} \frac{1}{\rho_{\sigma}} \sqrt{\dfrac{\log n}{n}} \nonumber\\
    &\leq  \frac{1}{2} - \frac{3C_0}{2\sqrt{2n\rho_n}} - \frac{\sqrt{2}C_0+C_1+1}{(n\alpha\rho_n)^{1/4}},
\end{align}
under the assumption that $n\alpha\rho_n \gg 1$. 
By induction, we can show 
\begin{equation}\label{Gs_Gammas_bdd_all_s_SBM}
G_s < 0.35,\Gamma_s < 0.4
\end{equation}
for all $s> 1$. Then, $\Gamma_s \leq 0.4 = \frac{1- (1/5)}{2}$, with $C_{\Gamma} = \frac{1}{5}$. 
Lemma \ref{Lemma_G_s_SBM} gives
\begin{align*} 
G_{s+1} & \leq \frac{4}{\alpha} \exp\left( -C_3\beta^2n\rho_n \right) + \frac{2}{\alpha}\exp\left( -C_4\beta^2n\rho_n \right) + \frac{32\log n}{n\alpha} \\
&+ \frac{8}{\alpha}\sqrt{\frac{2\log n}{n}}\left\{2\exp\left( -\frac{C_3}{2}\beta^2n\rho_n \right) + \exp\left( -\frac{C_4}{2}\beta^2n\rho_n \right) \right\} + \dfrac{4\Gamma_s^2}{ \sqrt{n\alpha \rho_n}},
\end{align*}
where $\beta = \frac{1}{2}C_{\Gamma} - o(1) = 0.1 - o(1) $. Hence, using Lemmas \ref{Lemma_G_s_SBM} and \ref{Lemma_Gamma_s_SBM} along with $n\alpha\rho_n \gg 1$, and $C_{\Gamma} = \frac{1}{5}$, for all $s >1$,
\begin{align*}
\Gamma_s & \leq \frac{\epsilon}{\Delta_{\min}} + 2G_s\Gamma_{s-1} 
+ (1+\sqrt{2})\frac{\sqrt{KG_s}}{\rho_{\sigma}} 
+ \frac{\sqrt{2}C_2}{C_1} \frac{1}{\rho_{\sigma}} \sqrt{\frac{\log n}{n}} \\ 
& \leq \frac{\epsilon}{\Delta_{\min}} + 2G_s\Gamma_{s-1} 
+ \frac{1+\sqrt{2}}{\rho_{\sigma}} \sqrt{KG_s} 
+ \frac{\sqrt{2}C_2}{C_1} \frac{1}{\rho_{\sigma}} \sqrt{\frac{\log n}{n}} \\
& \leq \frac{\epsilon}{\Delta_{\min}} + 0.7\Gamma_{s-1} 
+ \frac{(1+\sqrt{2})\sqrt{K}}{\rho_{\sigma}} 
\Bigg\{\frac{2}{\sqrt{\alpha}} \exp\left( -\frac{C_3}{2}\beta^2n\rho_n \right) 
+ 4\sqrt{\frac{2\log n}{n\alpha}}\\
& + \frac{\sqrt{2}}{\sqrt{\alpha}} \exp\left( -\frac{C_4}{2}\beta^2n\rho_n \right) 
+ \frac{2\sqrt{2}}{\sqrt{\alpha}}\left(\frac{2\log n}{n}\right)^{1/4}
\Bigg[\sqrt{2}\exp\left( -\frac{C_3}{4}\beta^2n\rho_n \right)  \\
& \quad + \exp\left( -\frac{C_4}{4}\beta^2n\rho_n \right)\Bigg] + \frac{2\Gamma_s}{(n\alpha\rho_n)^{1/4}} \Bigg\} 
+ \frac{\sqrt{2}C_2}{C_1} \frac{1}{\rho_{\sigma}} \sqrt{\frac{\log n}{n}} \\
& \leq \left(\frac{c_1}{(n\alpha\rho_n)^{1/4}} + 0.7  \right)\Gamma_{s-1} 
+ \frac{c_1}{\sqrt{n\alpha\rho_n}} + 4\sqrt{\frac{2\log n}{n\alpha}}.
\end{align*}
for some constant $c_1$, using the assumption $n\rho_n \gg 1$ for all $s>1$. Therefore, when $n\rho_n$ is large enough, we have
$$\Gamma_s \leq \frac{c_2}{2\sqrt{n\alpha\rho_n}}  + \frac{c_2}{2}\sqrt{\frac{\log n}{n\alpha}}, $$
for some constant $c_2$ and all $s \geq 2\log n$.
Define       
\begin{equation}\label{eq:beta_1_SBM_thm}
\beta_1 = 1 - \frac{c_2}{\sqrt{n\alpha\rho_n}} - c_2\sqrt{\frac{\log n}{n\alpha}},
\end{equation}
which ensures $\beta_1 \leq 1-2\Gamma_s $ for all $s\geq 2\log n$. Furthermore, define
\begin{align*}
& \beta_{1} = \beta_{1,\sigma} + \beta_{1,\epsilon}  \\
& \beta_{1,\sigma} = \beta_{\sigma} + \beta_{2,\sigma} + \beta_{3,\sigma} + \beta_{4,\sigma}  \\
& \beta_{1,\epsilon} = \beta_{\epsilon} + \beta_{2,\epsilon},
\end{align*}
and     
\begin{align*}
    & \beta_{2,\sigma} = \frac{8\sqrt{C_1}}{(n\alpha\rho_n)^{1/4}} \\ 
    & \beta_{3,\sigma} = 4C_2C_6\dfrac{\log n}{n^{3/2}\alpha\rho_n} \max\left\{ \dfrac{\log n}{\sqrt{n\alpha\rho_n}}, 1\right\}\\
    & \beta_{4,\sigma} = \dfrac{3C_0}{\sqrt{n\rho_n}} \\
    & \beta_{1, \epsilon} = \dfrac{2C_0+1}{\sqrt{n\alpha\rho_n}},\beta_{\epsilon} = \dfrac{2C_0}{\sqrt{n\rho_n}}, \beta_{2,\epsilon} = \dfrac{C_0}{\sqrt{n\alpha\rho_n}}, 
\end{align*}
where the constants are defined in Lemmas \ref{Lemma_1_SBM} - \ref{Lemma_5_SBM} and \eqref{eqn:epsilon}. 
These imply that
$$
\beta_{\sigma}  = \beta_1 - \beta_{1,\epsilon} - \beta_{2,\sigma} - \beta_{3,\sigma} - \beta_{4,\sigma} = 1 - \frac{c_2}{\sqrt{n\rho_n}} -\beta_{1,\epsilon} - \beta_{2,\sigma} - \beta_{3,\sigma} - \beta_{4,\sigma} ,
$$
which implies that $\beta_{\sigma} = 1-c$ for some sufficiently small constant $c$. We assume that $C_1$ and $C_3$ are large enough so that $\beta >0$. 
We will bound $A_{s+1}$ by Markov's inequality, for which we need an upper bound on $\E(A_{s+1})$.

\step{2: Decomposition of the misclassification indicator}
Combining \eqref{indicator_rewritten_SBM}, \eqref{eqn:1643}, $\beta_1 \leq 1-2\Gamma_s$, and the definitions above, we obtain
\begin{align*}
\cI(z_i \neq \zhat_i^{(s+1)}, \zhat_i^{(s+1)} = h) 
&\leq \cI( \beta_1 \|\bmu_{z_i} - \bmu_h\|^2 < 2 \langle \by_i-\bmu_{z_i},\wh\bmu\s_h - \wh\bmu\s_{z_i}\rangle )\\
&\leq \cI( \beta_{1,\sigma} \|\bmu_{z_i} - \bmu_h\|^2 < 2 \langle \by_i^*-\bmu_{z_i},\wh\bmu\s_h - \wh\bmu\s_{z_i}\rangle )\\
&+ \cI( \beta_{1,\epsilon}  \|\bmu_{z_i} - \bmu_h\|^2 \leq 2 \langle \by_i-\by_i^*,\wh\bmu\s_h - \wh\bmu\s_{z_i}\rangle )\\
&= \cI(\frac{\beta_{1,\sigma} }{2} \|\bmu_{z_i}- \bmu_h\|^2 < \langle \bw_i,\bmu_h-\bmu_{z_i} +  \bzeta_h -\bzeta_{z_i}\rangle )\\
&+ \cI(\frac{\beta_{1,\epsilon}}{2} \|\bmu_{z_i}- \bmu_h\|^2 \leq \langle \be_i,\bmu_h-\bmu_{z_i} +  \bzeta_h -\bzeta_{z_i} \rangle )
\end{align*}
where $\bzeta_h := \wh\bmu\s_h - \bmu_h$. 
Define 
\begin{equation}\label{eq:nu_and_phi_defs_SBM}
\bphi_{h} = \dfrac{1}{\nhat\s_h}\sum_{i \in \cC_h^{(s)}} (\by_i - \by^*_i ), \mvnu_h = \dfrac{1}{\nhat\s_h}\sum_{i \in \cC_h} (\by^*_i - \bmu_h),  \mvvarphi_h = (\wh\bmu\s_h - \bmu_h) -  \bphi_{h} - \mvnu_h.
\end{equation}
We decompose $\bzeta_h$ as
\[
\bzeta_h = \bphi_h + \mvnu_h + \mvvarphi_h.
\]
Using the definition of $\epsilon$, we have 
\begin{align}
\| \bphi_{h}\| = \left\|\dfrac{1}{\nhat\s_h}\sum_{i \in C_h\s} (\by_i - \by^*_i ) \right\| \leq \epsilon \label{eq:phi_bound_SBM}.
\end{align}
By Lemma \ref{Lemma_4_SBM}, 
\begin{align}
\|\mvnu_h\| & = 
\left\| \dfrac{1}{\nhat\s_h}\sum_{i \in \cC_h} (\by^*_i - \bmu_h) \right\| \leq \frac{2C_2}{n\alpha}
\frac{\sqrt{n_h\log n}}{n\sqrt{\rho_n}}
\leq \frac{2C_2}{n^2\alpha}\sqrt{\dfrac{n_h\log n}{\rho_n}}\label{nu_bound_SBM}.
\end{align}
From the proof of Lemma \ref{Lemma_Gamma_s_SBM}, 
\begin{align}
& \|\mvvarphi_h\| \\
=& \left\| (\wh\bmu\s_h - \bmu_h) -  \bphi_{h} - \mvnu_h \right\| \nonumber\\
\leq& \left\| \sum_{g\neq h} \dfrac{\wh n\s_{gh}}{\nhat_h\s} (\wb \by_{U\s_{gh}} - \bmu_h) - \frac{1}{\nhat\s_h} \sum_{g\neq h}\sum_{i\in U\s_{gh}}(\by_i-\by_i^*) - \frac{1}{\nhat\s_h} \sum_{g\neq h}\sum_{i\in U\s_{hg}}(\by_i^*-\bmu_h) \right\| \nonumber \\
\leq &
\left\| \sum_{g\neq h} \dfrac{\wh n\s_{gh}}{\nhat_h\s} (\wb \by_{U\s_{gh}} - \bmu_h) \right\| + \left\|\frac{1}{\nhat\s_h} \sum_{g\neq h}\sum_{i\in U\s_{gh}}(\by_i-\by_i^*) \right\| \\
& + \left\| \frac{1}{\nhat\s_h} \sum_{g\neq h}\sum_{i\in U\s_{hg}}(\by_i^*-\bmu_h) \right\| \nonumber \\
\leq & 2\epsilon G_s + 2\Gamma_{s-1}G_s\Delta_{\min} + \frac{C_1(\sqrt{2K}+2)}{n}\sqrt{\frac{G_s}{\alpha\rho_n}}.
\label{varphi_bound_SBM}
\end{align}

For the first part,
\begin{align*}
& \cI(\frac{\beta_{1,\sigma} }{2} \|\bmu_{z_i}- \bmu_h\|^2 < \langle \bw_i,\bmu_h-\bmu_{z_i} + \bzeta_h -\bzeta_{z_i} \rangle )\\
\leq & \cI(\frac{\beta_{\sigma} }{2} \|\bmu_{z_i} - \bmu_h\|^2 < \langle \bw_i,\bmu_h-\bmu_{z_i}\rangle ) 
+ \cI(\frac{\beta_{2,\sigma} }{2} \|\bmu_{z_i} - \bmu_h\|^2 \leq \langle  \bw_i, \mvvarphi_h - \mvvarphi_{z_i} \rangle) \\
+& \cI(\frac{\beta_{3,\sigma} }{2} \|\bmu_{z_i} - \bmu_h\|^2 \leq \langle \bw_i, \mvnu_{h} - \mvnu_{z_i} \rangle)
+ \cI(\frac{\beta_{4,\sigma} }{2} \|\bmu_{z_i} - \bmu_h\|^2 < \langle \bw_i, \bphi_h-\bphi_{z_i} \rangle ). 
\end{align*}
Define
\begin{align*}
J_{1,\sigma} & =  \sum_{h\in [K]} \dfrac{1}{n} \sum_{i = 1}^n  \cI \left(\frac{\beta_{\sigma} }{2} \|\bmu_{z_i} - \bmu_h\|^2 <  \langle \bw_i,\bmu_h-\bmu_{z_i}\rangle \right), \\
J_{2,\sigma} & = \sum_{h\in [K]} \dfrac{1}{n} \sum_{i = 1}^n \cI\left(\frac{\beta_{2,\sigma} }{2} \|\bmu_{z_i} - \bmu_h\|^2 \leq \langle \bw_i, \mvvarphi_h - \mvvarphi_{z_i} \rangle\right), \\
J_{3,\sigma} & = \sum_{h\in [K]} \dfrac{1}{n} \sum_{i = 1}^n  \cI\left(\frac{\beta_{3,\sigma} }{2} \|\bmu_{z_i} - \bmu_h\|^2 \leq \langle \bw_i, \mvnu_h - \mvnu_{z_i}  \rangle\right), \\
J_{4,\sigma} & = \sum_{h\in [K]} \dfrac{1}{n} \sum_{i = 1}^n \cI\left(\frac{\beta_{4,\sigma} }{2} \|\bmu_{z_i} - \bmu_h\|^2 \leq \langle \bw_i, \bphi_h - \bphi_{z_i} \rangle\right).
\end{align*}
On the other hand, we can define $J_{1,\epsilon}, J_{2,\epsilon}$ as 
\begin{align*}
J_{1,\epsilon} & =  \sum_{h\in [K]} \dfrac{1}{n} \sum_{i = 1}^n  \cI \left(\frac{\beta_{\epsilon} }{2} \|\bmu_{z_i} - \bmu_h\|^2 \leq  \langle \be_i,\bmu_h-\bmu_{z_i}\rangle \right), \\
J_{2,\epsilon} & = \sum_{h\in [K]} \dfrac{1}{n} \sum_{i = 1}^n \cI\left(\frac{\beta_{2,\epsilon} }{2} \|\bmu_{z_i} - \bmu_h\|^2 \leq \langle \be_i, \bzeta_h -\bzeta_{z_i} \rangle\right),
\end{align*}
and get the bound
\begin{equation}\label{As_decomp_SBM}
\E(A_{s+1}) \leq \E((J_{1,\sigma}+ J_{2,\sigma} + J_{3,\sigma} +  J_{4,\sigma} 
+ J_{1,\epsilon}+ J_{2,\epsilon} )\cI(\cG)) + \bbP (\cG^c),
\end{equation}
where $\cG$ is the event in which the results of Lemmas \ref{Lemma_1_SBM}, \ref{Lemma_2_SBM}, \ref{Lemma_4_SBM}, and \ref{Lemma_5_SBM} hold.

\step{3: Bounding $J_{1,\sigma}$ (main tail term)}

To bound $\E (J_{1,\sigma})$ in \eqref{As_decomp_SBM}, we use Lemma \ref{Lemma_SBM} to get 
\begin{align*}
\E(J_{1,\sigma}) &= \dfrac{1}{n} \sum_{h \in [k]}\sum_{i=1}^n \bbP\left(\frac{\beta_{\sigma} }{2}\| \bmu_{z_i} - \bmu_h\|^2 \leq \langle \by_i^* - \bmu_{z_i}, \bmu_h - \bmu_{z_i} \rangle \right)   \\
&\leq\dfrac{1}{n} \sum_{h \in [k]}\sum_{i=1}^n \exp\left( -\frac{C_5\beta_{\sigma}^2}{4}n\rho_n\right)\\
&\leq K \exp\left( -\frac{C_5\beta_{\sigma}^2}{4}n\rho_n\right)
\leq \exp\left( -\frac{C_5\beta_{\sigma}^2}{8}n\rho_n\right),
\end{align*}
using $n\rho_n\gg1$ and $K$ fixed. 
Note that by \eqref{varphi_bound_SBM}, for all $h \in [k]$,
\begin{align*}
\|\mvvarphi_h\|  &\leq 2\epsilon G_s + 2\Gamma_{s-1}G_s\Delta_{\min} + \frac{C_1(\sqrt{K}+\sqrt{2})}{n}\sqrt{\frac{2G_s}{\alpha\rho_n}}\\
& \leq 2\epsilon G_s +  G_s \Delta_{\min} + C_1(\sqrt{K}+\sqrt{2})\sqrt{\dfrac{G_s}{n\alpha\rho_n}} \Delta_{\min}  \\
& \leq 2(\Delta_{\min} + \epsilon )\sqrt{G_s}
\end{align*}
when $n\alpha\rho_n \gg 1$ and $G_s \leq 0.35$ as in \eqref{Gs_Gammas_bdd_all_s_SBM}. 
Then using $G_s \leq 0.35$ again,
\begin{align*}
\|\mvvarphi_h - \mvvarphi_{z_i}\| \leq 4(\Delta_{\min} + \epsilon)\sqrt{G_s},
\end{align*}

\step{4: Bounding $J_{2,\sigma}$}

Conditionally on $\cG$, we have
\begin{align}
J_{2,\sigma} 
& = \sum_{h\in [K]} \dfrac{1}{n} \sum_{i = 1}^n \cI\left(\frac{\beta_{2,\sigma} }{2} \|\bmu_{z_i} - \bmu_h\|^2 \leq \langle \bw_i, \mvvarphi_h - \mvvarphi_{z_i} \rangle\right) \nonumber\\         
&\leq \sum_{h\in [K]} \dfrac{4}{n\beta_{2,\sigma}^2 \Delta_{\min}^4} \sum_{i=1}^n \langle \by_i^* - \bmu_{z_i}, \mvvarphi_h - \mvvarphi_{z_i}\rangle ^2 \nonumber\\
& \leq \dfrac{4}{n\beta_{2,\sigma}^2 \Delta_{\min}^4} \bigg (  \frac{C_1^2}{n\rho_n} \|\mvvarphi_h - \mvvarphi_{z_i}\|^2 \bigg ) \label{J2_lemma2_SBM}\\ 
&\leq \dfrac{128C_1^2 G_s ( \Delta_{\min}^2 + \epsilon^2 ) }{n^2\rho_n\beta_{2,\sigma}^2 \Delta_{\min}^4}    \leq \dfrac{128C_1^2 A_s}{n^2\alpha\rho_n\beta_{2,\sigma}^2 \Delta_{\min}^2} + \dfrac{128C_1^2 A_s \epsilon^2}{n^2\alpha\rho_n\beta_{2,\sigma}^2 \Delta_{\min}^4}  \label{J2_beta2_rho_SBM}  \\
& \leq \dfrac{64C_1^2 A_s}{n\alpha\rho_n\beta_{2,\sigma}^2} + \dfrac{32C_1^2 A_s }{n^2\alpha\rho_n^2\beta_{2,\sigma}^2} \leq \left( \dfrac{1}{\rho_{\sigma}} + \dfrac{1}{\rho_{\sigma} \rho_{\epsilon}^2} \right)A_s\nonumber.
\end{align}
where \eqref{J2_lemma2_SBM} follows from Lemma \ref{Lemma_2_SBM}, \eqref{J2_beta2_rho_SBM} follows from the fact that $G_s  \leq  \frac{1}{\alpha}A_s$, and $\beta_{2,\sigma}^2 = \dfrac{64C_1}{\sqrt{n\alpha\rho_n}}$ and the definitions of $\rho_{\sigma}, \rho_{\epsilon}$. Hence, 
\begin{equation}
\E (J_{2,\sigma} \cI(\cG)) \leq  \left( \dfrac{1}{ \rho_{\sigma}} + \dfrac{1}{\rho_{\sigma} \rho_{\epsilon}^2} \right)\E (A_s). 
\end{equation}

\step{5: Bounding $J_{3,\sigma}$}
For the third term, we bound the probability 
\begin{align}
&\bbP\left( \dfrac{\beta_{3,\sigma}\Delta_{\min}^2}{2} 
\leq \langle \by_i^* - \bmu_{z_i}, \mvnu_h-\mvnu_{z_i}  \rangle \right) \nonumber \\
\leq & \bbP\left(\dfrac{\beta_{3,\sigma}\Delta_{\min}^2}{2} \leq \|\bw_i\|  \|\mvnu_h-\mvnu_{z_i} \| \right)
\leq \bbP\left(\dfrac{\beta_{3,\sigma}\Delta_{\min}^2}{2} \leq \|\bw_i\| \frac{4C_2}{n^2\alpha}\sqrt{\dfrac{n_h\log n}{\rho_n}} \right) \nonumber \\
\leq & \bbP\left(\dfrac{\beta_{3,\sigma}\Delta_{\min}^2}{2} \leq \|\bw_i\| \frac{4C_2}{n\alpha}\sqrt{\dfrac{\log n}{n\rho_n}} \right)
\end{align}
using \eqref{nu_bound_SBM}. From Lemma \ref{Lemma_norm_SBM}, 
$$\|\bw_i\| \leq C_6 \max\left\{ \frac{\log n}{n\sqrt{n\alpha}\rho_n}, \frac{1}{n}\sqrt{\frac{\log n}{\rho_n}}\right\} $$
for some constant $C_6$ with probability $1-n^{-3}$, 
and 
$$
\|\mvnu_h-\mvnu_{z_i} \| \leq \|\mvnu_h\| + \| \mvnu_{z_i} \| \leq \frac{4C_2}{n^2\alpha}\sqrt{\dfrac{n_h\log n}{\rho_n}} \leq \frac{4C_2}{n\alpha}\sqrt{\dfrac{\log n}{n\rho_n}}
$$
under Lemma \ref{Lemma_4_SBM}. Therefore,
$$
\|\bw_i\|\|\mvnu_h-\mvnu_{z_i} \| \leq \frac{8C_2C_6}{n\alpha}\sqrt{\frac{\log n}{n\rho_n}} \max\left\{ \frac{\log n}{n\sqrt{n\alpha}\rho_n}, \frac{1}{n}\sqrt{\frac{\log n}{\rho_n}}\right\}
$$
with probability at least $1-n^{-3}$. 
Choosing
$\beta_{3,\sigma} = 4C_2C_6\dfrac{\log n}{n^{3/2}\alpha\rho_n} \max\left\{ \dfrac{\log n}{\sqrt{n\alpha\rho_n}}, 1\right\}$,
$$
\bbP\left( \dfrac{\beta_{3,\sigma}\Delta_{\min}^2}{2} 
\leq \langle \by_i^* - \bmu_{z_i}, \mvnu_h-\mvnu_{z_i}  \rangle \right) \leq  \frac{1}{n^3}. 
$$

\step{6: Bounding $J_{4,\sigma}$}
The last term $J_{4,\sigma}$ is bound using $\beta_{4,\sigma} = \dfrac{3C_0}{\sqrt{n\rho_n}}$ and Bernstein inequality for sum of Bernoulli entries similar to the proof of Lemma \ref{Lemma_norm_SBM},
\begin{align*}
\bbE J_{4,\sigma} & = \sum_{h\in [K]} \dfrac{1}{n} \sum_{i = 1}^n \bbP \left(\frac{\beta_{4,\sigma} }{2} \|\bmu_{z_i} - \bmu_h\|^2 \leq \langle \bw_i, \bphi_h - \bphi_{z_i} \rangle\right)\\
& \leq \sum_{h\in [K]} \dfrac{1}{n} \sum_{i = 1}^n \bbP \left(\frac{\beta_{4,\sigma} }{2} \frac{\Delta_{\min}^2}{2\epsilon}  \leq  \|\bw_i\| \right) \\
& \leq \sum_{h\in [K]} \dfrac{1}{n} \sum_{i = 1}^n \bbP \left(\frac{\beta_{4,\sigma} }{2} \frac{\sqrt{\rho_n}}{C_0} \leq \|\bw_i\| \right) \\
& \leq K\exp\left( -2C_7 n\rho_n  \right) \leq \exp\left( -C_7 n\rho_n  \right),
\end{align*}
for some constant $C_7$ using $n\rho_n \gg 1$ and the fact that $K$ is fixed.

\step{7: Bounding $J_{1,\epsilon}, J_{2,\epsilon}$}
Then we consider $J_{1,\epsilon}$ and $J_{2,\epsilon}$. With $\beta_{\epsilon} = \dfrac{2C_0}{\sqrt{n\rho_n}}$, we have
\begin{align*}
&\cI\left(\frac{\beta_\epsilon}{2}\left\|\bmu_{z_i}-\bmu_h\right\|^2 \leq \langle \be_i, \bmu_h-\bmu_{z_i}\rangle\right) \\
\leq & \cI\left(\beta_\epsilon\sqrt{\frac{1}{2n}} \leq \epsilon \right) \\
\leq & \cI\left( \dfrac{\sqrt{2}C_0}{n\sqrt{\rho_n}} \leq \frac{C_0}{n\sqrt{\rho_n}} \right) = 0
\end{align*}
and 
$$J_{1,\epsilon} = 0. $$
Finally, we can deal with $J_{2,\epsilon}$ using 
$$
\| \bzeta_h -\bzeta_g\| \leq 2\Gamma_s \|\bmu_h-\bmu_g\|    
$$
and $\beta_{2,\epsilon} = \dfrac{C_0}{\sqrt{n\rho_n}}$ to achieve 
\begin{align*}
& \cI\left(\frac{\beta_{2,\epsilon} }{2} \|\bmu_{z_i} - \bmu_h\|^2 \leq \langle \be_i, \bzeta_h -\bzeta_{z_i} \rangle\right) \\
\leq & \cI\left(\frac{\beta_{2,\epsilon} }{2} \|\bmu_{z_i} - \bmu_h\|^2 \leq  2\epsilon\Gamma_s \|\bmu_h-\bmu_{z_i}\| \right) \\
\leq & \cI\left( \frac{\beta_{2,\epsilon} }{2} \|\bmu_{z_i} - \bmu_h\|^2 <  \epsilon \|\bmu_h-\bmu_{z_i}\| \right) \\
\leq & \cI\left(\frac{\beta_{2,\epsilon} }{\sqrt{n}} < \epsilon \right)  = 0. 
\end{align*}
Summarizing the results above, 
\begin{align*}
\E (A_{s+1}) &\leq \E (J_{1,\sigma}) 
+ \E(J_{2,\sigma}\cI(\cG)) 
+ \E(J_{3,\sigma}\cI(\cG)) 
+ \E (J_{4,\sigma}) \\
&\quad + \E(J_{1,\epsilon}) 
+ \E(J_{2,\epsilon}\cI(\cG)) 
+ \bbP (\cG^c) \\
&\leq \exp\left( -\frac{C_5\beta_{\sigma}^2}{8}n\rho_n\right) 
+ \left( \dfrac{1}{ \rho_{\sigma}} + \dfrac{1}{\rho_{\sigma} \rho_{\epsilon}^2} \right)\E (A_s) \\
&\quad + \frac{1}{n^3} 
+ \exp\left( -C_7 n\rho_n \right) 
+ \frac{1}{n^3},
\end{align*}
where $\beta_\sigma = 1 - \frac{c_2}{\sqrt{n\rho_n}} -\beta_{1,\epsilon} - \beta_{2,\sigma} - \beta_{3,\sigma} - \beta_{4,\sigma} = 1 - o(1) \geq \frac{2}{3}$
using the assumption $n\rho_n\gg 1$. 
Denote $\fp = \dfrac{1}{ \rho_{\sigma}} + \dfrac{1}{\rho_{\sigma} \rho_{\epsilon}^2}$.
By recursion, 
\begin{align*}
\E (A_s) &\leq \left( \fp \right)^s \E(A_s) + 
\dfrac{1 -\left( \fp  \right)^{s+1} }{1- \fp  } \left[ \exp\left(-\frac{C_5}{18}n\rho_n\right) + \exp\left( -C_7 n\rho_n \right) + \frac{2}{n^3} \right].
\end{align*}
Since $\rho_{\sigma} \gg 1 $, $\rho_{\epsilon} \gg 1$,
\begin{equation}
\dfrac{1 -\bigg ( \dfrac{1}{ \rho_{\sigma}} + \dfrac{1}{\rho_{\sigma} \rho_{\epsilon}^2} \bigg )^{s+1} }{1- \dfrac{1}{ \rho_{\sigma}} - \dfrac{1}{\rho_{\sigma} \rho_{\epsilon}^2}\ }  \leq 2,
\end{equation}
and when $s \geq 4\log n$,
\begin{equation*}
\left( \dfrac{1}{ \rho_{\sigma}} + \dfrac{1}{\rho_{\sigma} \rho_{\epsilon}^2} \right)^s \E(A_s) \leq  \left( \dfrac{1}{ \rho_{\sigma}} + \dfrac{1}{\rho_{\sigma} \rho_{\epsilon}^2} \right)^{\log (n^4)} \E(A_s) \leq (n^4)^{\log \left( \frac{1}{ \rho_{\sigma}} + \frac{1}{\rho_{\sigma} \rho_{\epsilon}^2} \right)} \leq \dfrac{1}{n^3} .
\end{equation*}
Thus, when $s \geq 4\log n$, 
\begin{equation*}
\E (A_{s+1}) \leq  2 \exp\left( -\frac{C_5}{18}n\rho_n\right) + 2\exp\left( -C_7 n\rho_n \right) +  \frac{5}{n^3} .
\end{equation*}
By Markov's inequality, for any $t>0$,
\begin{equation}
\mathbb{P}\left( A_s \geq t\right) \leq \frac{1}{t} \mathbb{E} A_s \leq \frac{2}{t}\exp\left( -\frac{C_5}{18}n\rho_n \right) + \frac{2}{t}\exp\left( -C_7 n\rho_n \right) + \frac{5}{t n^3} .\label{A_s_SBM}
\end{equation}
If $n\rho_n \leq \max\left\{ \frac{36}{C_5},\frac{2}{C_7}\right\}\log n $, choose 
$$t=\exp \left(-\left(C- \frac{1}{\sqrt{n\rho_n}}\right)n\rho_n \right)
$$ 
where $C=\min\left\{\frac{C_5}{18} , C_7\right\} $
and we have
$$
\mathbb{P}\left\{A_s \geq t \right\} \leq \frac{1}{n}+ 4\exp(-\sqrt{n\rho_n}) .
$$
Otherwise, since $A_s$ only takes discrete values of $\left\{0, \frac{1}{n}, \cdots, 1\right\}$, choosing $t=\frac{1}{n}$ in \eqref{A_s_SBM} leads to
$$
\mathbb{P}\left\{A_s>0\right\}=\mathbb{P}\left\{A_s \geq \frac{1}{n}\right\} \leq 2n \exp (-2 \log n)+ 2 n \exp (-2\log n) + \frac{3}{n^2} \leq \frac{5}{n}. 
$$
The proof is complete.

\end{proof}

\subsection{Spectral clustering in noisy stochastic block models}
\textbf{Proof of Theorem~\ref{thm:noisy-sbm}.}
\begin{proof}
Denote $\zb \in \bbR^{n}$ as the community label vector, with $\zb : [n] \to [K]$. Under the noisy SBM described in Section~\ref{sec:noisy-SBM}, we have
\[
\bbP(\Yb_{ij} = 1) = (1-\alpha_n - \beta_n)\,\Bb_{\zb_i \zb_j} + \alpha_n .
\]
Therefore, the observed network $\Yb$ follows a Bernoulli network model with expectation
\[
\bbE(\Yb) = (1-\alpha_n - \beta_n)\Ab^* + \alpha_n =: \Bb_{\Yb}.
\]
Under Assumption~\ref{asm:noisy-SBM}(c), it is straightforward to verify that $\operatorname{rank}(\Bb_{\Yb}) = K$ and that $\Bb_{\Yb}$ preserves the community structure. Consequently, we can apply an argument similar to that used in the proof of Theorem~\ref{thm:sbm_K} to the spectral embedding of $\Yb$ to obtain the desired result.
\end{proof}

\subsection*{Useful results}
We provide some facts and technical lemmas that will be used to prove the results in Sections~\ref{sec:SC} and~\ref{sec:MDS}. 

We adopt notation from \cite{wang2019some} and  \cite{abbe2022ellp} to make the proof clear. Define the sub-Gaussian norms $\|X\|_{\psi_2}=\sup _{p \geq 1}\left\{p^{-1 / 2} \mathbb{E}^{1 / p}|X|^p\right\}$ for random variable $X$ and $\|\boldsymbol{X}\|_{\psi_2}=\sup _{\|\boldsymbol{u}\|_2=1}\|\langle\boldsymbol{u}, \boldsymbol{X}\rangle\|_{\psi_2}$ for random vector $\boldsymbol{X}$. We adopt the following convenient notation from \cite{wang2019some} to make probabilistic statements compact.

\begin{dfn}
Let $\left\{X_n\right\}_{n=1}^{\infty},\left\{Y_n\right\}_{n=1}^{\infty}$ be two sequences of random variables and $\left\{r_n\right\}_{n=1}^{\infty} \subseteq(0,+\infty)$ be deterministic. We write
$$
X_n=O_{\mathbb{P}}\left(Y_n ; r_n\right)
$$
if there exists a constant $C_1>0$ such that
$$
\forall C>0, \exists C^{\prime}>0 \text { and } N>0 \text {, s.t. } \quad \mathbb{P}\left(\left|X_n\right| \geq C^{\prime}\left|Y_n\right|\right) \leq C_1 e^{-C r_n}, \quad \forall n \geq N .
$$
We write $X_n=o_{\mathbb{P}}\left(Y_n ; r_n\right)$ if $X_n=O_{\mathbb{P}}\left(w_n Y_n ; r_n\right)$ holds for some deterministic sequence $\left\{w_n\right\}_{n=1}^{\infty}$ tending to zero.
\end{dfn}

Consider the sub-Gaussian mixture model defined in Section \ref{sec:SC}.

\begin{lemma}\label{subG_tech_lemma}
    \begin{equation}
    \left\|\widebar\bw\right\| = O_{\bbP}\left(\sigma\max\left\{1,\sqrt{\frac{p}{n}}\right\};n \right) = O_{\bbP}\left(\sigma\max\left\{\sqrt{\frac{p}{n}},\sqrt{\frac{\log n}{n}}\right\};\log n \right),      \label{eqn:tech_1}
    \end{equation}
    \begin{equation}
    \sup_{\bu\in\bbS^{n-1}}\left\|\sum_{i=1}^nu_i\bw_i\right\| = O_{\bbP}\left( \sigma\max\left\{\sqrt{n},\sqrt{p}\right\} ;n \right), \label{eqn:tech_2}
    \end{equation}
    \begin{equation}
    \max _{i \in[n]}\left\|\bw_i\right\| = O_{\bbP}\left(\sigma \max \left\{\sqrt{n},\sqrt{p}\right\} ; \log n\right).\label{eqn:tech_3}
    \end{equation}
    For any fixed $m\in[n]$,
    \begin{equation}
    \max _{i \in[n]} \left| \langle \bw_m, \widebar\bw \rangle \right| = O_{\bbP}\left( \sigma^2\left(\frac{1}{\sqrt{n}}\max\left\{\sqrt{p\log n},\log n\right\} + \frac{p}{n} \right) ;\log n \right), \label{eqn:tech_4}
    \end{equation}
\begin{equation}
    \|\widebar\bLambda^{-1/2} \widebar\Ub^{\top}(\Jb\widebar\Xb)\bw_m\|_{\psi_2} \lesssim \sqrt{\kappa}\sigma.\label{eqn:tech_5}
\end{equation}
    
\end{lemma}

\begin{proof}
    From Lemma \ref{lem:subg-sum} and \ref{lem:cluster-sum}, we obtain \eqref{eqn:tech_1}.

    For \eqref{eqn:tech_2},
    $$
    \left\|\sum_{i=1}^nu_i\bw_i\right\| \leq \|\Wb\| = O_{\bbP}\left( \sigma\max\left\{\sqrt{n},\sqrt{p}\right\} ;n \right),
    $$
    where the equality uses the third conclusion of Lemma H.1 in \cite{abbe2022ellp}.
    
    From the third result of Lemma H.1 in \cite{abbe2017community}, we have \eqref{eqn:tech_3}.

    From Lemma \ref{lem:inner-prod} in the proof of the main theorem \ref{thm-lloyd-guarantee} with \\
    $t = \max\{4\sqrt{p\log n},4\log n \}$ and $\delta = n^{-4}$, we obtain \eqref{eqn:tech_4}. 
    For conclusion \eqref{eqn:tech_5}
$$
\|\widebar{\Xb}^{\top}\widebar{\Ub}\widebar{\bLambda}^{-1/2}\|_2 \leq \|\widebar{\Xb}\|_2 \|\widebar{\Ub}\|_2 \|\widebar{\bLambda}^{-1/2}\|_2 = \left(\frac{\bar\lambda_1}{\bar\lambda_K}\right)^{1/2} = \kappa^{1/2},
$$
using the fact that $\|\widebar{\Xb}\|_2 = \bar\lambda_1^{1/2}$ and $\|\widebar{\bLambda}^{-1/2}\|_2 = \bar\lambda_K^{1/2}$.
    
\end{proof}

\subsection{Spectral clustering of mixture models}
Theorem \ref{thm:spect-clust} is proved in this section. 

We outline the main steps of the proof below:
\begin{enumerate}
    \item 
    Derive bounds on the random perturbation errors 
    $\epsilon_1 := \|\Eb_{11}\|_{2\to\infty}$ and 
    $\epsilon_2 := \|\Eb_{12}\|$;
    \item 
    Establish a lower bound on the minimum inter-cluster separation 
    $\Delta_{\min}$;
    \item 
    Verify that the conditions of Theorem~\ref{thm-lloyd-guarantee} are satisfied;
    \item 
    Obtain the misclustering error bound by applying 
    Theorem~\ref{thm-lloyd-guarantee}.
\end{enumerate}

\begin{proof}
Since $\Xb = \bar{\Xb} + \Wb$, where $\bar{\Xb}$ is of rank $K$ by Assumption~\ref{asm:SC} and $\Wb$ has row-independent sub-Gaussian error, it suffices to verify the regularity and concentration assumptions required to apply Corollary~2.1 in \cite{abbe2022ellp}. From the proof of Theorem 3.1 in \cite{abbe2022ellp}, we know that the regularity assumption is satisfied by Assumption \ref{asm:SC} (a) and the concentration assumption is satisfied by Assumption \ref{asm:SC} (b) and (c) since 
$$
\operatorname{SNR}  = \min\left\{ \dfrac{\bar{s}^2}{\sigma^2}, \dfrac{n\bar{s}^4}{p\sigma^4}\right\}
\geq \min\left\{ \dfrac{\bar{s}^2}{\sigma^2}, \dfrac{n\bar{s}^4}{\log n(p+\log n)\sigma^4} \right\} \gg 1. 
$$


\textbf{Control $\epsilon_1 := \|\Eb_{11}\|$. }
Therefore we can apply Corollary 2.1. in \cite{abbe2022ellp} to get the decomposition of $\Ub\bLambda^{1/2}$ as 
$$
\Ub\bLambda^{1/2}\Ob = \widebar{\Ub}\widebar{\bLambda}^{1/2} + \cH\left(\Wb \Xb^{\top}\right) \widebar{\Ub}\widebar{\bLambda}^{-1/2} + \Eb_{12},
$$
where $\Ob\in \bbR^{K\times K}$ is a rotation matrix. Define $$ \dfrac{1}{\sqrt{\log n}} \|\widebar{\Ub}\|_{2\rightarrow\infty}\|\widebar{\bLambda}^{1/2}\|_2 \lesssim \sqrt{\frac{\widebar\lambda_1}{n\log n}} =: \epsilon_2$$ 
then we have $\|\Eb_{12}\|_{2\rightarrow\infty} \lesssim \epsilon_2$ with probability $1-o(1)$ by Corollary 2.1 of \cite{abbe2022ellp}. 
We can further decompose the embedding matrix as:
$$
\Ub\bLambda^{1/2}\Ob = \widebar{\Ub}\widebar{\bLambda}^{1/2} + \Eb_{11}+ \Eb_{12} + \Eb_2,
$$
where 
$\Eb_2 = \cH\left(\Wb \widebar{\Xb}^{\top}\right) \widebar{\Ub}\widebar{\bLambda}^{-1/2}$
and 
$\Eb_{11} = \cH\left(\Wb \Wb^{\top}\right) \widebar{\Ub}\widebar{\bLambda}^{-1/2}$.

In order to apply Theorem \ref{thm-lloyd-guarantee}, the goal then is to show $\Eb_2$ has sub-Gaussian concentration properties and $\|\Eb_{11}\|_{2\rightarrow\infty}$ is small. 

The matrix $\Eb_2$ has row-independent sub-Gaussian with parameter $\sigma_{S} =\sqrt{\kappa}\sigma$ from \eqref{eqn:tech_5}. 

\textbf{Control $\epsilon_2 := \|\Eb_{12}\|$. }
For the matrix $\Eb_{11}$, we would like to bound the $2\rightarrow\infty$ norm using
$$
\left\| \cH\left(\Wb \Wb^{\top}\right) \widebar{\Ub}\widebar{\bLambda}^{-1/2}\right\|_{2\rightarrow\infty} \leq 
\left\| \cH\left(\Wb \Wb^{\top}\right) \widebar{\Ub}\right\|_{2\rightarrow\infty} \left\|\widebar{\bLambda}^{-1/2} \right\|_2,
$$
and thus
\begin{align*}
&\bbP\left( \left\|\cH\left(\Wb \Wb^{\top}\right) \widebar{\Ub}\right\|_{2,\infty} > t \right) 
= \bbP\left( \max_{i\in[n]}\left\|\left[ \cH\left(\Wb \Wb^{\top}\right) \widebar{\Ub}\right]_{i\cdot}\right\|_{2} > t \right) \\
\leq & \sum_{i=1}^n \bbP\left( \left\|\left[ \cH\left(\Wb \Wb^{\top}\right) \widebar{\Ub}\right]_{i\cdot}\right\|_{2} > t \right) .
\end{align*}

Define the event $\cG$ as the setting where the above inequality and the inequality in \ref{lem:cluster-sum} hold. Then we have
$\bbP(\cG)\geq 1-\frac{2}{n^3}$. Assume that the event $\cG$ holds.
Based on the bound 
\begin{align*}
\|\sum_{j \neq i} \bw_j \widebar{\Ub}_{jl}\|_2 
&\leq \sum_{s=1}^K \|\sum_{j\in\cC_s,j\neq i} \bw_j\widebar{\Vb}_{sl}/\sqrt{n_s}\|_2 = \sum_{s=1}^K \frac{|\widebar{\Vb}_{sl}|}{\sqrt{n_s}}\| \sum_{j\in\cC_s,j\neq i} \bw_j \|_2 \\
&\leq 
3\sigma\sqrt{p+\log n}\sum_{s=1}^K|\widebar{\Vb}_{sl}| + 2\sqrt{2}\sigma \sqrt{\log n}/\sqrt{n\alpha} =: t_{l},
\end{align*}

we have 
\begin{align*}
&\bbP\left( \langle \bw_i,\sum_{j \neq i} \bw_j \widebar{\Ub}_{jl}\rangle > t \right) 
= \bbP\left( \langle \bw_i,
\frac{\sum_{j \neq i} \bw_j \widebar{\Ub}_{jl}}{\|\sum_{j \neq i} \bw_j \widebar{\Ub}_{jl}\|_2}\rangle > \frac{t}{\|\sum_{j \neq i} \bw_j \widebar{\Ub}_{jl}\|_2} \right) \\
\leq & \bbP\left( \langle \bw_i,
\frac{\sum_{j \neq i} \bw_j \widebar{\Ub}_{jl}}{\|\sum_{j \neq i} \bw_j \widebar{\Ub}_{jl}\|_2}\rangle > \frac{t}{t_{l}} \right) \leq \inf_{\theta>0}e^{-\theta t/t_l} \exp\left(\frac{1}{2}\theta^2\sigma^2\right) \\
=& \exp\left(-\frac{t^2}{2t_l^2\sigma^2}\right). 
\end{align*}
Therefore, 
$$
\left|\langle \bw_i,\sum_{j \neq i} \bw_j \widebar{\Ub}_{jl}\rangle\right| \leq  \sqrt{2}t_l\sigma\sqrt{\log(2/\delta)},
$$
with probability greater than $1-\delta$ for each fixed $l\in[K]$.  
Following the decomposition, 
\begin{align*}
&\left\|\left[ \cH\left(\Wb \Wb^{\top}\right) \widebar{\Ub}\right]_{i\cdot}\right\|_{2}^2 \\
=& \left\|\left( \langle \bw_i,\sum_{j=1,j\neq i}^n \bw_j \widebar{\Ub}_{jl}\rangle \right)_{l=1}^k\right\|_{2}^2 
= \sum_{l=1}^K \left(\langle \bw_i,\sum_{j \neq i} \bw_j \widebar{\Ub}_{jl}\rangle  \right)^2\\
\leq& \sum_{l=1}^K \left(\sqrt{2}t_l\sigma\sqrt{\log(2/\delta)}\right)^2 
= 2\sigma^2 \log(2/\delta) \sum_{l=1}^K t_l^2 \\
= & 2\sigma^2 \log(2/\delta) \sum_{l=1}^K \left( 3\sigma\sqrt{p +\log n}\sum_{s=1}^K|\widebar{\Vb}_{sl}| + 2\sqrt{2}\sigma \sqrt{\log n}/\sqrt{n\alpha} \right)^2 \\
\leq & 4\sigma^2\log(2/\delta)\left( 
9\sigma^2 (p+\log n)\sum_{s=1}^K (\sum_{s=1}^K|\widebar{\Vb}_{sl}| )^2 + 8\sigma^2K\frac{\log n}{n\alpha}
\right)\\
\leq & 4K\sigma^4\log(2/\delta)\left( 
9K (p+\log n) + \frac{8\log n}{n\alpha}
\right),
\end{align*}
with probability greater than $1-K\delta$. 
Therefore, 
\begin{equation}\label{eqn:tight_concentration}
\left\|\cH\left(\Wb \Wb^{\top}\right) \widebar{\Ub}\right\|_{2\rightarrow\infty}^2 \leq  4K\sigma^4\log(2/\delta)\left( 
9K (p+\log n) + \frac{8\log n}{n\alpha}
\right),     
\end{equation}
with probability $1-nK\delta$ using the union argument. 
Choosing $\delta = \frac{1}{n^3K}$, we obtain  
\begin{align*}
& \left\| \cH\left(\Wb \Wb^{\top}\right) \widebar{\Ub}\widebar{\bLambda}^{-1/2}\right\|_{2\rightarrow\infty}\\
\leq &
\sqrt{4K\sigma^4\log(2Kn^3)\left( 
9K (p+\log n) + \frac{8\log n}{n\alpha}
\right)}/\sqrt{\bar\lambda_K}=:\epsilon_1,
\end{align*}
with probability $1-\frac{1}{n^2}$. 

From the proof of Theorem 3.1 in \cite{abbe2022ellp}, we have
$$
\Delta_{\min}:= \min_{i\in\cC_h,j\in\cC_g,i\neq j}\|(\widebar\Ub\widebar\bLambda^{1/2})_i - (\widebar\Ub\widebar\bLambda^{1/2})_j\| \geq \frac{\widebar\lambda_K^{1/2}}{\max_{l\in[K]}\sqrt{n_l}} \geq \frac{\widebar\lambda_K^{1/2}}{\sqrt{n}},
$$
and 
$$
\kappa \leq \kappa_0^2,\quad \widebar\lambda_K \geq \frac{\bar{s}^2 n}{2\kappa_0^2}. 
$$

\textbf{Check the conditions under the main theorem.}
Under Assumption \ref{asm:SC}, the conditions required in applying Theorem \ref{thm-lloyd-guarantee} are satisfied since 
$$\rho_{\sigma} = \frac{\Delta_{\min}}{\sigma_S}\sqrt{\frac{\alpha}{1+\frac{K r}{n}}} \gtrsim \dfrac{\widebar\lambda_K^{1/2}}{\sqrt{n\kappa}\sigma} 
\geq \dfrac{\bar{s}}{\sqrt{2}\kappa_0^2\sigma}  \gtrsim  \dfrac{\bar{s}}{\sigma} 
\geq C$$
and 
\begin{align*}
\rho_{\epsilon} 
& \gtrsim
\min\left\{
\frac{\widebar\lambda_K}{\sqrt{n \log n\left( p+\log n\right)}\sigma^2}
, \sqrt{\frac{\log n}{\kappa}}
\right\} \\
& \gtrsim \min\left\{
\frac{\sqrt{n}\bar{s}^2}{\sqrt{\log n\left( p+\log n\right)}\sigma^2}
, \sqrt{\frac{\log n}{\kappa}}
\right\} \geq C
\end{align*}
for some large $C$ and assuming a large sample size. 
Applying  Theorem \ref{thm-lloyd-guarantee}, we have that under initialization condition $G_0 < (1-\epsilon_0)\sqrt{\frac{c_1}{C_1}}$ for some small $\epsilon_0$,
\begin{align*}
A_s \leq \max \Bigg\{ 
& \exp \left( - \frac{\bar{s}^2}{32\kappa_0^4\beta \sigma^2} \right), 
\exp\left( -c\frac{\bar{s}^{3}\sqrt{n}}{\sigma^3\sqrt{\log n(p+\log n)}} \right), \notag \\
& \exp\left( -c\frac{\sqrt{\log n}\bar{s}}{\sigma} \right)
\Bigg\}.
\end{align*}
where $\beta = \dfrac{\max_l n_l}{n}$ for all $s\geq 4\log n $ for some constant $c$ with probability 
\begin{align*}
    & 1- \exp \left( - \frac{\bar{s}}{ \sigma}  \right) -  \exp\left( - \left(\frac{\bar{s}}{\sigma} \right)^{3/2} \left(\frac{n}{\log n(p+\log n)} \right)^{1/4} \right) \\
    & - \exp\left( -(\log n)^{1/4}\sqrt{\frac{\bar{s}}{ \sigma}} \right) - o(1).
\end{align*}


\end{proof}

\subsection{Multidimensional scaling of mixture models}
In this section, we prove Theorem~\ref{thm:mds}. We first present several auxiliary lemmas used to control the perturbation terms arising in the analysis.

The proof proceeds as follows:
\begin{enumerate}
\item 
In Section~\ref{lem:mds}, we establish auxiliary lemmas that control the relevant perturbation terms.
\item 
In Section~\ref{proof:mds}, we derive bounds on the random perturbation errors
\[
\epsilon_1 := \|\Eb_{11}\|_{2\to\infty},\qquad  
\epsilon_2 := \|\Eb_{12}\|,\qquad 
\epsilon_3 := \|\Eb_{13}\|,
\]
and verify that these bounds satisfy the conditions of Theorem~\ref{thm-lloyd-A_s_bound}, which completes the proof.
\end{enumerate}

\subsubsection{Technical lemmas}\label{lem:mds}

\begin{lemma}\label{lem:spectral_norm_MDS_mean}
$$
\|\widebar\bLambda^{-1/2} \widebar\Ub^{\top}(\Jb\widebar\Xb)\widebar\bw\| = O_{\bbP}\left(\sqrt{\frac{K+\log n}{n}\kappa}\sigma; \log n\right).
$$
\end{lemma}

\begin{proof}
    By sub-Gaussian property of $\{\bw_m\}_{m=1}^n$, we have
    $$    \|\widebar\bLambda^{-1/2} \widebar\Ub^{\top}(\Jb\widebar\Xb)\widebar\bw_m\|_{\psi_2} \lesssim \sigma \|\Jb\widebar\Xb\|\lambda_{K-1}^{-1/2} = \sqrt{\kappa}\sigma 
    $$
    and using independence of $\{\bw_m\}_{m=1}^n$,
    $$    \left\|\sum_{m=1}^{n}\widebar\bLambda^{-1/2} \widebar\Ub^{\top}(\Jb\widebar\Xb)\widebar\bw_m\right\|_{\psi_2} \lesssim \sqrt{n\kappa}\sigma.
    $$
    By Lemma \ref{lem:cluster-sum} in the proof of the main theorem, we have
    $$
    \left\|\frac{1}{n}\sum_{m=1}^{n}\widebar\bLambda^{-1/2} \widebar\Ub^{\top}(\Jb\widebar\Xb)\widebar\bw\right\| = O_{\bbP}\left(\sqrt{\frac{K+\log n}{n}\kappa}\sigma; \log n\right).
    $$
\end{proof}

In order to decompose the embedding matrix $\Ub\bLambda^{1/2}$ in a similar fashion as in the proof of Theorem \ref{thm:spect-clust}, we need the following technical lemmas. 

\begin{lemma}[Modified Lemma H.1 of  \cite{abbe2022ellp}]\label{lemma_H_1}
    Under assumption that $\{\bw_i\}_{i=1}^n$ are independent, zero-mean sub-Gaussian random vectors in $\bbR^p$ with parameter $\sigma^2$, we have
\begin{align} 
&\left\|\cH\left((\Jb\Wb) (\Jb\Wb)^{\top}\right)\right\|_2 = O_{\bbP}\left( \sigma^2\max\{n,p\};n \right), \\ 
& \max_{i\in[n]}\left\|(\Jb\Wb)_i\right\|^2=O_{\bbP}\left( \sigma^2\max\{n,p\};n \right), \\ 
& \left\|(\Jb\Wb) (\Jb\Wb)^{\top}\right\|_2=O_{\bbP}\left( \sigma^2\max\{n,p\};n \right). 
\end{align}
    
\end{lemma}

\begin{proof}
    By definition,
    \begin{align}
       &\left\|\cH\left((\Jb\Wb) (\Jb\Wb)^{\top}\right)\right\|_2\\
       =&\sup_{\bu \in \bbS^{n-1}}\left|\bu^{\top} \cH\left((\Jb\Wb)(\Jb\Wb)^{\top}\right) \bu\right| \nonumber\\
       =&\sup _{\bu \in \bbS^{n-1}}\left|\sum_{i \neq j} u_i u_j\left\langle\bw_i-\widebar\bw, \bw_j-\widebar\bw \right\rangle\right| \nonumber\\
       \leq& \sup _{\bu \in \bbS^{n-1}}\left|\sum_{i \neq j} u_i u_j\left\langle\bw_i,\bw_j\right\rangle\right|+ 2\sup _{\bu \in \bbS^{n-1}}\left|\sum_{i \neq j} u_i u_j\left\langle\bw_i,\widebar\bw\right\rangle\right| + \sup _{\bu \in \bbS^{n-1}}\left|\sum_{i \neq j} u_i u_j\left\langle\widebar\bw,\widebar\bw\right\rangle\right| \label{lemma_H_1_decomp}.
    \end{align}
Using the result from Lemma H.1 in \cite{abbe2022ellp},
the first part in \eqref{lemma_H_1_decomp} can be bounded as
\begin{equation}
    \sup _{\bu \in \bbS^{n-1}}\left|\sum_{i \neq j} u_i u_j\left\langle\bw_i,\bw_j\right\rangle\right| = O_{\bbP}\left(\sigma^2\max \left\{n, \sqrt{np}\right\} ; n\right).\label{lemma_H_1_bound_1}
\end{equation}
Using Lemma \ref{subG_tech_lemma}, we can control the second term of \eqref{lemma_H_1_decomp} as 
\begin{align*}
    \sup _{\bu \in \bbS^{n-1}}\left|\sum_{i \neq j} u_i u_j\left\langle\bw_i,\widebar\bw\right\rangle \right| 
    &=  \sup _{\bu \in \bbS^{n-1}}\left|\sum_{i=1}^n (\sum_{j\neq i}u_j)u_i\left\langle\bw_i,\widebar\bw\right\rangle \right| \\
    &\leq  \sup _{\bu \in \bbS^{n-1}}\sqrt{n}\left|  \left\langle\sum_{i=1}^nu_i\bw_i,\widebar\bw\right\rangle \right|\\
    &\leq  \sup _{\bu \in \bbS^{n-1}}\sqrt{n} \left\|\sum_{i=1}^nu_i\bw_i\right\|\left\|\widebar\bw\right\|\\
    &= O_{\bbP}\left( \sigma^2\max\{n,p\};n \right),
\end{align*}
and the third part is bounded via
\begin{align*}
    \sup _{\bu \in \bbS^{n-1}}\left|\sum_{i \neq j} u_i u_j\left\langle\widebar\bw,\widebar\bw\right\rangle\right| 
    &=  \sup _{\bu \in \bbS^{n-1}}\left|\sum_{i \neq j} u_i u_j\right| \left\|\widebar\bw\right\|^2  \\
    &\leq  \sup _{\bu \in \bbS^{n-1}}n \left\|\widebar\bw\right\|^2 \\
    &= O_{\bbP}\left( \sigma^2\max\{n,p\};n \right).
\end{align*}
Combing the three parts together, we obtain the first conclusion.

Using Lemma \ref{subG_tech_lemma}
$$
\begin{aligned}
 \max_{i\in[n]}\left\|(\Jb\Wb)_i\right\|^2 =& \max_{i\in[n]}\left\|\bw_i-\widebar\bw\right\|^2 \\
\leq & \max_{i\in[n]}\left\|\bw_i\right\|+\left\|\widebar\bw\right\|^2\\
= & O_{\bbP}\left(\sigma^2\max \left\{n,p\right\} ; n\right),
\end{aligned}
$$
and
\begin{align*}
& \left\|(\Jb\Wb) (\Jb\Wb)^{\top}\right\|_2 \\
\leq & 
\left\|\cH\left((\Jb\Wb) (\Jb\Wb)^{\top}\right)\right\|_2 
+ \max_{i\in[n]}\left\|(\Jb\Wb)_i\right\|^2 = O_{\bbP}\left(\sigma^2\max \left\{n,p\right\} ; n\right). 
\end{align*}

\end{proof}

\begin{rem}
    This modified version differs from Lemma H.1 of \cite{abbe2022ellp} in that the first bound becomes larger in our MDS setting due to centering. Specifically, in the high-dimensional setting where $p\gtrsim n$, the bound for $\left\|\cH\left((\Jb\Wb) (\Jb\Wb)^{\top}\right)\right\|_2$ is is multiplied by $\sqrt{\dfrac{p}{n}}$ owing to the fact that the mean vector $\widebar{\bw}$ has dimension $p$, leading to its influence accumulating as $p$ grows larger. Consequently, this leads to a stronger condition for a similar decomposition, as presented in the modified Corollary 2.1 of \cite{abbe2022ellp}, in the high-dimensional setting where $p\gtrsim n$.
\end{rem}

\begin{lemma}[Modified Lemma H.2 of \cite{abbe2022ellp}]\label{lemma_H_2}
    Assume $\{\bw_i\}_{i=1}^n$ are independent, zero-mean sub-Gaussian random vectors in $\bbR^p$ with parameter $\sigma^2$. Let $\{\Vb^{(m)}\}\subset \bbR^{n\times r}$ be random matrices such that $\Vb^{(m)}$ is independent of $\bw_m$. Then,
\begin{align*}
& \max_{m\in[n]}\left\|\cH\left((\Jb\Wb)(\Jb\Wb)^{\top}\right)_m\Vb^{(m)}\right\| \\
= & \sqrt{r}\max_{m\in[n]}\|\Vb^{(m)}\|_2 O_{\bbP}\left(\sigma^2\max\left\{\dfrac{p}{\sqrt{n}},\sqrt{p\log n},\sqrt{n\log n}\right\};\log n\right).     
\end{align*}

\end{lemma}

\begin{proof}
    By definition
    \begin{align*}
        &\max_{m\in[n]}\left\|\cH\left((\Jb\Wb)(\Jb\Wb)^{\top}\right)_m\Vb^{(m)}\right\|\\
        = & \max_{m\in[n]}\left\| \sum_{j\neq m}\langle (\Jb\Wb)_m, (\Jb\Wb)_j \rangle \Vb^{(m)}_j\right\| \\
        = & \max_{m\in[n]}\left\| \sum_{j\neq m}\langle \bw_m-\widebar\bw, \bw_j-\widebar\bw \rangle \Vb^{(m)}_j\right\| \\
        \leq & \max_{m\in[n]}\left\| \sum_{j\neq m}\langle \bw_m, \bw_j \rangle \Vb^{(m)}_j\right\| + 
        \max_{m\in[n]}\left\| \sum_{j\neq m}\langle \bw_m, \widebar\bw \rangle \Vb^{(m)}_j\right\|  \\
        &+ \max_{m\in[n]}\left\| \sum_{j\neq m}\langle \widebar\bw, \bw_j \rangle \Vb^{(m)}_j\right\| + \max_{m\in[n]}\left\| \sum_{j\neq m}\langle \widebar\bw, \widebar\bw \rangle \Vb^{(m)}_j\right\|.
    \end{align*}
    We are going to bound each of the four terms.
    From Lemma H.2 of \cite{abbe2022ellp} with $p \asymp \log n$, we have 
    $$
    \max_{m\in[n]}\left\| \sum_{j\neq m}\langle \bw_m, \bw_j \rangle \Vb^{(m)}_j\right\| = \sqrt{r\log n}\max_{m\in[n]}\|\Vb^{(m)}\|_2O_{\bbP}\left(\sigma^2\max\{\sqrt{n},\sqrt{p}\};\log n\right).
    $$
    Using the fact $\left\|\sum_{j\neq m}\Vb^{(m)}_j\right\| = \left\|\Vb^{(m)\top}(\Ib_n-\be_m\be_m^{\top})\bone_n\right\| \leq \sqrt{n}\|\Vb^{(m)}\|$ and Lemma \ref{subG_tech_lemma} \eqref{eqn:tech_4}
    \begin{align*}
        & \max_{m\in[n]}\left\| \sum_{j\neq m}\langle \bw_m, \widebar\bw \rangle \Vb^{(m)}_j\right\| 
        = \max_{m\in[n]}\left\| \langle \bw_m, \widebar\bw \rangle \sum_{j\neq m}\Vb^{(m)}_j\right\| \\
        \leq & \max_{m\in[n]}\left\| \langle \bw_m, \widebar\bw \rangle \right\|  \max_{m\in[n]}\left\|\sum_{j\neq m}\Vb^{(m)}_j\right\| \\
        = & O_{\bbP}\left(\sigma^2 \left( \max\left\{\sqrt{p\log n},\log n \right\} + \dfrac{p}{\sqrt{n}} \right); \log n\right)\max_{m\in[n]}\|\Vb^{(m)}\| 
    \end{align*}
    For the third term, 
    \begin{align*}
        \left\| \sum_{j\neq m}\langle \widebar\bw, \bw_j \rangle \Vb^{(m)}_j\right\| ^2
        = & \sum_{l=1}^r \left| \sum_{j\neq m}\langle \widebar\bw, \bw_j \rangle \Vb^{(m)}_{jl}\right|^2 \\
        = & \sum_{l=1}^r \left| \langle \widebar\bw, \sum_{j\neq m}\Vb^{(m)}_{jl}\bw_j \rangle \right|^2\\
        \leq & \left\|\widebar\bw\right\|^2 
        \sum_{l=1}^r \left\| \sum_{j\neq m}\Vb^{(m)}_{jl}\bw_j \right\|^2 \\
        \leq & \left\|\widebar\bw\right\|^2 
        \sum_{l=1}^r \left\| \Wb^\top\left( \Ib_n - \be_m\be_m^{\top}\right)\Vb_l^{(m)} \right\| \\
        \leq & \left\|\widebar\bw\right\|^2 
        \sum_{l=1}^r \left\| \Vb^{(m)} \right\|_2^2\left\| \Wb\Wb^\top \right\|_2\\
        = & O_{\bbP}\left(r\sigma^4\left\| \Vb^{(m)}\right\|_2^2\max\left\{\frac{p}{n},\frac{\log n}{n} \right\}\max\left\{n,p\right\};\log n\right). 
    \end{align*}
    Therefore,
    \begin{align*}
& \max_{m\in[n]}\left\| \sum_{j\neq m}\langle \widebar\bw, \bw_j \rangle \Vb^{(m)}_j\right\| \\
= & \sqrt{r}\max_{m\in[n]}\left\| \Vb^{(m)}\right\|_2 O_{\bbP}\left(\sigma^2\max\left\{\sqrt{\frac{p}{n}},\sqrt{\frac{\log n}{n}} \right\}\max\left\{\sqrt{n},\sqrt{p}\right\};\log n\right).    \end{align*}
For the last part, we have 
\begin{align*}
    &\max_{m\in[n]} \left\| \sum_{j\neq m}\langle \widebar\bw, \widebar\bw \rangle \Vb^{(m)}_j\right\| \\
    \leq& \max_{m\in[n]} \left\|\widebar\bw \right\|^2\left\|\sum_{j\neq m}\Vb^{(m)}_j\right\|\\
    \leq& \sqrt{n}\max_{m\in[n]}\left\| \Vb^{(m)}\right\|_2O_{\bbP}\left(\sigma^2\max\left\{\dfrac{p}{n},\dfrac{\log n}{n}\right\};\log n \right). 
\end{align*}
By combining four parts together, we obtain the final result. 
\end{proof}

Define $\Hb=\Ub^{\top} \widebar{\Ub} \in \bbR^{r \times r}$ and let $\tilde{\Ub} \tilde{\bLambda} \tilde{\Vb}^{\top}$ denote its singular value decomposition. Define 
$\operatorname{sgn}(\Hb) = \tilde{\boldsymbol{U}} \tilde{\boldsymbol{V}}^{\top}$. 
Based on Lemmas \ref{lemma_H_1}-\ref{lemma_H_2}, we can modify Corollary 2.1 of  \cite{abbe2022ellp} to fit in our setting. 

\begin{theorem}[Modified Corollary 2.1 in \cite{abbe2022ellp}]
\label{thm:modifed_l_p}
    If Assumption \ref{asm:MDS} holds, then \begin{align*}
    \left\|\Ub \bLambda^{1 / 2} \operatorname{sgn}(\Hb)-\Gb \widebar{\Ub} \widebar{\bLambda}^{-1 / 2}\right\|_{2, \infty} & =o_{\bbP}\left(\|\widebar{\Ub}\|_{2, \infty}\left\|\widebar{\bLambda}^{1 / 2}\right\|_2 ; \log n\right), \\ \left\|\Ub \bLambda^{1 / 2} \operatorname{sgn}(\Hb)-\left[\widebar{\Ub} \widebar{\bLambda}^{1 / 2}+\cH\left(\Zb \Xb^{\top}\right) \widebar{\Ub} \widebar{\bLambda}^{-1 / 2}\right]\right\|_{2, \infty} & =o_{\bbP}\left(\|\widebar{\Ub}\|_{2, \infty}\left\|\widebar{\bLambda}^{1 / 2}\right\|_2 ; \log n\right) .
    \end{align*}
\end{theorem}

\begin{proof}[Proof of Theorem \ref{thm:modifed_l_p}]
Consider the centered version of the sub-Gaussian model as 
$$
\Jb\Xb = \Jb\bar{\Xb} + \Jb\Wb.
$$  

Compared with Theorem 2.1 in \cite{abbe2022ellp}, one key difference under the centered sub-Gaussian model is that 
$$
\left\|\mathcal{H}\left((\mathbf{J W})(\mathbf{J W})^{\top}\right)\right\|_2=O_{\mathbb{P}}\left(\sigma^2 \max \{n, p\} ; n\right)
$$
as shown in Lemma \ref{lemma_H_1} while under the model in Section \ref{sec:SC} we have 
$$
\left\|\cH\left(\Wb\Wb^{\top}\right)\right\|_2=O_{\mathbb{P}}\left(\sigma^2 \max \{\sqrt{np}, n \} ; n\right).
$$
Another difference is that, due to the centering effect, we can only establish Lemma \ref{lemma_H_2}, which is a modified version of Lemma H.2 in \cite{abbe2022ellp} under the \(2\to\infty\) norm.

With slight modification, under Assumption \ref{asm:MDS} (a) and (b), the regularity condition holds, and under Assumption \ref{asm:MDS} (c) and (d), the concentration condition holds.

Therefore, going through the same procedure of proving Theorem 2.1 in \cite{abbe2022ellp} using modified Lemma \ref{lemma_H_1} and \ref{lemma_H_2} and under a slightly strong assumption \ref{asm:MDS}, we can show Theorem \ref{thm:modifed_l_p}. 

\end{proof}

\subsubsection{Main proof of of Theorem \ref{thm:mds}}
\label{proof:mds}
\begin{proof}
Define $\widebar\bw = \frac{1}{n}\sum_{i=1}^n\bw_i$. 
Similar to the proof of Theorem \ref{thm:spect-clust}, we can still apply a modified version of Corollary 2.1 of \cite{abbe2022ellp}, as shown in our Theorem \ref{thm:modifed_l_p}, under a stronger Assumption \ref{asm:MDS} compared with Assumption \ref{asm:SC}. 

Therefore, we can decompose the embedding matrix as:
\begin{equation}\label{eqn:0104}
\Ub\bLambda^{1/2}\Ob = \widebar{\Ub}\widebar{\bLambda}^{1/2} + \cH\left((\Jb\Wb) (\Jb\Xb)^\top\right) \widebar{\Ub}\widebar{\bLambda}^{-1/2} + \Eb_{13},    
\end{equation}
where $\Ob\in \bbR^{K\times K}$ is a rotation matrix. Define $$ \dfrac{1}{\sqrt{\log n}} \|\widebar{\Ub}\|_{2\rightarrow\infty}\|\widebar{\bLambda}^{1/2}\|_2 \lesssim \sqrt{\frac{\widebar\lambda_1}{n\log n}} =: \epsilon_3$$ 
then we have $\|\Eb_{13}\|_{2\rightarrow\infty} \lesssim \epsilon_3$ with probability $1-o(1)$ by Theorem \ref{thm:modifed_l_p}. 
We can further control the second term on the right-hand side of \eqref{eqn:0104} 
\begin{align*}
  & \cH\left((\Jb\Wb) (\Jb\Xb)^\top\right) \widebar{\Ub}\widebar{\bLambda}^{-1/2} \\
  = & \cH\left(\Wb (\Jb\widebar\Xb)^\top\right) \widebar{\Ub}\widebar{\bLambda}^{-1/2} + \cH\left((\Jb-\Ib_n)\Wb(\Jb\widebar\Xb)^\top\right) \widebar{\Ub}\widebar{\bLambda}^{-1/2} \\
  & \quad + \cH\left((\Jb\Wb) (\Jb\Wb)^\top\right) \widebar{\Ub}\widebar{\bLambda}^{-1/2} \\
  = &: \Eb_{2} + \Eb_{11} + \Eb_{12}. 
\end{align*}
For matrix $\Eb_2$, the rows are  independent sub-Gaussian random vectors with parameter $\sigma_{M} = \sqrt{\kappa}\sigma$
since $\|(\Jb\widebar\Xb)^\top\widebar{\Ub}\widebar{\bLambda}^{-1/2}\| \leq \sqrt{\kappa}\sigma$. 

For matrices $\Eb_{11}$ and $\Eb_{12}$, we would like to control the $2\rightarrow\infty$ norm for both.

Notice that 
$$\| \cH\left((\Jb-\Ib_n)\Wb(\Jb\widebar\Xb)^\top\right)\widebar{\Ub}\widebar{\bLambda}^{-1/2}\|_{2\rightarrow\infty} \leq \|(\Jb-\Ib_n)\Wb(\Jb\widebar\Xb)^\top\widebar{\Ub}\widebar{\bLambda}^{-1/2}  \|_{2\rightarrow\infty}.$$ 
Using Lemma \ref{lem:spectral_norm_MDS_mean}, we have 
\begin{align*}
    &\left\|(\Jb-\Ib_n)\Wb(\Jb\widebar\Xb)^\top\widebar{\Ub}\widebar{\bLambda}^{-1/2} \right\|_{2\rightarrow\infty} \\
    = & \left\|\frac{1}{n}\bone_n \widebar\bw^\top(\Jb\widebar\Xb)^\top\widebar\Ub\widebar{\bLambda}^{-1/2} \right\|_{2\rightarrow\infty}\\
    =& \frac{1}{n}\|\widebar\bLambda^{-1/2} \widebar\Ub^{\top}(\Jb\widebar\Xb)\widebar\bw\| \\
    = & O_{\bbP}\left(\frac{1}{n}\sqrt{\frac{K+\log n}{n}\kappa}\sigma; \log n\right).
\end{align*}
Define $\epsilon_1 = \dfrac{1}{n}\sqrt{\dfrac{K+\log n}{n}\kappa}\sigma$. Then we have 
$\|\Eb_{11}\|_{2\rightarrow\infty} \leq \epsilon_1$ with probability $1-o(1)$. 

Combining Lemma \ref{lemma_H_2} and \eqref{eqn:tight_concentration}, we have 
\begin{align*}
    \|\Eb_{12}\|_{2\rightarrow\infty} 
 = O_{\bbP}\left( \dfrac{\sigma^2}{\sqrt{\lambda_{K-1}}} \max\left\{\dfrac{p}{\sqrt{n}},\sqrt{p\log n},\log n \right\};\log n\right) 
\end{align*}
Therefore, we can define $\epsilon_2 =\dfrac{\sigma^2}{\sqrt{\lambda_{K-1}}}\max\left\{\dfrac{p}{\sqrt{n}},\sqrt{p\log n},\log n \right\} $. 

In order to apply Theorem \ref{thm-lloyd-guarantee}, similar to Section \ref{sec:SC}, we can verify that 
the conditions are satisfied under Assumption \ref{asm:MDS} (c). 
Then we can use Theorem \ref{thm-lloyd-guarantee} to get 
\begin{align*}
A_s \leq & \max \left \{ \exp \left( - \frac{\bar{s}^2}{32\kappa_0^4\beta \sigma^2} \right) , 
\exp\left( -c\frac{\bar{s}^{3}\sqrt{n}}{\sigma^3\left(\sqrt{\log n(p+\log n) }\right)} \right), \right.\\
& \left. \exp\left( -c\frac{n \bar{s}^{3}}{p\sigma^3} \right), \exp\left( -c\frac{\sqrt{\log n}\bar{s}}{\sigma} \right)
\right \}  ,
\end{align*}
where $\beta = \dfrac{\max_l n_l}{n}$ for all $s\geq 4\log n $ for some constant $c$ with probability 
\begin{align*}
& 1- \exp \left( - \frac{\bar{s}}{ \sigma}  \right) -  \exp\left( - \left[\left(\frac{n}{\log n(p+\log n)} \right)^{1/4}+\sqrt{\frac{n}{p}} \right] \left(\frac{\bar{s}}{\sigma} \right)^{3/2}  \right)\\
&- \exp\left( -(\log n)^{1/4}\sqrt{\frac{\bar{s}}{ \sigma}} \right) - o(1).
\end{align*}

\end{proof}

\subsection{Random dot product graphs}
We make use of the following lemma.
\begin{lemma}(Lemma 5, \cite{lyzinski:2014}) \label{lyzinski_lemma}
    Suppose $\Pb$ is of rank $r$ and that the non-zero eigenvalues of $\Pb$ are distinct. If there exists $0 < \eta < \frac{1}{2}$ such that $\gamma n \geq 4\sqrt{\cD \log\frac{n}{\eta}}$, then with probability at least $1 - 2\eta$, 
    \begin{equation}\label{RDPG_eps_bound}
    \| \Yb - \Yb^* \|_{2\to\infty} \leq \dfrac{85r\cD^3 \log \frac{n}{\eta}}{(\gamma n)^{7/2}}.
\end{equation}
\end{lemma}

The assumptions of Theorem~\ref{thm-RDPG} ensure that the conditions of Lemma~\ref{lyzinski_lemma} 
are satisfied. Setting $
\gee := \frac{85 r \cD^3 \log(n/\eta)}{(\gamma n)^{7/2}},$
the union bound implies that the probability that both \eqref{thm-lloyd-A_s_bound} and 
\eqref{RDPG_eps_bound} hold is at least $
1 - \delta(n,\sigma,\Delta_{\min},\gee) - 2\eta.$
Substituting this value of $\gee$ into \eqref{thm-lloyd-A_s_bound} yields the desired result.

\subsection{Community dynamic factor models}
Recall the setting and assumptions of CDFM in Theorem \ref{thm:cdfm}. Under Assumption \ref{Uematsu_assumps}, \cite{uematsu:2019} prove the following result on the estimation error of the loading matrix $\mvgL$.
\begin{lemma}{[\cite{uematsu:2019}, Lemma 6]}\label{weak-factor-lemma}
Suppose Assumption \ref{Uematsu_assumps} holds on the DFM \eqref{DFM}. Then, with probability at least $1 - O((d \vee T)^{-v})$,
\begin{equation}\label{lambda max bound}
\|\wh\mvgL - \mvgL^{(0)} \|_{\max} \leq C \left( \dfrac{\log( \max\{n,T\})}{T} \right)^{\frac{1}{2}},
\end{equation}
for some $C>0$.
\end{lemma}

Note that both \eqref{convergence of ests} and \eqref{lambda max bound} indicate that the estimate $\wh\mvgL$ is close to the rotated loading matrix $\mvgL^{(0)}$. Since our primary task is clustering, and rotation preserves the mixture structure when each mixing distribution is rotated the same way, we may simply ignore the rotation issue and treat $\mvgL^{(0)}$ as the true loading matrix. Hence, 
\begin{equation}
    \|\wh\mvgl_i - \mvgl_i\| \leq C r \left( \dfrac{\log(\max\{n,T\})}{T} \right)^{\frac{1}{2}}, \quad i \in [n],
\end{equation}
for some $C>0$. Thus, we may set $\gee = C r\left( \dfrac{\log(\max\{n,T\})}{T} \right)^{\frac{1}{2}}$. 

Assumption \ref{CDFM_gee_assumption} implies that 
$$
\sqrt{\alpha} \gD \geq \sqrt{C_{3}} \sqrt K r \sqrt{\dfrac{\log(\max\{n,T\})}{T}} = \dfrac{\sqrt{C_{3}}}{C}\sqrt K \gee
$$
and 
$$
\gD^2 \geq \sqrt{C_{3}} r^2 \sigma \sqrt{\dfrac{\log(\max\{n,T\})}{T}} =\dfrac{\sqrt{C_{3}}}{C} r\sigma\gee.
$$
As long as $C_{3} \geq C^2$, conditions \eqref{rho_sigma_assump} and \eqref{interaction_assump} of Theorem \ref{thm-lloyd-guarantee} are satisfied. Thus, an application of Theorem \ref{thm-lloyd-guarantee} finishes the proof.

\end{appendix}

\bibliographystyle{asa} 
\bibliography{references.bib}       %

\end{document}

%% file: preamble.tex
\theoremstyle{plain}

\newtheorem{theorem}{Theorem}[section]
\newtheorem{lemma}[theorem]{Lemma}

\newtheorem{thm}{Theorem}[section]

\newtheorem{corollary}[thm]{Corollary}

\theoremstyle{remark}
\newtheorem{rem}{Remark}[section]

\theoremstyle{remark}
\newtheorem{dfn}{Definition}[section]
\newtheorem{asm}{Assumption}[section]

\theoremstyle{remark}

\let\ga=\alpha    \let\gee=\epsilon
     \let\gl=\lambda            \let\gt=\tau 
  \let\gz=\zeta
\let\gC=\Gamma \let\gD=\Delta

\newcommand{\eb}{\mathbf{e}}

\newcommand{\pb}{\mathbf{p}}

\newcommand{\vb}{\mathbf{v}}
\newcommand{\yb}{\bm{y}}
\newcommand{\zb}{\mathbf{z}}

\newcommand{\ba}{\bm{a}}

\newcommand{\bc}{\bm{c}}

\newcommand{\be}{\bm{e}}

\newcommand{\bu}{\bm{u}}
\newcommand{\bv}{\bm{v}}
\newcommand{\bw}{\bm{w}}
\newcommand{\bx}{\bm{x}}
\newcommand{\by}{\bm{y}}

\newcommand{\Ab}{\mathbf{A}}
\newcommand{\Bb}{\mathbf{B}}

\newcommand{\Eb}{\mathbf{E}}

\newcommand{\Gb}{\mathbf{G}}
\newcommand{\Hb}{\mathbf{H}}
\newcommand{\Ib}{\mathbf{I}}
\newcommand{\Jb}{\mathbf{J}}

\newcommand{\Mb}{\mathbf{M}}

\newcommand{\Ob}{\mathbf{O}}
\newcommand{\Pb}{\mathbf{P}}
\newcommand{\Qb}{\mathbf{Q}}
\newcommand{\Rb}{\mathbf{R}}
\newcommand{\Sbb}{\mathbf{S}}

\newcommand{\Ub}{\mathbf{U}}
\newcommand{\Vb}{\mathbf{V}}
\newcommand{\Wb}{\mathbf{W}}
\newcommand{\Xb}{\mathbf{X}}
\newcommand{\Yb}{\mathbf{Y}}
\newcommand{\Zb}{\mathbf{Z}}

\newcommand{\bB}{\bm{B}}
\newcommand{\bC}{\bm{C}}

\newcommand{\bN}{\bm{N}}

\newcommand{\bR}{\bm{R}}

\newcommand{\bmu}{\bm{\mu}}

\newcommand{\bphi}{\bm{\phi}}

\newcommand{\bzeta}{\bm{\zeta}}

\newcommand{\bLambda}{\bm{\Lambda}}

\newcommand{\bbE}{\mathbb{E}}

\newcommand{\bbP}{\mathbb{P}}

\newcommand{\bbR}{\mathbb{R}}
\newcommand{\bbS}{\mathbb{S}}

\newcommand{\bbZ}{\mathbb{Z}}

\newcommand{\cC}{\mathcal{C}}
\newcommand{\cD}{\mathcal{D}}

\newcommand{\cF}{\mathcal{F}}
\newcommand{\cG}{\mathcal{G}}
\newcommand{\cH}{\mathcal{H}}
\newcommand{\cI}{\mathcal{I}}

\newcommand{\cN}{\mathcal{N}}

\newcommand{\cX}{\mathcal{X}}

\newcommand{\fp}{\mathfrak{p}}

\newcommand{\ind}{\mathds{1}}

\newcommand{\mvzero}{\boldsymbol{0}}

\newcommand{\mvD}{\boldsymbol{D}}\newcommand{\mvE}{\boldsymbol{E}}\newcommand{\mvF}{\boldsymbol{F}}
\newcommand{\mvI}{\boldsymbol{I}}

\newcommand{\mvQ}{\boldsymbol{Q}}

\newcommand{\mvX}{\boldsymbol{X}}

\newcommand{\mvf}{\boldsymbol{f}}

\newcommand{\mvu}{\boldsymbol{u}}

\newcommand{\mvgee}{\boldsymbol{\epsilon}}

\newcommand{\mvgl}{\boldsymbol{\lambda}}\newcommand{\mvgL}{\boldsymbol{\Lambda}}

\newcommand{\mvgP}{\boldsymbol{\Pi}}
\newcommand{\mvgS}{\boldsymbol{\Sigma}}

\newcommand{\mvgz}{\boldsymbol{\zeta}}
\newcommand{\mvdelta}{\boldsymbol{\delta}}

\newcommand{\mvlambda}{\boldsymbol{\lambda}}\newcommand{\mvLambda}{\boldsymbol{\Lambda}}\newcommand{\mvmu}{\boldsymbol{\mu}}
\newcommand{\mvnu}{\boldsymbol{\nu}}

\newcommand{\mvxi}{\boldsymbol{\xi}}

\newcommand{\mvvarphi}{\boldsymbol{\varphi}}\newcommand{\mvPsi}{\boldsymbol{\Psi}}

\newcommand{\s}{^{(s)}}

\newcommand{\zhat}{\widehat{z}}
\newcommand{\nhat}{\widehat{n}}
\newcommand{\wb}{\bar}
\newcommand{\wh}{\widehat}

\DeclareMathOperator{\E}{\mathds{E}}
\DeclareMathOperator{\pr}{\mathds{P}}

\DeclareMathOperator{\var}{Var}

\newcommand{\set}[1]{\left\{#1\right\}}

\newcommand{\probc}{\stackrel{\mathrm{P}}{\longrightarrow}}

\newcommand{\weakc}{\stackrel{\mathrm{w}}{\longrightarrow}}
\newcommand{\convas}{\stackrel{\mathrm{a.s.}}{\longrightarrow}}

\newcommand{\bzero}{\mathbf{0}}	
\newcommand{\bone}{\mathbf{1}}	

\newcommand{\bern}{\mathsf{Bernoulli}}